%% file: main.tex
\documentclass{article}
\usepackage{amssymb}
\usepackage{amsmath}
\usepackage{amsthm}

\usepackage[preprint]{corl_2025} 

\usepackage{color}
\usepackage{bm}
\usepackage{cleveref}
\usepackage{algorithm}
\usepackage{algpseudocode}
\usepackage{wrapfig}
\usepackage{booktabs}
\usepackage{graphicx}
\usepackage{xspace}
\usepackage{subcaption}
\usepackage{tabularx}
\usepackage{enumitem}

\input{macros}

\algtext*{EndIf}   
\algtext*{EndFor}  
\algtext*{EndWhile} 
\algtext*{EndFunction} 

\theoremstyle{plain}      

\newtheoremstyle{boldremark}  
  {12pt}                      
  {12pt}                      
  {\normalfont}               
  {}                          
  {\bfseries}                 
  {.}                         
  {5pt plus 1pt minus 1pt}    
  {}                          
\theoremstyle{boldremark}

\newcommand{\loose}{\looseness=-1}

\newcommand{\arxiv}[1]{}
\newcommand{\forapp}[1]{}
\newtoggle{arxiv}
\togglefalse{arxiv}
\newtoggle{arxiv2} 
\toggletrue{arxiv2}

\title{Steering Your Diffusion Policy with Latent Space Reinforcement Learning}

\author{
  Andrew Wagenmaker\thanks{Core contributor. Please see appendix for individual contributions. Correspondence to: Andrew Wagenmaker \{\texttt{ajwagen@berkeley.edu}\}.} \\
  UC Berkeley
  \And
  Mitsuhiko Nakamoto$^{*}$ \\
  UC Berkeley
  \And
  Yunchu Zhang$^*$ \\
  University of Washington
  \And
  Seohong Park \\
  UC Berkeley
  \And
  Waleed Yagoub \\
  University of Washington
  \And
  Anusha Nagabandi \\
  Amazon 
  \And
  Abhishek Gupta$^*$ \\
  University of Washington
  \And
  Sergey Levine$^*$ \\
  UC Berkeley
}

\begin{document}
\maketitle


\begin{abstract}
Robotic control policies learned from human demonstrations have achieved impressive results in many real-world applications. However, in scenarios where initial performance is not satisfactory, as is often the case in novel open-world settings, such behavioral cloning (BC)-learned policies typically require collecting additional human demonstrations to further improve their behavior---an expensive and time-consuming process. In contrast, reinforcement learning (RL) holds the promise of enabling autonomous online policy improvement, but often falls short of achieving this due to the large number of samples it typically requires. In this work we take steps towards enabling fast autonomous adaptation of BC-trained policies via efficient real-world RL. Focusing in particular on diffusion policies---a state-of-the-art BC methodology---we propose \emph{diffusion steering via reinforcement learning} (\algname): adapting the BC policy by running RL over its latent-noise space. We show that \algname is highly sample efficient, requires only black-box access to the BC policy, and enables effective real-world autonomous policy improvement. Furthermore, \algname avoids many of the challenges associated with finetuning diffusion policies, obviating the need to modify the weights of the base policy at all. We demonstrate \algname on simulated benchmarks, real-world robotic tasks, and for adapting pretrained generalist policies, illustrating its sample efficiency and effective performance at real-world policy improvement.
Website: \url{https://diffusion-steering.github.io}.
\end{abstract}


\input{body/introduction}
\input{body/related}

\input{body/method}

\input{body/experiments_arxiv}

\section{Discussion}
Diffusion policies have seen widespread success in real-world robotic control problems, yet developing efficient and stable approaches to improve them in deployment has remained a persistent challenge.
In this work we have proposed a method, \algname, which enables efficient RL-based improvement of diffusion policies, and demonstrates strong performance in both simulated and real-world settings. We believe this work opens the door for several interesting follow-up directions:
\begin{itemize}
    \item While in this work we have focused on optimizing the input noise, one could consider, for example, also modifying the input observation or prompt given to the policy. Can this lead to effective policy improvement as well, and perhaps enable \algname-like improvement with autoregressive transformer-based policies?
    \item While we have focused on control domains, the underlying principle of \algname---running RL over the latent-noise space of a diffusion model to optimize some objective---may have much broader potential. Could \algname be applied to, for example, image generation or protein modeling, other domains where diffusion models have found much success?
    \item What are the theoretical properties of diffusion steering? Can we make any claims about the expressivity of the noise-space optimization we employ or the structure of the diffusion policy's noise space and ease of learning a policy over it?
\end{itemize}

\newpage

\iftoggle{arxiv2}{}{
\section*{Limitations}

Our experiments show that \algname provides an effective and highly efficient method for improving diffusion and flow policies using either online or offline experience. However, our approach does have a number of limitations. First, the exploration capabilities of \algname are inherently determined by the underlying diffusion policy, and while this seems to work well in practice, we do not have a guarantee that \emph{all} diffusion policies are steerable, nor that all such policies provide adequate exploration. A highly concentrated action distribution (e.g., a policy trained on very narrow data) might not provide enough ``options'' for our method to select from, limiting its ability to improve, and our approach does not currently offer a clear way to quantify in advance how much improvement is possible with a given diffusion policy. Studying this question both empirically and theoretically could help us characterize the steerability of diffusion policies. Our method also has similar limitations to any RL approach: it requires reward signals, online rollouts (for the online RL variant), and resets (if learning in the real world). While the efficiency of our approach alleviates some of the challenges associated with RL in the real world, it still introduces additional requirements beyond the standard imitation learning methods that can be used to train the base diffusion policy. Reducing these limitations, for example by automating rewards or further improving sample efficiency, is an important direction for future work. We hope that studying these open questions will lead to even more practical methods for robotic RL, and that \algname will provide a practical and useful tool for autonomous improvement in robotic learning.}

\acknowledgments{This research was partly supported by ONR N00014-22-1-2773 and N00014-25-1-2060. YZ, AG and WY are partly supported by funding from Amazon. We would like to thank Pranav Atreya, Homer Walke, and Laura Smith for their advice on the $\pi_0$ setup. This research used the Savio computational cluster resource provided by the Berkeley Research Computing
program at UC Berkeley.}


\bibliography{bibliography}  

\newpage
\appendix
\input{body/appendix}

\end{document}

%% file: macros.tex
\newcommand{\algname}{\textsc{Dsrl}\xspace}
\newcommand{\algnamena}{\textsc{Dsrl-Na}\xspace}

\newcommand{\cS}{\mathcal{S}}
\newcommand{\cA}{\mathcal{A}}

\newcommand{\R}{\mathbb{R}}
\newcommand{\cN}{\mathcal{N}}
\newcommand{\cM}{\mathcal{M}}
\newcommand{\Exp}{\mathbb{E}}

\newcommand{\cW}{\mathcal{W}}

\newcommand{\pidp}{\pi_{\mathrm{dp}}}
\newcommand{\pidpw}{\pi_{\mathrm{dp}}^{\scriptscriptstyle \cW}}

\newcommand{\Pdp}{P^{\scriptscriptstyle \cW}}
\newcommand{\rdp}{r^{\scriptscriptstyle \cW}}
\newcommand{\cMdp}{\mathcal{M}^{\scriptscriptstyle \cW}}
\newcommand{\piw}{\pi^{\scriptscriptstyle \cW}}
\newcommand{\rmd}{\mathrm{d}}

\newcommand{\bw}{\bm{w}}
\newcommand{\bx}{\bm{x}}
\newcommand{\bs}{\bm{s}}
\newcommand{\ba}{\bm{a}}
\newcommand{\beps}{\bm{\epsilon}}
\newcommand{\epsnet}{\beps_{\theta}^{(t)}}

\newcommand{\frakD}{\mathfrak{D}}
\newcommand{\frakB}{\mathfrak{B}}

\newcommand{\Qw}{Q^{\scriptscriptstyle \cW}}
\newcommand{\Qa}{Q^{\scriptscriptstyle \cA}}
\newcommand{\Qabar}{\bar{Q}^{\scriptscriptstyle \cA}}

%% file: body/introduction.tex
\section{Introduction}

Robotic learning methods have achieved impressive results on many real-world applications by shifting the burden of controller design from human engineering to data-driven end-to-end learning \cite{stepputtis2020language,shafiullah2022behavior,gu2023rt,team2024octo,zhao2024aloha,black2024pi_0}. Learning policies from expert data via supervised learning, often referred to as behavioral cloning (BC), has been particularly effective in translating large, offline demonstration datasets into robotic control policies. Such methods are appealing because of their simplicity, scalability, and applicability across a broad range of generative model parameterizations, ranging from diffusion or flow models~\cite{chi2023diffusion, black2024pi_0} to autoregressive transformers~\cite{kim2024openvla}. \iftoggle{arxiv}{However, since these methods require expert data to learn, they are unable to directly make use of additional experience collected during deployment to improve policy behavior.}{However, since these methods require expert data to learn, they are unable to directly utilize experience collected during deployment.} In settings where the BC policy does not solve the goal task perfectly, as is common in many open-world deployments, utilizing this experience to further refine and improve the policy is highly desirable.\loose

Reinforcement learning (RL) can, in principle, achieve this type of policy improvement. While learning policies via RL from scratch is often too inefficient for practical use, RL has served as an effective \emph{finetuning} tool in domains such as language modeling, showing impressive results adapting models pretrained on large corpora of offline ``demonstrations'' \cite{ouyang2022training,bai2022training,team2025kimi,guo2025deepseek}. However, this finetuning process is still expensive in terms of both compute and time (i.e., samples), making it difficult to directly apply to the adaptation of real-world robotic control policies. 
Given this, developing methods for RL-based finetuning of pretrained BC policies that are simple, stable, broadly applicable, and efficient enough to apply to real-world robot learning has remained a persistent challenge.

\begin{figure}[tbp]
    \centering
    \includegraphics[width=\textwidth]{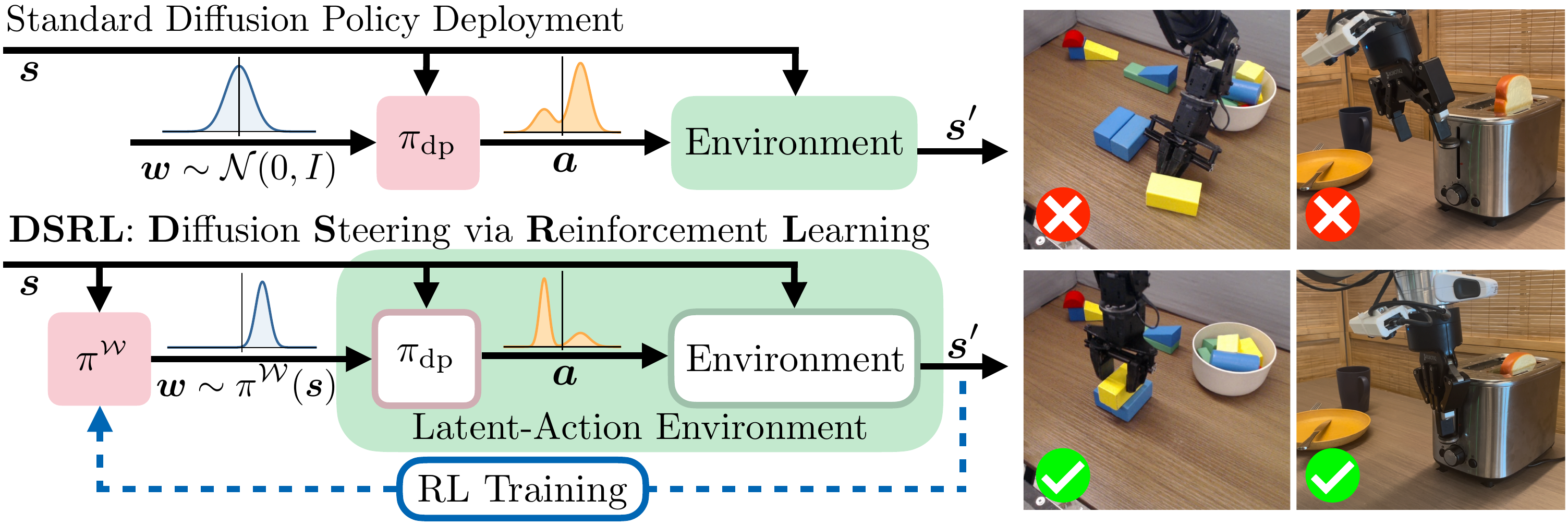}
    \caption{Overview of our proposed approach, \textbf{D}iffusion \textbf{S}teering via \textbf{R}einforcement \textbf{L}earning (\algname). Standard deployment of a BC-trained diffusion policy $\pidp$ first samples noise $\bw \sim \cN(0,I)$ that is then denoised through the reverse diffusion process to produce an action $\ba$. We propose modifying the initial distribution of $\bw$ with an RL-trained latent-noise space policy $\piw$, that instead of choosing $\bw \sim \cN(0,I)$, chooses $\bw$ to \emph{steer} the distribution of actions produced by $\pidp$ in a desirable way, enabling highly sample efficient real-world adaptation of robot policies.\loose 
    }
    \label{fig:main}
\vspace{-1.5em}
\end{figure}

In this work, we take steps towards addressing this challenge, and seek to understand how we can utilize novel experience to adapt pretrained robotic control policies in a simple and efficient way.
Focusing in particular on BC policies parameterized as diffusion models \cite{sohl2015deep,ho2020denoising,song2020denoising}, which has become the de-facto standard policy parameterization in robotics \cite{chi2023diffusion,ankile2024juicer,zhao2024aloha,ze20243d,sridhar2024nomad,dasari2024ingredients,team2024octo,black2024pi_0,bjorck2025gr00t},
we propose a lightweight approach that overcomes many of the challenges faced by existing RL-based finetuning methods. In particular, rather than modifying the \emph{weights} of a pretrained diffusion-based BC policy, we instead modify its \emph{sampling process}, altering the input noise distribution the diffusion model utilizes to generate samples. We show that, by running RL over this input noise distribution, we can quickly learn to steer the policy's behavior as new experience is collected online, and that this approach is stable and highly sample efficient, while being surprisingly expressive. Our approach relies only on forward passes of the denoising process and is fully black-box, not requiring direct access to the policy's weights at all. This allows for the adaptation of policies through pure API access and eliminates the need for potentially unstable back-propagation through the diffusion chain, a primary challenge in applying standard RL finetuning to diffusion policies \cite{ren2024diffusion,mark2024policy,park2025flow}. Furthermore, our approach avoids the challenges associated with finetuning the large, complex policy architectures often utilized in modern robot learning settings, and instead only requires training much smaller latent-space policies. See \Cref{fig:main} for an overview of our approach.

Our main contribution is to formalize this \emph{diffusion steering} process, which we refer to as \emph{diffusion steering via reinforcement learning} (\algname), and provide a practical instantiation that can be used for adapting robot policies in the real world. While we argue that virtually any RL algorithm can be utilized for diffusion steering, we develop an actor-critic-based procedure that takes advantage of structure in the diffusion policy to enable highly sample-efficient learning from both online and offline data. We empirically demonstrate the effectiveness of \algname in both simulation and on real-world robotic embodiments, demonstrating state-of-the-art performance in adapting behavior both from online interaction and offline data. 
We show that our approach also leads to sample-efficient real-world refinement of robot policies, in some cases improving success from $20\%$ to $90\%$ with less than $50$ episodes of online interaction. Furthermore, we demonstrate that our approach can be used to improve the behavior of existing state-of-the-art generalist robot policies, in particular $\pi_0$ \cite{black2024pi_0}, in real-world deployment. 

%% file: body/related.tex
\iftoggle{arxiv2}{}{\vspace{-0.5em}}
\section{Related Work}
\iftoggle{arxiv2}{}{\vspace{-0.5em}}

\textbf{Behavioral cloning and diffusion models in control \& robotics.}
BC methods have seen widespread use throughout the control and robotics communities \cite{argall2009survey,ross2011reduction,bojarski2016end,zhang2018deep,rahmatizadeh2018vision,mandlekar2021matters}. While we focus primarily on diffusion-based BC, other approaches---for example, autoregressive next-token prediction---have found success as well \cite{stepputtis2020language,brohan2022rt,nair2022learning,gu2023rt,shafiullah2022behavior,cui2022play,black2023zero,kim2024openvla}.
Diffusion-based BC has led to state-of-the-art results in real-world robotic domains ranging from single-task applications \cite{chi2023diffusion,ankile2024juicer,zhao2024aloha,ze20243d,sridhar2024nomad,dasari2024ingredients} to ``generalist'' policies \cite{team2024octo,black2024pi_0,bjorck2025gr00t}. 
Diffusion models have also been applied to control settings in a \emph{model-based} fashion, enabling planning-based control \cite{janner2022planning,ajay2022conditional,zheng2023guided,li2023hierarchical,liang2023adaptdiffuser,suh2023fighting,chen2024simple,wang2024poco}. 
In contrast to these works, our focus is not on developing better methodology for training diffusion policies, but on adapting already-trained diffusion policies as new experience is collected.\loose

\textbf{Reinforcement learning with diffusion.}
Beyond the pure BC setting, many works have considered applying diffusion models in RL settings where the learner has a reward signal they wish to maximize. 
In the purely offline setting, approaches include training a diffusion policy on an objective weighted by the value of an action \cite{lu2023contrastive,kang2023efficient,zhang2025energy,ding2024diffusion}, optimizing a diffusion policy directly to maximize a reward \cite{wang2022diffusion,he2023diffcps,ding2023consistency,ada2024diffusion,zhang2024entropy,park2025flow}, rejection sampling of the actions produced by the diffusion policy \cite{chen2022offline,hansen2023idql,he2024aligniql,nakamoto2024steering}, in addition to several other approaches \cite{venkatraman2023reasoning,chen2024aligning,chen2024diffusion,chen2023score,mao2024diffusion}.
In the online setting, past work has applied PPO \cite{schulman2017proximal} to finetune the first few steps of a diffusion policy's denoising process \cite{ren2024diffusion}, trained the diffusion policy by matching the score of the $Q$-function \cite{psenka2023learning}, or applied iterative BC to clone actions that lead to large $Q$-function values \cite{yang2023policy,li2024learning,mark2024policy}.
Other works seek to post-process the outputs of a pretrained diffusion policy by learning an additive residual policy with RL \cite{ankile2024imitation, yuan2024policy}.
Compared to these works, we take a different approach and, instead of modifying the diffusion policy's weights or post-processing the actions, adjust the sampling process by modifying the input noise distribution. As we will show, this leads to faster policy improvement and does not require access to the policy's weights.\loose

\textbf{Noise optimization in generative modeling.}
Our proposed approach optimizes the noise sampled at the start of the diffusion policy's denoising process. Other works have considered similar strategies for improving the behavior of diffusion models. 
In image diffusion, it has been shown that optimizing the initial noise of the diffusion process to maximize some metric on image quality can lead to substantial gains in performance \cite{eyring2024reno,mao2024lottery,samuel2024generating} or mimic the effects of diffusion guidance \cite{ahn2024noise}.
In addition to considering a very different application domain, these works differ methodologically from our work as we consider applying RL to optimize the initial noise, while these works consider non-RL based approaches. 
In the control domain, \cite{singh2020parrot} trains a \emph{normalizing flow}-based policy \cite{dinh2016density} on offline data, then runs RL over the noise space of this policy to improve its performance online. The key difference between this work and ours is that we utilize a diffusion or flow matching model rather than a normalizing flow. Normalizing flows are invertible by construction,
which ensures that the agent loses no expressivity over actions when optimizing the noise. As this property is not guaranteed to hold for diffusion models, it is much less clear that optimizing over the noise space of a diffusion policy is expressive enough for effective online adaptation.
More importantly, diffusion models have seen much greater success in practice---while there are many examples successfully applying diffusion policies to robotic control, normalizing flows have been shown to perform poorly in such settings \cite{dalal2021accelerating,shafiullah2022behavior}. Given this, developing practical approaches to effectively adapting diffusion models by modifying their input noise is both a technical advancement from \cite{singh2020parrot}, and, we believe, of much greater practical relevance.\loose

%% file: body/method.tex

\iftoggle{arxiv2}{}{\vspace{-0.5em}}
\section{Preliminaries}\label{sec:prelim}
\iftoggle{arxiv2}{}{\vspace{-0.5em}}

\iftoggle{arxiv}{\paragraph{Markov decision processes.}}{\textbf{Markov decision processes.}}
A Markov Decision Process (MDP) is denoted by a tuple \mbox{$\cM = (\cS, \cA, P, p_0, r, \gamma)$},
where $\cS$ is a set of states, $\cA$ a set of actions (which we assume to be a subset of $\R^d$), $P : \cS \times \cA \rightarrow \triangle_{\cS}$ a transition kernel, $p_0 \in \triangle_{\cS}$ the initial state distribution, $r : \cS \times \cA \rightarrow \R$ a reward function, and $\gamma \in [0,1]$ the discount factor.
An episode begins by drawing state $\bs_0 \sim p_0$, and proceeds with, at state $\bs$, the agent choosing action $\ba$, receiving reward $r(\bs,\ba)$, and transitioning to $\bs' \sim P(\cdot \mid \bs, \ba)$.
A policy $\pi : \cS \rightarrow \triangle_{\cA}$ denotes a mapping from states to actions. 
The $Q$-function for policy $\pi$, $Q^\pi(\bs,\ba)$, denotes the expected discounted return of policy $\pi$ from taking action $\ba$ in state $\bs$, $Q^\pi(\bs,\ba) := \Exp^{\pi}[\sum_{t \ge 0} \gamma^t r(\bs_t, \ba_t) \mid \bs_0 = \bs, \ba_0 = \ba]$. The typical objective in RL is to find a policy that achieves maximum expected discounted return.

\iftoggle{arxiv}{\paragraph{Diffusion models and flow matching.}}{\textbf{Diffusion models and flow matching.}}
Diffusion models are generative models that aim to transform an easy-to-sample-from distribution into a more complex goal distribution $q(\bx_0)$ \cite{sohl2015deep,ho2020denoising}. While various instantiations exist, denoising diffusion \cite{ho2020denoising,song2020denoising} has emerged as perhaps the most popular. Denoising diffusion models this transformation process with a distribution $p_\theta(\bx_0)$ parameterized as
$p_{\theta}(\bx_0) = \int p_\theta(\bx_{0:T}) \rmd \bx_{1:T}$, for $p_\theta(\bx_{0:T}) = p_{\theta}(\bx_T) \cdot \prod_{t=1}^T p_{\theta}^{(t)}(\bx_{t-1} \mid \bx_t)$, and typically takes $p_{\theta}^{(t)}(\bx_{t-1} \mid \bx_t)$ to be a normal distribution.
The \emph{forward diffusion process} adds noise to samples $\bx_0 \sim q(\bx_0)$, and trains a network $\epsnet$ to denoise these samples.
The \emph{reverse diffusion process}---the denoising or sampling process---begins by sampling $\bx_T \sim \cN(0,I)$, and then applies the update\loose
\begin{align}\label{eq:reverse_process}
\bx_{t-1} = \alpha_t (\bx_t- \beta_t \epsnet(\bx_t)) + \sigma_t \beps_t, \quad \beps_t \sim \cN(0,I),
\end{align}
for $\alpha_t, \beta_t,$ and $\sigma_t$ some coefficients. This ultimately produces a ``denoised'' sample $\bx_0$ distributed approximately as $q(\bx_0)$. 
The reverse diffusion process can be run with both $\sigma_t > 0$ (typically referred to as DDPM sampling \cite{ho2020denoising}) or $\sigma_t = 0$ (DDIM sampling \cite{song2020denoising}). Notably, DDIM sampling makes the reverse process \emph{deterministic} given the initial noise sample $\bx_T$ and, furthermore, can often lead to more efficient generation by reducing the number of required denoising steps \cite{song2020denoising}, a property especially useful in real-world robotic deployment where it is critical that actions are generated at a high enough control frequency \cite{chi2023diffusion}. 
\iftoggle{arxiv}{With DDIM sampling, we can therefore think of the reverse process as simply a function that maps a ``latent'' $\bx_T$ to some point in the support of $q(\bx_0)$.}{}

Beyond denoising diffusion, \emph{flow matching} is a similar generative modeling technique---also based on an iterative denoising procedure---that has demonstrated impressive results across a range of domains \cite{lipman2022flow,liu2022flow,albergo2022building,black2024pi_0,park2025flow}. While a full discussion of flow matching is beyond the scope of this work (see e.g. \cite{lipman2024flow}) we highlight that, similar to denoising diffusion, flow matching operates by first sampling noise from some tractable distribution, and then processes this noise through a (by construction) deterministic process to produce a sample from a goal distribution.

\iftoggle{arxiv}{\paragraph{Behavioral cloning with diffusion and flow policies.}}{\textbf{Behavioral cloning with diffusion and flow policies.}}
BC aims to learn a policy that models the state-conditional distribution of actions from a dataset of demonstrations $\frakD = \{ (\bs^i, \ba^i) \}_{i=1}^N$. This is simply a generative modeling problem and diffusion models or flow matching can be applied directly, with $\bx_0 \leftarrow \ba^i$ and the BC policy, $\pidp$, the entire denoising process, \eqref{eq:reverse_process}.
As noted, for fixed input noise $\bx_T \leftarrow \bw$, a DDIM or flow-based policy is a deterministic process. Given this, for a diffusion policy $\pidp$ utilizing DDIM sampling or a flow-based policy, we define $\pidpw : \cS  \times \cW \rightarrow \cA$ the deterministic output of the denoising process at state $\bs$ when input noise $\bw$ initializes the denoising process, and denote by $\cW := \R^d$ the latent-noise space.

\iftoggle{arxiv}{\paragraph{Problem setting.}}{\textbf{Problem setting.}}
In this work we assume we are given a pretrained diffusion policy $\pidp$, and our goal is to adapt its behavior to maximize some reward $r$ in environment $\cM$, the standard objective of RL. 
While we do not make explicit assumptions on the behavior of $\pidp$, for our approach to be effective $\pidp$ must be ``steerable'' in a sense that will become clear (please see \Cref{sec:understanding_ds} for further discussion of when $\pidp$ is steerable).
We consider both the online setting---where the learner can interact with $\cM$---as well as the offline setting---where the learner is given a dataset of transitions from $\cM$.
We assume we may select any $\bs \in \cS$ and $\bw \in \cW$ and observe $\pidpw(\bs,\bw)$, but do not assume any other access to $\pidpw$ (for example, its weights or intermediate steps in its denoising process).\loose

\iftoggle{arxiv2}{}{\vspace{-0.5em}}
\section{Diffusion Steering via Reinforcement Learning}
\iftoggle{arxiv2}{}{\vspace{-0.5em}}

\begin{wrapfigure}{r}{0.4\textwidth}
\vspace{-1em}
\centering
\includegraphics[width=0.38\textwidth]{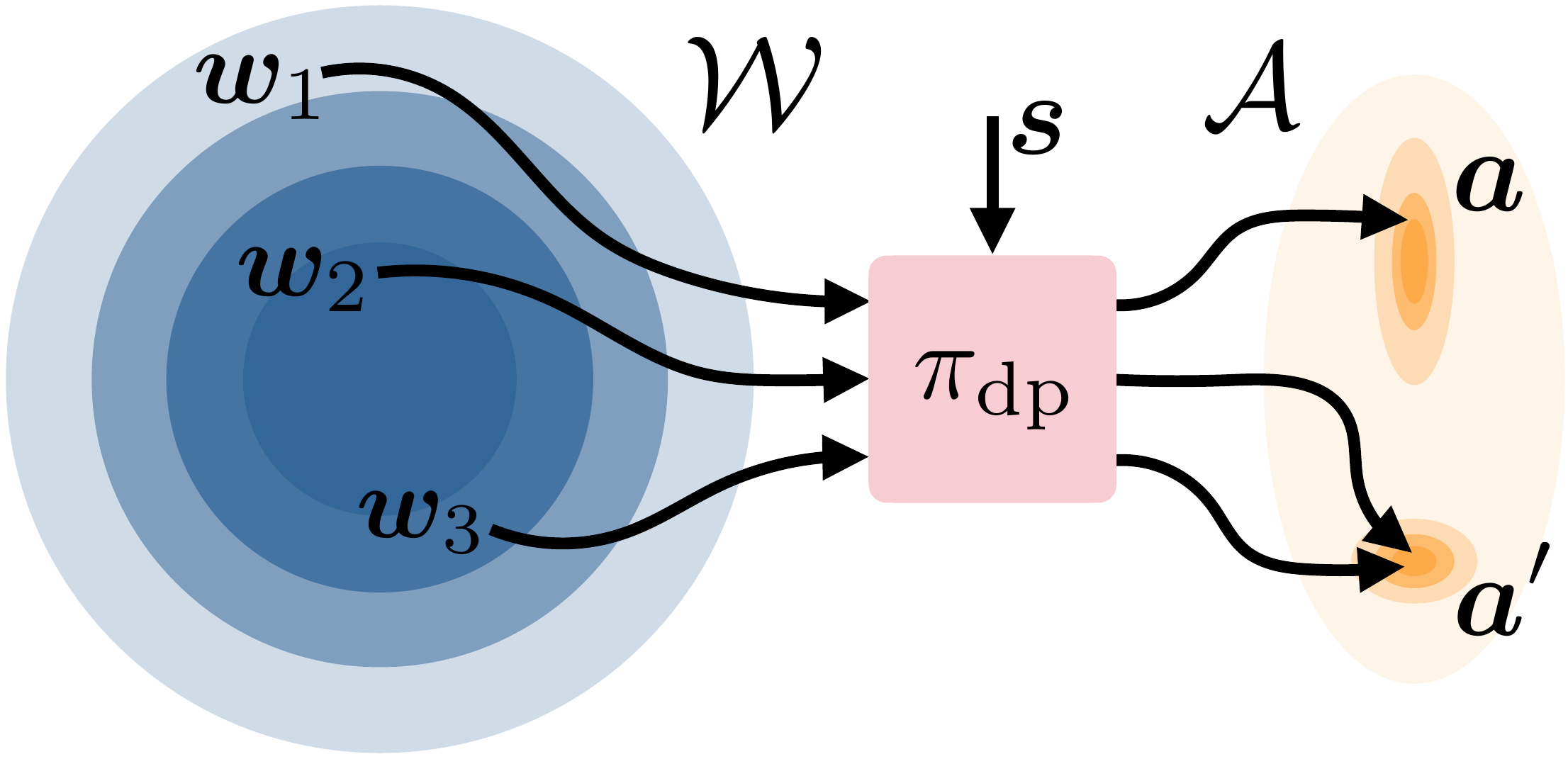}
\caption{By selecting which point in the latent-noise space to denoise, we can steer the action produced by $\pidp$ to a desired mode. At the same time, different points in the latent-noise space may in some cases be denoised to the same point in the original action space, allowing one to infer the behavior of noise actions $\bw_2$ and $\bw_3$ at $\bs$ given only observation $(\bs,\ba')$, forming the basis of our noise aliasing approach (\Cref{sec:noise_aliasing}).}
\label{fig:noise_aliasing_illustration}
\vspace{-0.5em}
\end{wrapfigure}

In typical operation of a diffusion policy, $\bw$ is sampled from a standard Gaussian, $\bw \sim \cN(0,I)$, and processed through $\pidpw$ to produce an output action. $\pidpw$ is trained so that the resulting action distribution, $\ba \sim \pidpw(\bs,\bw)$ for $\bw \sim \cN(0,I)$, matches the action distribution of the demonstration data at $\bs$. At deployment, this only holds true, however, if $\bw \sim \cN(0,I)$---if $\bw$ is chosen some other way, $\pidp$ will not necessarily produce actions that match the distribution of the demonstrator.
This suggests a straightforward way to modify the behavior of $\pidp$ is to  modify the distribution of $\bw$, and instead of $\bw \sim \cN(0,I)$, choose $\bw$ so that the resulting action $\ba \leftarrow \pidpw(\bs,\bw)$ leads to a desired result (see \Cref{fig:noise_aliasing_illustration}). 
This observation forms the backbone of our proposed approach, \emph{diffusion steering}: rather than modifying the weights of the diffusion policy or postprocessing its output, diffusion steering simply ``steers'' the diffusion policy to produce desired actions by altering its input noise distribution.\loose

\iftoggle{arxiv2}{}{\vspace{-0.5em}}
\subsection{Diffusion Steering as Latent-Noise Space Policy Optimization}
\iftoggle{arxiv2}{}{\vspace{-0.5em}}
While selecting $\bw$ may give us some control over the output of $\pidp$, it remains to determine \emph{which} $\bw$ we should choose. To answer this, we propose reinterpreting the role of $\pidpw$ from a \emph{policy} to an \emph{action space transformation}.
At any state $\bs$, we can produce an action $\ba \in \cA$ by selecting $\bw \in \cW$ and setting $\ba \leftarrow \pidpw(\bs,\bw)$. 
We can think of this mapping as transforming the action space of $\cM$ from the original action space $\cA$ to the latent-noise space $\cW$. In particular, defining
\begin{align*}
\Pdp(\cdot \mid \bs,\bw) := P(\cdot \mid \bs, \pidpw(\bs,\bw)) \quad \text{and} \quad \rdp(\bs,\bw) := r(\bs, \pidpw(\bs,\bw)),
\end{align*}
we see that $\cMdp := (\cS, \cW, \Pdp, p_0, \rdp, \gamma)$ forms a transformed version of our original MDP, which we refer to as the \emph{latent-action MDP}. 
Notably, we can treat interaction with $\cMdp$ just as we would treat interaction with $\cM$, completely black-boxing the contribution of $\pidpw$---we now take actions in $\cW$, and filter them through $\pidpw$ to obtain some action in $\cA$ which we play in $\cM$, but this filtering process is completely encapsulated in the transformed environment. Given this, we may consider $\pidpw$ simply part of the environment, and choosing $\bw$ reduces to a standard policy optimization problem over $\cMdp$, allowing us to apply virtually \emph{any} RL algorithm to learn a latent-noise space policy. 
We refer to this procedure---adapting the behavior of $\pidp$ by running RL over its latent-noise space---as \textbf{D}iffusion \textbf{S}teering via \textbf{R}einforcement \textbf{L}earning (\algname). 

\iftoggle{arxiv}{\paragraph{Computational efficiency of \algname.}}{\textbf{Computational efficiency of \algname.}}
Typical policy optimization procedures require back-propagating through a policy to compute a gradient direction in which to update the parameters. Such methods are challenging to apply to diffusion policies, however, as the diffusion policy generates actions via multi-step denoising \eqref{eq:reverse_process}, and back-propagation through this entire process is computationally intensive and numerically unstable \cite{ren2024diffusion,mark2024policy,park2025flow}. 
\algname avoids this issue entirely by lifting policy optimization from the diffusion policy itself to a secondary policy operating in the diffusion policy's latent-noise space. 
This latent-space policy can be parameterized in whatever form is most convenient, and only requires forward passes of the diffusion policy's denoising process to optimize.
This also allows \algname to circumvent the need to optimize large diffusion policies common in modern applications, yielding further computational improvements over standard finetuning procedures, and reduces the model access required by \algname to only choosing $\bw \in \cW$ and computing $\ba \leftarrow \pidpw(\bs,\bw)$, rather than requiring access to the weights of $\pidpw$ or intermediate steps within the denoising process. Such weak access could enable, for example, adaptation of proprietary models where the base weights are not publicly available, or adaptation over API access.\loose

\iftoggle{arxiv2}{}{\vspace{-0.5em}}
\subsection{Efficient \algname with Noise Aliasing}\label{sec:noise_aliasing}
\iftoggle{arxiv2}{}{\vspace{-0.5em}}

While in principle \algname can be instantiated with virtually any RL algorithm, in this section we introduce an approach which makes particular use of the diffusion policy's structure to increase sample efficiency.
Our approach builds off standard actor-critic algorithms \cite{silver2014deterministic,lillicrap2015continuous,haarnoja2018soft} which, in our setting, would require fitting a critic, $Q : \cS \times \cW \rightarrow \R$, that quantifies the value of latent-noise actions, and an actor, $\pi : \cS \rightarrow \triangle_{\cW}$, which maps states to high value latent-noise actions. 
Such an approach can be applied in online learning---where we generate latent-noise actions $\bw$ ourself---but fails to apply in standard offline learning settings, where we are given a dataset of transitions $(\bs,\ba,r,\bs')$ labeled with actions in the \emph{original} action space, rather than the latent-noise action space, as the above actor-critic procedure requires. 
A standard application of an actor-critic algorithm does not, furthermore, fully exploit the structure of the diffusion policy. There may exist $\bw' \neq \bw$ such that $\pidpw(\bs,\bw) \approx \pidpw(\bs,\bw')$---indeed, we would expect a diffusion policy to exhibit this behavior in settings where the demonstrator has a relatively narrow action distribution, causing many $\bw$ to map to the same $\ba$. As we assume sampling access to $\pidpw$, we can in principle infer that $\pidpw(\bs,\bw) \approx \pidpw(\bs,\bw')$, which could allow us, for example, to determine the behavior of $\cMdp$ when taking action $\bw'$ without ever actually taking action $\bw'$, as illustrated in \Cref{fig:noise_aliasing_illustration}. Exploiting this aliasing behavior promises to reduce the amount of exploration over the latent-noise space required to learn in $\cMdp$, yet standard actor-critic algorithms fail to take advantage of this. To address these issues, we propose an instantiation of \algname, \emph{noise-aliased \algname}, which we state in \Cref{alg:main}.

\begin{figure}[H]
\vspace{-1.5em}
    \centering
    \begin{minipage}{\linewidth}
    \begin{algorithm}[H]
      \caption{Noise-Aliased Diffusion Steering via Reinforcement Learning (\algnamena)}
      \begin{algorithmic}[1]
        \State \textbf{input}: pretrained diffusion policy $\pidpw$, offline data $\frakD_{\mathrm{off}}$ and/or online environment $\cM$
        \State Initialize replay buffer $\frakB \leftarrow \frakD_{\mathrm{off}}$, $\cA$-critic $\Qa$, latent-noise critic $\Qw$, latent-noise actor $\piw$
        \For{$t = 1,\ldots, T$}
          \State Update $\Qa$: $\textstyle \min_{\Qa} \Exp_{(\bs,\ba,r,\bs') \sim \frakB, \ba' \sim \pidpw(\bs',\piw(\bs'))} \big [ ( \Qa(\bs,\ba) - r - \gamma \Qabar(\bs', \ba'))^2 \big ]$
     \State Update $\Qw$: $\min_{\Qw} \Exp_{\bs \sim \frakB, \bw \sim \cN(0,I)}[(\Qw(\bs, \bw) - \Qa(\bs,\pidpw(\bs,\bw)))^2]$
     \State Update $\piw$: $\max_{\piw} \Exp_{\bs \sim \frakB} \big [ \Qw(\bs,\piw(\bs)) \big ]$
        \If{access to online environment $\cM$}
          \State Sample latent-noise action $\bw_t \sim \piw(\bs_t)$ and compute $\ba_t \leftarrow \pidpw(\bs_t, \bw_t)$
          \State Play $\ba_t$ in $\cM$, observe $r_t$ and next state $\bs_{t+1}$, and add $(\bs_t, \ba_t, r_t, \bs_{t+1})$ to $\frakB$
        \EndIf
        \EndFor
      \end{algorithmic}
    \label{alg:main}
\end{algorithm}
\vspace{-2.5em}
\end{minipage}
\end{figure}

\algnamena maintains two critics, $\Qa : \cS \times \cA \rightarrow \R$ and $\Qw : \cS \times \cW \rightarrow \R$. 
$\Qa$ is trained via temporal-difference learning on the original action space, and is therefore able to incorporate standard offline data with $\cA$-actions. $\Qw$ operates on the latent-noise action space and distills $\Qa$ by sampling $\bw \sim \cN(0,I)$ and generating the corresponding $\cA$-action, $\ba \leftarrow \pidpw(\bs,\bw)$. This allows $\Qw$ to internalize the dynamics information encoded within $\Qa$ and, furthermore, map the value of latent-noise actions that have never been taken to the value of corresponding $\cA$-actions that have been taken, fully exploiting the aliasing nature of $\pidpw$.\loose

We remark that, in the purely offline setting, if $\pidp$ is trained on the same offline dataset $\frakD_{\mathrm{off}}$ as \algnamena, then \algnamena naturally handles conservatism. When using standard diffusion policy training, by construction $\pidpw$ will only output in-distribution actions, which implies that any $\bw$ will be mapped by $\pidpw$ to an in-distribution action. Since \algnamena only queries $\Qa$ on actions in $\frakD_{\mathrm{off}}$ and actions produced by $\pidpw$, it therefore naturally avoids querying the value of unseen actions, and achieves this while allowing $\piw$ to be optimized in an unrestricted manner over the latent-noise action space. 
Notably, this enables conservative policy optimization without explicitly enforcing a conservatism penalty, as is typically done in offline RL \cite{kumar2020conservative,kostrikov2021offline,jin2021pessimism,xie2021bellman}.

%% file: body/experiments_arxiv.tex

\newcommand{\dppo}{\textsc{Dppo}\xspace}
\newcommand{\idql}{\textsc{Idql}\xspace}
\newcommand{\dql}{\textsc{Dql}\xspace}
\newcommand{\dipo}{\textsc{Dipo}\xspace}
\newcommand{\qsm}{\textsc{Qsm}\xspace}
\newcommand{\sac}{\textsc{Sac}\xspace}
\newcommand{\rlpd}{\textsc{Rlpd}\xspace}
\newcommand{\algnamesac}{\textsc{Dsrl-Sac}\xspace}

\iftoggle{arxiv2}{}{\vspace{-0.75em}}
\section{Experiments}\label{sec:experiments}
\iftoggle{arxiv2}{}{\vspace{-0.75em}}

In our experiments, we evaluate \algname on simulated environments---in online (\Cref{sec:exp_online}), offline (\Cref{sec:exp_offline}), and offline-to-online (\Cref{sec:exp_offline_to_online}) settings---and real-world robotic control settings (\Cref{sec:exp_widowx}). We also demonstrate that \algname enables steering of generalist robot policies, in both simulation and the real world (\Cref{sec:exp_generalist}), and in \Cref{sec:understanding_ds} investigate the design decisions critical to \algname.
Throughout this section we instantiate \algname with \algnamena, as well as by applying \sac \cite{haarnoja2018soft} directly to the latent-action MDP $\cMdp$ (which we refer to as \algnamesac), and evaluate \algname's performance across both diffusion policies (with DDIM sampling) and flow-based policies.
Please see the appendix for further details on all experiments.

\iftoggle{arxiv2}{}{\vspace{-0.5em}}
\subsection{\algname Enables Efficient Online Adaptation of Diffusion Policies}\label{sec:exp_online}
\iftoggle{arxiv2}{}{\vspace{-0.5em}}

\begin{figure}
    \centering
    \begin{subfigure}[b]{0.11\textwidth}
        \centering
        \includegraphics[width=\textwidth]{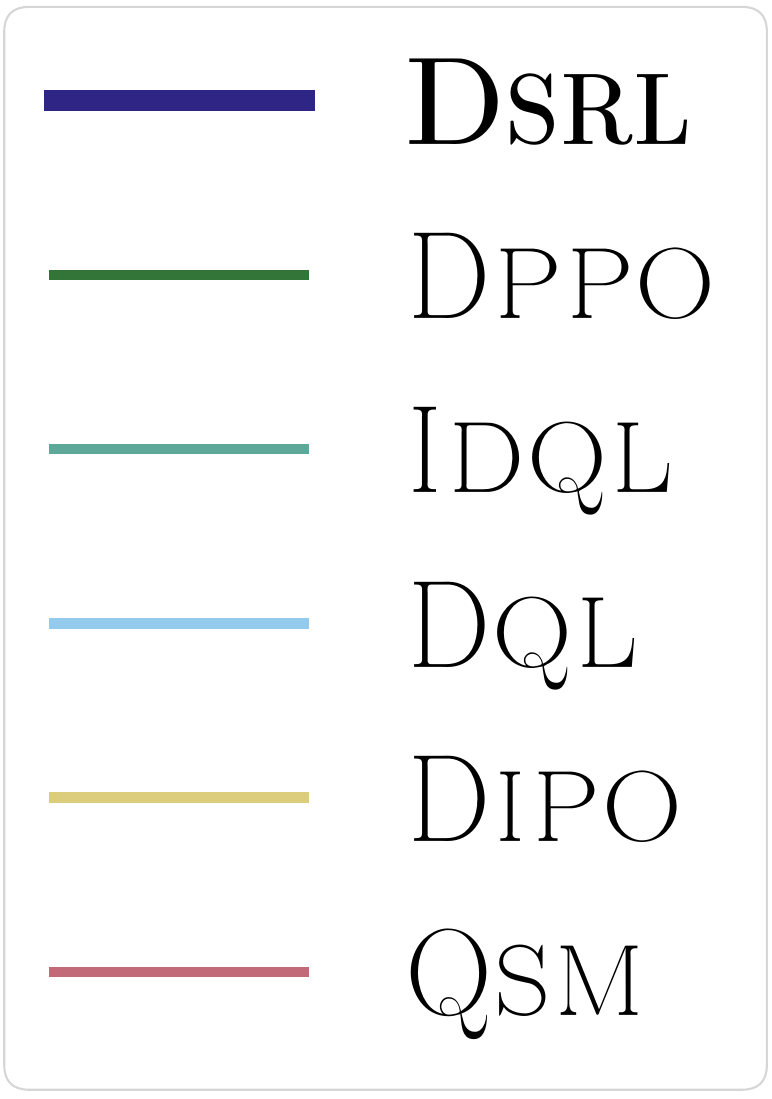}
        \vspace{0.05em}
    \end{subfigure}
    \hspace{0.5em}
    \begin{subfigure}[b]{0.24\textwidth}
        \centering
        \includegraphics[width=\textwidth]{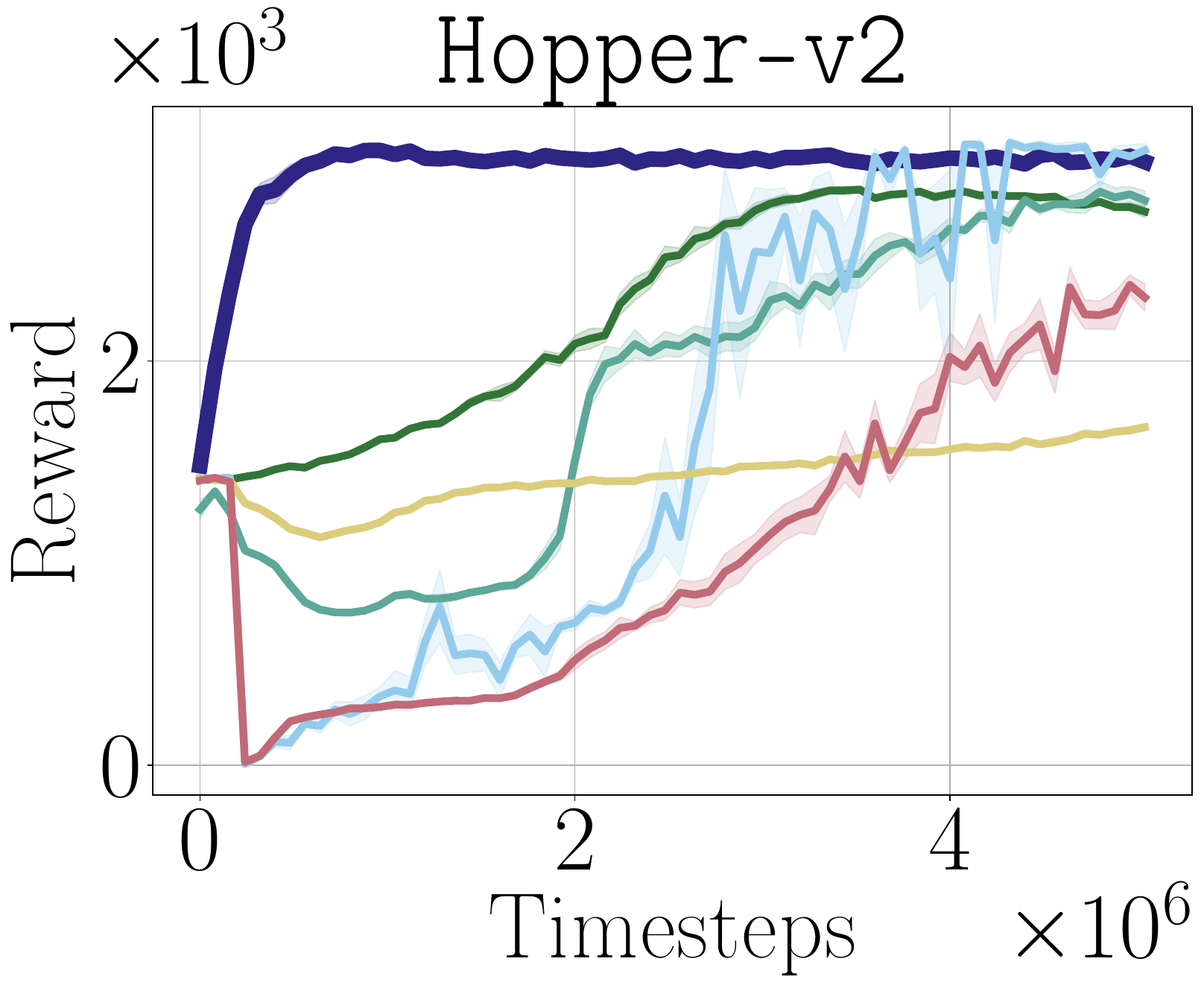}
    \end{subfigure}
    \hspace{0.5em}
    \begin{subfigure}[b]{0.22\textwidth}
        \centering
        \includegraphics[width=\textwidth]{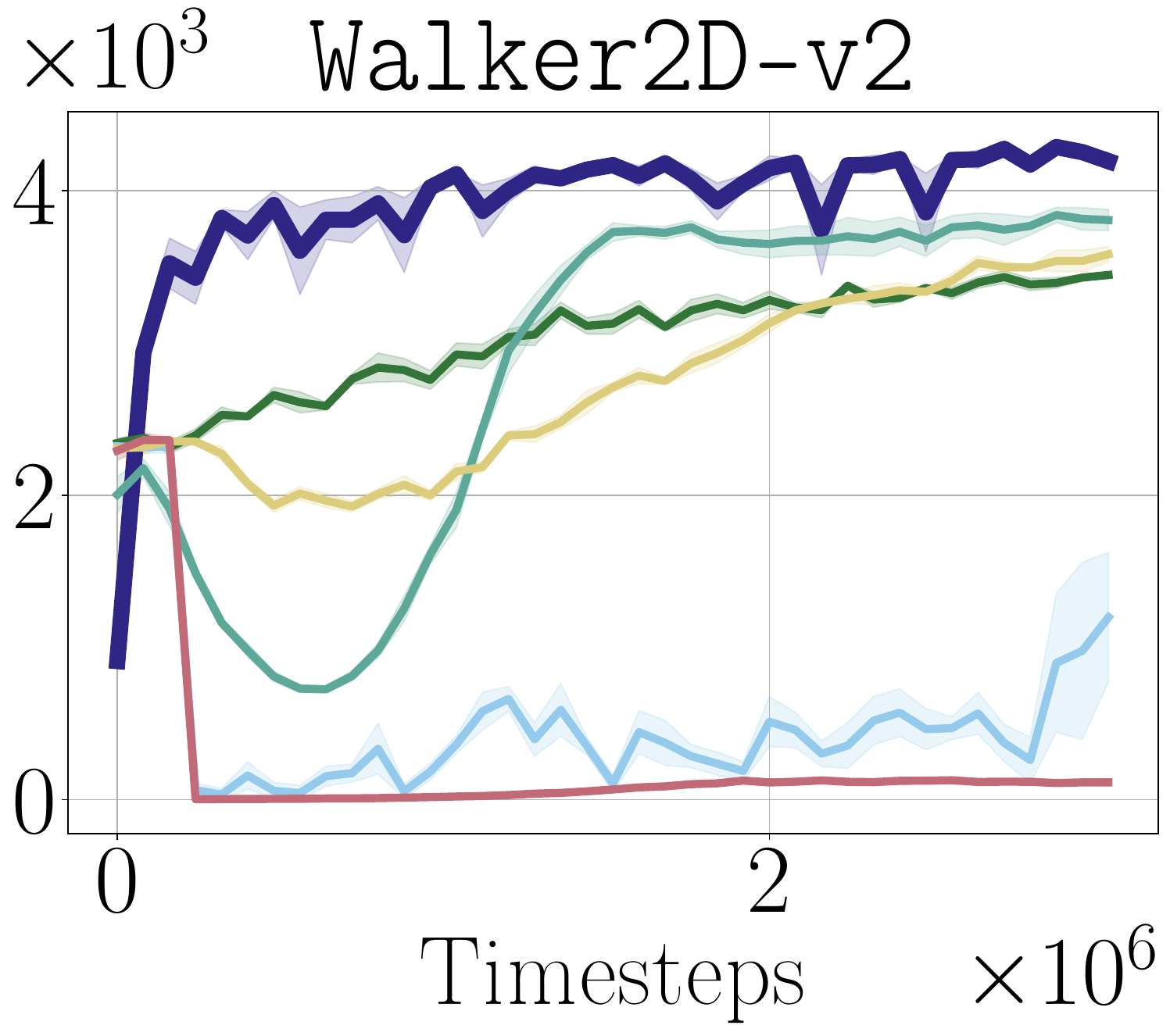}
    \end{subfigure}
    \hspace{0.5em}
    \begin{subfigure}[b]{0.23\textwidth}
        \centering
        \includegraphics[width=\textwidth]{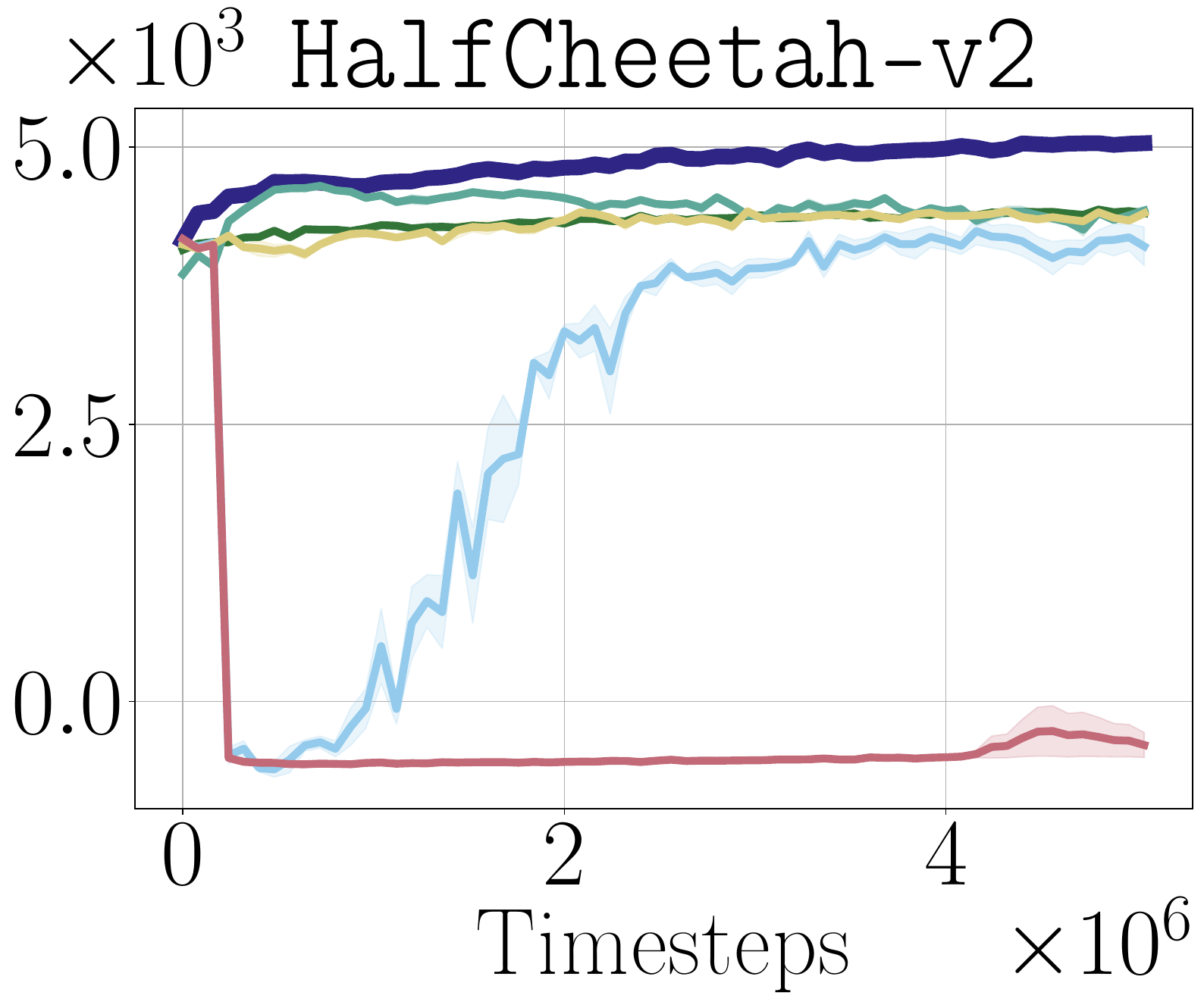}
    \end{subfigure}
    \caption{\algname enables online adaptation of pretrained diffusion policies on \texttt{OpenAI Gym} 
    \vspace{-0.5em}
    \cite{brockman2016openai}.}
    \label{fig:gym_results}
\end{figure}

\begin{figure}[t]
    \centering
    \begin{subfigure}[b]{0.245\textwidth}
        \centering
        \includegraphics[width=\textwidth]{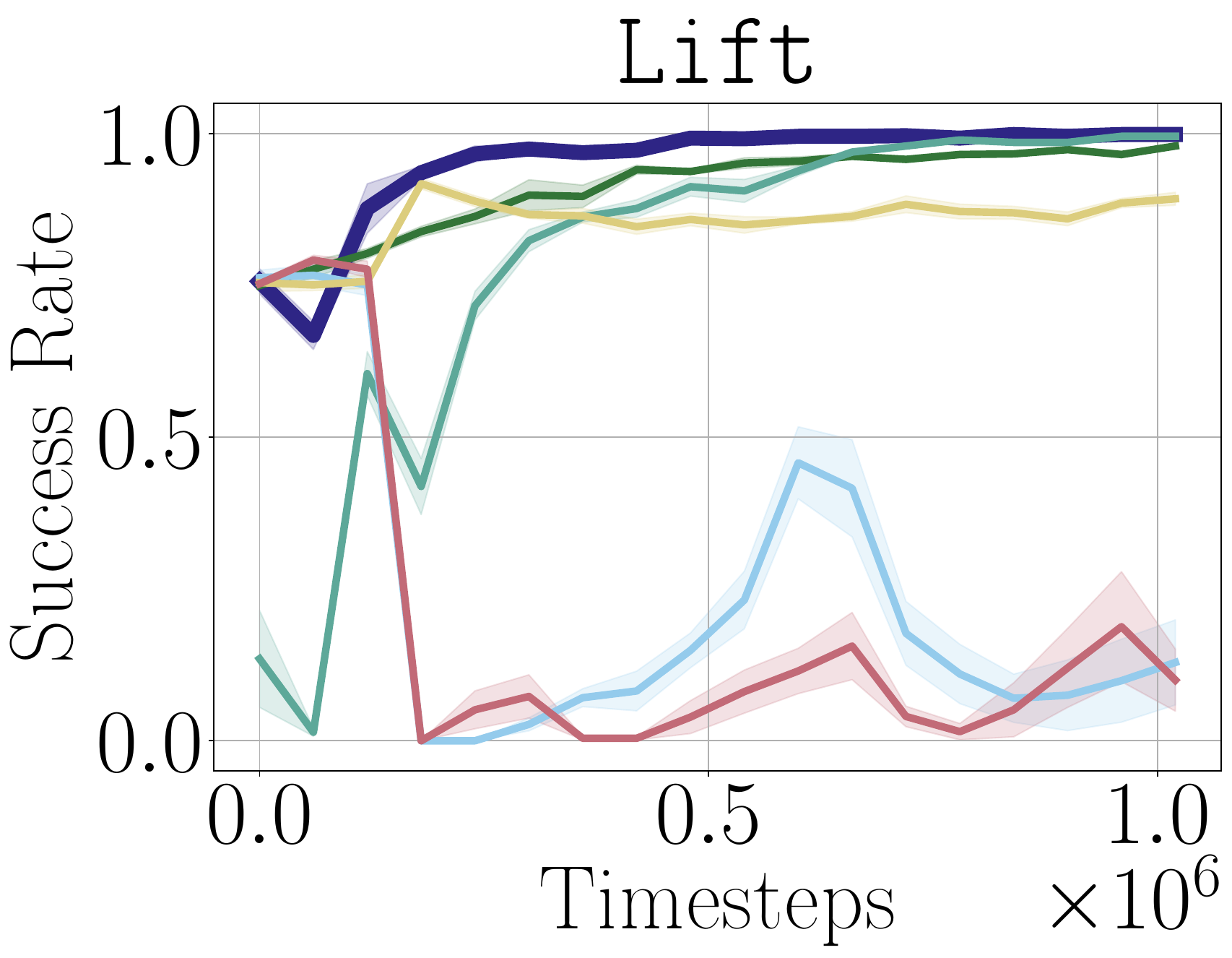}
    \end{subfigure}
    \hfill
    \begin{subfigure}[b]{0.23\textwidth}
        \centering
        \includegraphics[width=\textwidth]{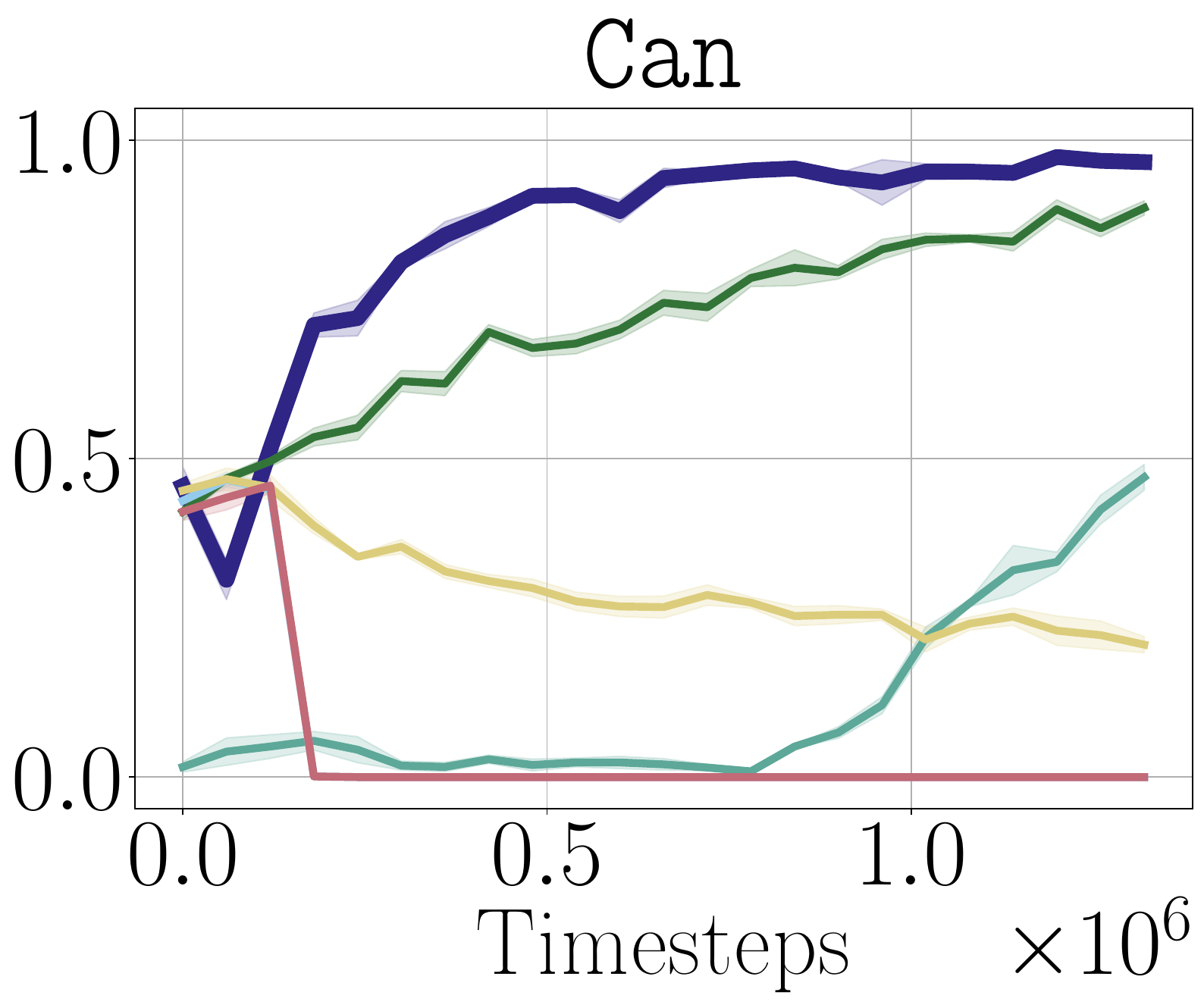}
    \end{subfigure}
    \hfill
    \begin{subfigure}[b]{0.23\textwidth}
        \centering
        \includegraphics[width=\textwidth]{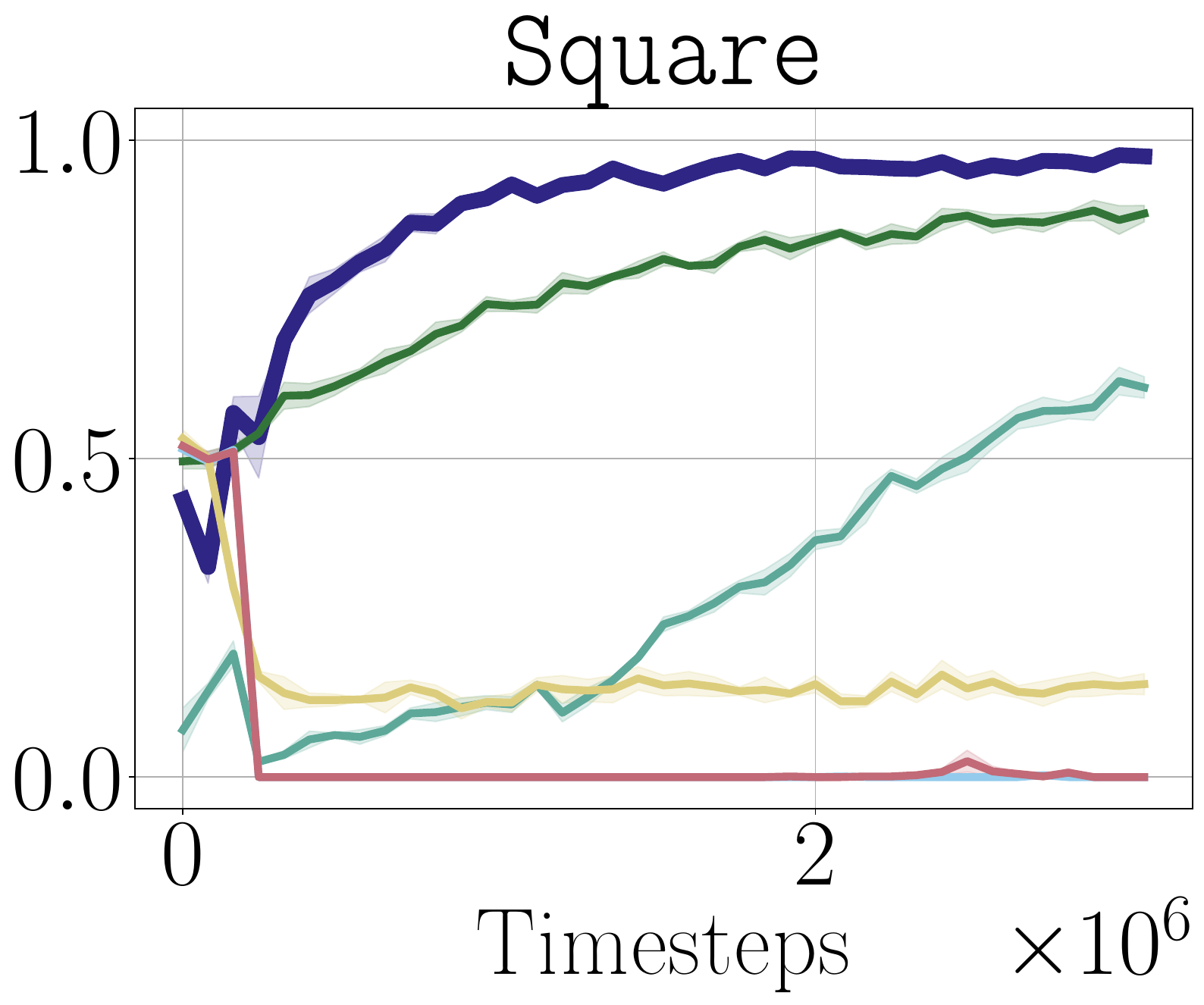}
    \end{subfigure}
    \hfill
    \begin{subfigure}[b]{0.23\textwidth}
        \centering
        \includegraphics[width=\textwidth]{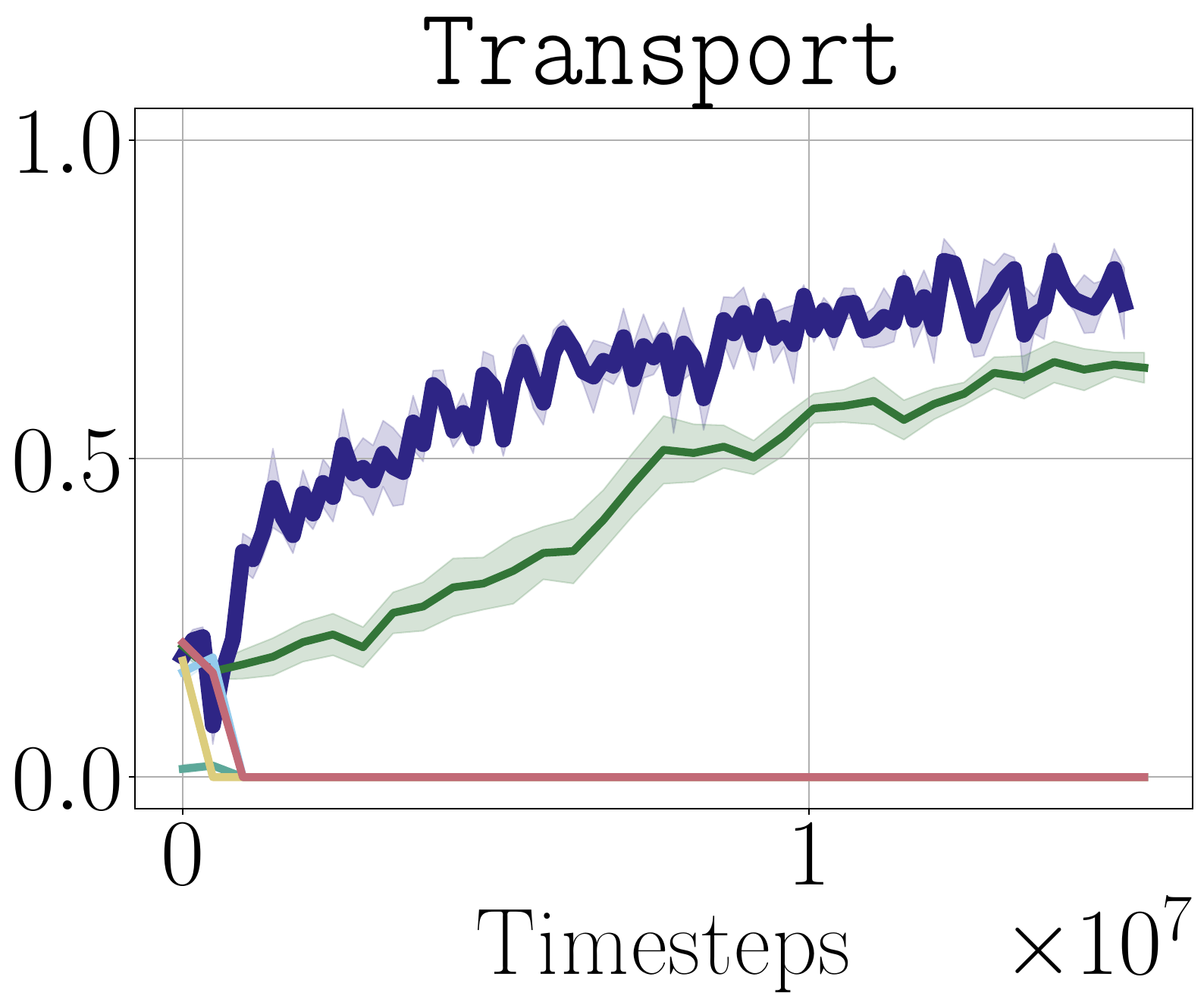}
    \end{subfigure}
    \caption{\algname enables online adaptation of pretrained diffusion policies on \texttt{Robomimic} \cite{robomimic2021}.}
    \label{fig:robomimic_results}
\end{figure}

We first study the online adaptation setting, where we assume we are given a diffusion policy $\pidp$ pretrained on demonstration data, and our goal is to adapt the behavior of $\pidp$ to maximize a given reward using only online samples. We evaluate on the \texttt{Robomimic} benchmark \cite{robomimic2021}, a challenging robotic manipulation benchmark, and three different tasks from the \texttt{OpenAI Gym} benchmark \cite{brockman2016openai}. 
\forapp{For \texttt{Robomimic}, $\pidp$ is pretrained with the demonstrations in the provided \texttt{Multi-Human} dataset, and for \texttt{Gym} environments, $\pidp$ is pretrained with the medium-level datasets from \texttt{D4RL} \cite{fu2020d4rl}.}
In all cases other than \texttt{Robomimic Square} and \texttt{Transport}, we set $\pidp$ to the diffusion policy checkpoints provided by \citet{ren2024diffusion}, while for \texttt{Square} and \texttt{Transport} we train a slightly higher-performing base diffusion policy. 
We compare against five other state-of-the-art methods for RL with diffusion policies, three of which---\dppo \cite{ren2024diffusion}, \idql \cite{hansen2023idql}, and \dql \cite{wang2022diffusion}---directly seek to adapt the behavior of $\pidp$ online, and two---\dipo \cite{yang2023policy} and \qsm \cite{psenka2023learning}---which utilize diffusion policies but learn from scratch (see \Cref{sec:online_exp_details} for additional details). 
Each experiment is averaged over 5 seeds, and all but \texttt{Transport} utilize \algnamena---for \texttt{Transport} we utilize \algnamesac.
Our results are given in \Cref{fig:robomimic_results,fig:gym_results}. As these figures illustrate, \algname is not only able to effectively modify the base policy's behavior, achieving near-optimal behavior on each task and demonstrating that diffusion steering is an expressive approach for policy adaptation, but also requires a substantially lower number of samples than existing approaches---at least a 5-10$\times$ improvement in efficiency---to reach this level of performance.

\iftoggle{arxiv2}{}{\vspace{-0.5em}}
\subsection{\algname Enables Efficient Offline Adaptation of Diffusion Policies}\label{sec:exp_offline}
\iftoggle{arxiv2}{}{\vspace{-0.5em}}

We next turn to the offline setting, evaluating how well the \algnamena variant of \algname can learn from offline data. We evaluate on 10 tasks from the \texttt{OGBench} offline RL benchmark \cite{park2024ogbench}, covering a variety of locomotion and manipulation settings. In all cases we utilize the provided offline dataset to train $\pidp$, then freeze $\pidp$ and apply \algnamena on this same offline dataset. 
We compare against 10 different baselines: Gaussian-policy based \textsc{Bc} (``\textsc{Bc ($\cN$)}''), \textsc{Iql} \cite{kostrikov2021offline}, and \textsc{Rebrac} \cite{tarasov2023revisiting}, diffusion-policy based \textsc{Idql} \cite{hansen2023idql}, \textsc{Srpo} \cite{chen2023score}, and \textsc{Cac} \cite{ding2023consistency}, and flow-policy based \textsc{Fawac}, \textsc{Fbrac}, \textsc{Ifql} (flow-based variants of \cite{nair2020awac}, \cite{wang2022diffusion}, and \textsc{Idql}, respectively),
\textsc{Fql} \cite{park2025flow}, and flow-based \textsc{Bc} (the base diffusion policy used by \algname, ``\textsc{Bc ($\pidp$)}'').
These baselines cover a wide range of standard offline RL methods, including methods that utilize diffusion or flow-based policies (e.g., \textsc{Fql}).\loose

\begin{table*}[t!]
\centering
\scalebox{0.55}
{
\begin{tabular}{lcccccccccccc}
\toprule
\texttt{Task} & \textsc{Bc ($\cN$)} & \textsc{Bc} ($\pidp$) & \textsc{Iql} & \textsc{Rebrac} & \textsc{Idql} & \textsc{Srpo} & \textsc{Cac} & \textsc{Fawac} & \textsc{Fbrac} & \textsc{Ifql} & \textsc{Fql} & \textbf{\algname} \\
\midrule

\texttt{antmaze-large-navigate-singletask} & $0$ {\tiny $\pm 0$} & $0$ {\tiny $\pm 0$} & $48$ {\tiny $\pm 9$} & $\mathbf{91}$ {\tiny $\pm 10$} & $0$ {\tiny $\pm 0$} & $0$ {\tiny $\pm 0$} & $42$ {\tiny $\pm 7$} & $1$ {\tiny $\pm 1$} & $70$ {\tiny $\pm 20$} & $24$ {\tiny $\pm 17$} & $80$ {\tiny $\pm 8$}  & $40$ {\tiny $\pm 29$} \\
\texttt{antmaze-giant-navigate-singletask} & $0$ {\tiny $\pm 0$} & $0$ {\tiny $\pm 0$} & $0$ {\tiny $\pm 0$} & $\mathbf{27}$ {\tiny $\pm 22$} & $0$ {\tiny $\pm 0$} & $0$ {\tiny $\pm 0$} & $0$ {\tiny $\pm 0$} & $0$ {\tiny $\pm 0$} & $0$ {\tiny $\pm 1$} & $0$ {\tiny $\pm 0$} & $4$ {\tiny $\pm 5$}  & $0$ {\tiny $\pm 0$} \\
\texttt{humanoidmaze-medium-navigate-singletask} & $1$ {\tiny $\pm 0$} & $0$ {\tiny $\pm 0$} & $32$ {\tiny $\pm 7$} & $16$ {\tiny $\pm 9$} & $1$ {\tiny $\pm 1$} & $0$ {\tiny $\pm 0$} & $38$ {\tiny $\pm 19$} & $6$ {\tiny $\pm 2$} & $25$ {\tiny $\pm 8$} & $\mathbf{69}$ {\tiny $\pm 19$} & $19$ {\tiny $\pm 12$}  & $34$ {\tiny $\pm 20$} \\
\texttt{humanoidmaze-large-navigate-singletask} & $0$ {\tiny $\pm 0$} & $0$ {\tiny $\pm 0$} & $3$ {\tiny $\pm 1$} & $2$ {\tiny $\pm 1$} & $0$ {\tiny $\pm 0$} & $0$ {\tiny $\pm 0$} & $1$ {\tiny $\pm 1$} & $0$ {\tiny $\pm 0$} & $0$ {\tiny $\pm 1$} & $6$ {\tiny $\pm 2$} & $7$ {\tiny $\pm 6$}  & $\mathbf{10}$ {\tiny $\pm 12$} \\
\texttt{antsoccer-arena-navigate-singletask} & $1$ {\tiny $\pm 0$} & $0$ {\tiny $\pm 0$} & $3$ {\tiny $\pm 2$} & $0$ {\tiny $\pm 0$} & $0$ {\tiny $\pm 1$} & $0$ {\tiny $\pm 0$} & $0$ {\tiny $\pm 0$} & $12$ {\tiny $\pm 3$} & $24$ {\tiny $\pm 4$} & $16$ {\tiny $\pm 9$} & $\mathbf{39}$ {\tiny $\pm 6$}  & $28$ {\tiny $\pm 9$} \\
\texttt{cube-single-play-singletask} & $3$ {\tiny $\pm 1$} & $3$ {\tiny $\pm 3$} & $85$ {\tiny $\pm 8$} & $92$ {\tiny $\pm 4$} & $\mathbf{96}$ {\tiny $\pm 2$} & $82$ {\tiny $\pm 16$} & $80$ {\tiny $\pm 30$} & $81$ {\tiny $\pm 9$} & $83$ {\tiny $\pm 13$} & $73$ {\tiny $\pm 3$} & $\mathbf{97}$ {\tiny $\pm 2$} & $\mathbf{93}$ {\tiny $\pm 14$} \\
\texttt{cube-double-play-singletask} & $0$ {\tiny $\pm 0$} & $0$ {\tiny $\pm 0$} & $1$ {\tiny $\pm 1$} & $7$ {\tiny $\pm 3$} & $16$ {\tiny $\pm 10$} & $0$ {\tiny $\pm 0$} & $2$ {\tiny $\pm 2$} & $2$ {\tiny $\pm 1$} & $22$ {\tiny $\pm 12$} & $9$ {\tiny $\pm 5$} & $36$ {\tiny $\pm 6$} &  $\mathbf{53}$ {\tiny $\pm 14$} \\
\texttt{scene-play-singletask} & $1$ {\tiny $\pm 1$} & $0$ {\tiny $\pm 0$} & $12$ {\tiny $\pm 3$} & $50$ {\tiny $\pm 13$} & $33$ {\tiny $\pm 14$} & $2$ {\tiny $\pm 2$} & $50$ {\tiny $\pm 40$} & $18$ {\tiny $\pm 8$} & $46$ {\tiny $\pm 10$} & $0$ {\tiny $\pm 0$} & $76$ {\tiny $\pm 9$} &  $\mathbf{88}$ {\tiny $\pm 9$} \\
\texttt{puzzle-3x3-play-singletask} & $1$ {\tiny $\pm 1$} & $0$ {\tiny $\pm 0$} & $2$ {\tiny $\pm 1$} & $2$ {\tiny $\pm 1$} & $0$ {\tiny $\pm 0$} & $0$ {\tiny $\pm 0$} & $0$ {\tiny $\pm 0$} & $1$ {\tiny $\pm 1$} & $2$ {\tiny $\pm 2$} & $0$ {\tiny $\pm 0$} & $\mathbf{16}$ {\tiny $\pm 5$}  & $0$ {\tiny $\pm 0$} \\
\texttt{puzzle-4x4-play-singletask} & $0$ {\tiny $\pm 0$} & $0$ {\tiny $\pm 0$} & $5$ {\tiny $\pm 2$} & $10$ {\tiny $\pm 3$} & $26$ {\tiny $\pm 6$} & $7$ {\tiny $\pm 4$} & $1$ {\tiny $\pm 1$} & $0$ {\tiny $\pm 0$} & $5$ {\tiny $\pm 1$} & $21$ {\tiny $\pm 11$} & $11$ {\tiny $\pm 3$}  & $\mathbf{37}$ {\tiny $\pm 13$} \\

\bottomrule
\end{tabular}
}
\caption{
\algname enables effective adaptation of flow policies from offline data on \texttt{OGBench} \cite{park2024ogbench}.
}
\vspace{-1em}
\label{table:offline}
\end{table*}

Our results are given in \Cref{table:offline}. All baseline values, with the exception of \textsc{Bc ($\pidp$)}, are taken from \citet{park2025flow}, and \algname and \textsc{Bc ($\pidp$)} are averaged over 4 random seeds.
We see that \algname achieves state-of-the-art performance on approximately half the tasks, demonstrating its effectiveness at learning from offline data.
We highlight in particular that \algname substantially improves over \textsc{Bc ($\pidp$)}---the performance of $\pidp$ before steering---again illustrating the large increase in policy performance diffusion steering can produce, even when utilizing offline data .
It is notable that the same \algnamena method that we use in our online RL experiments directly serves as a highly effective offline RL method without modification.\loose

\subsection{\algname Enables Efficient Offline-to-Online Adaptation of Diffusion Policies}\label{sec:exp_offline_to_online}
The results from \Cref{sec:exp_online,sec:exp_offline} demonstrate that \algname can effectively steer diffusion policies using both online interactions and offline data. Here we investigate whether \algname is an effective algorithm for offline-to-online settings. We consider again the \texttt{Robomimic Can} 
\begin{wrapfigure}{r}{0.6\textwidth}
    \centering
    \begin{subfigure}[b]{0.12\textwidth}
        \centering
        \includegraphics[width=\textwidth]{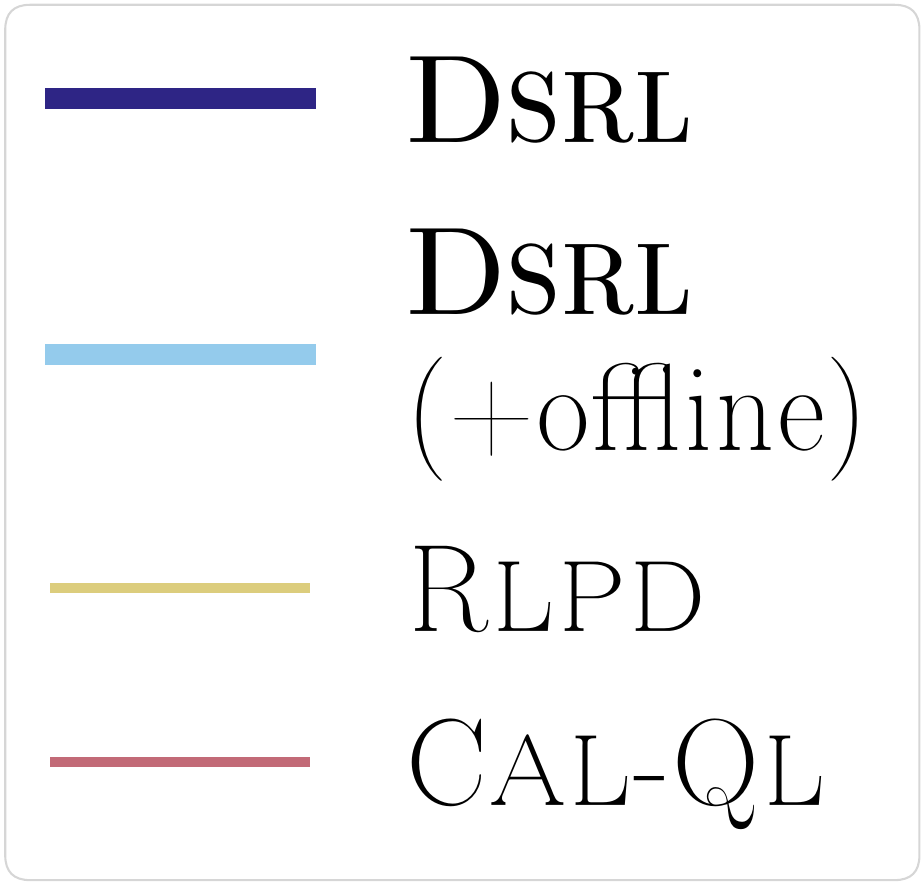}
        \vspace{0.35em}
    \end{subfigure}
    \hfill
    \begin{subfigure}[b]{0.23\textwidth}
        \centering
        \includegraphics[width=\textwidth]{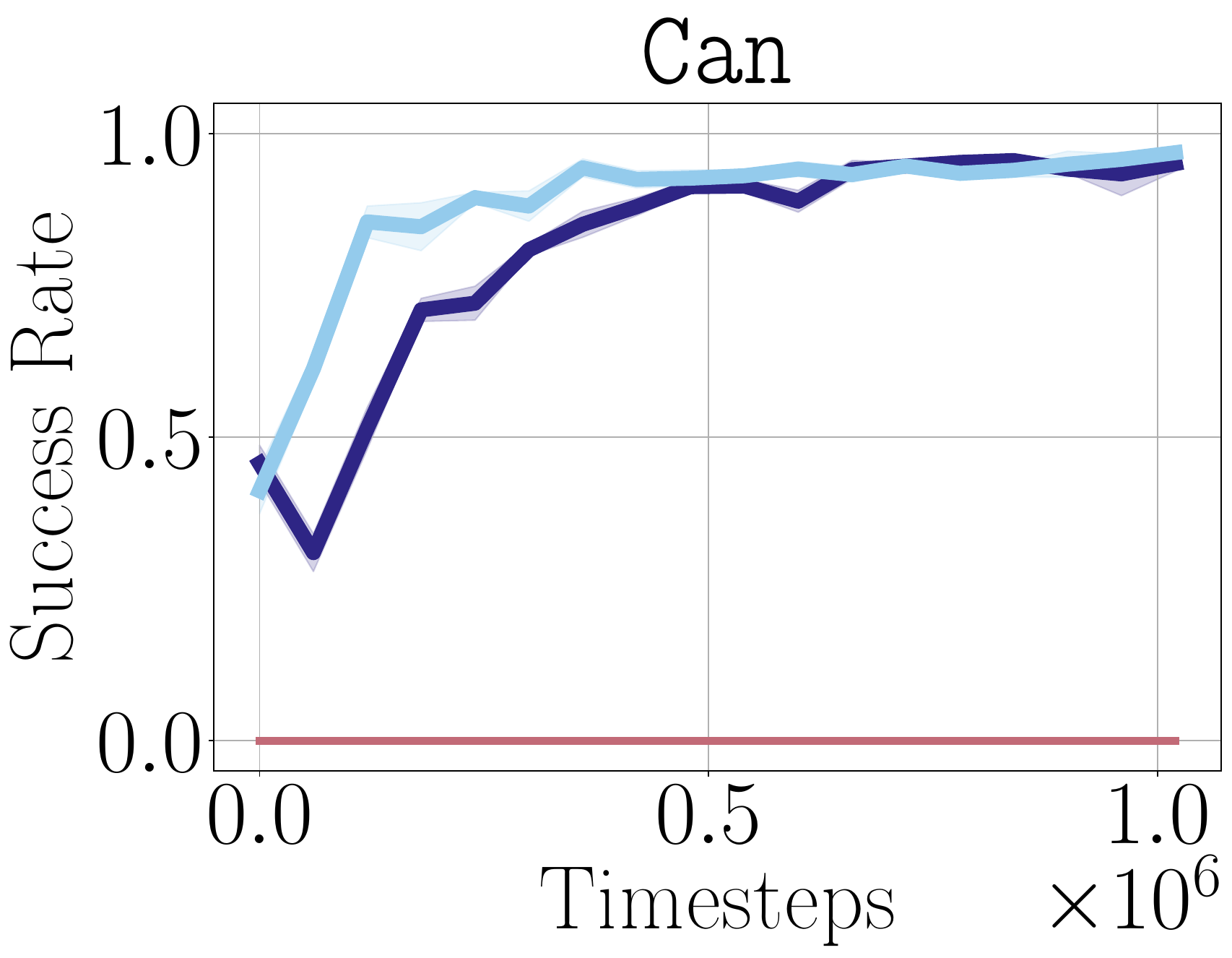}
    \end{subfigure}
    \hfill
    \begin{subfigure}[b]{0.215\textwidth}
        \centering
        \includegraphics[width=\textwidth]{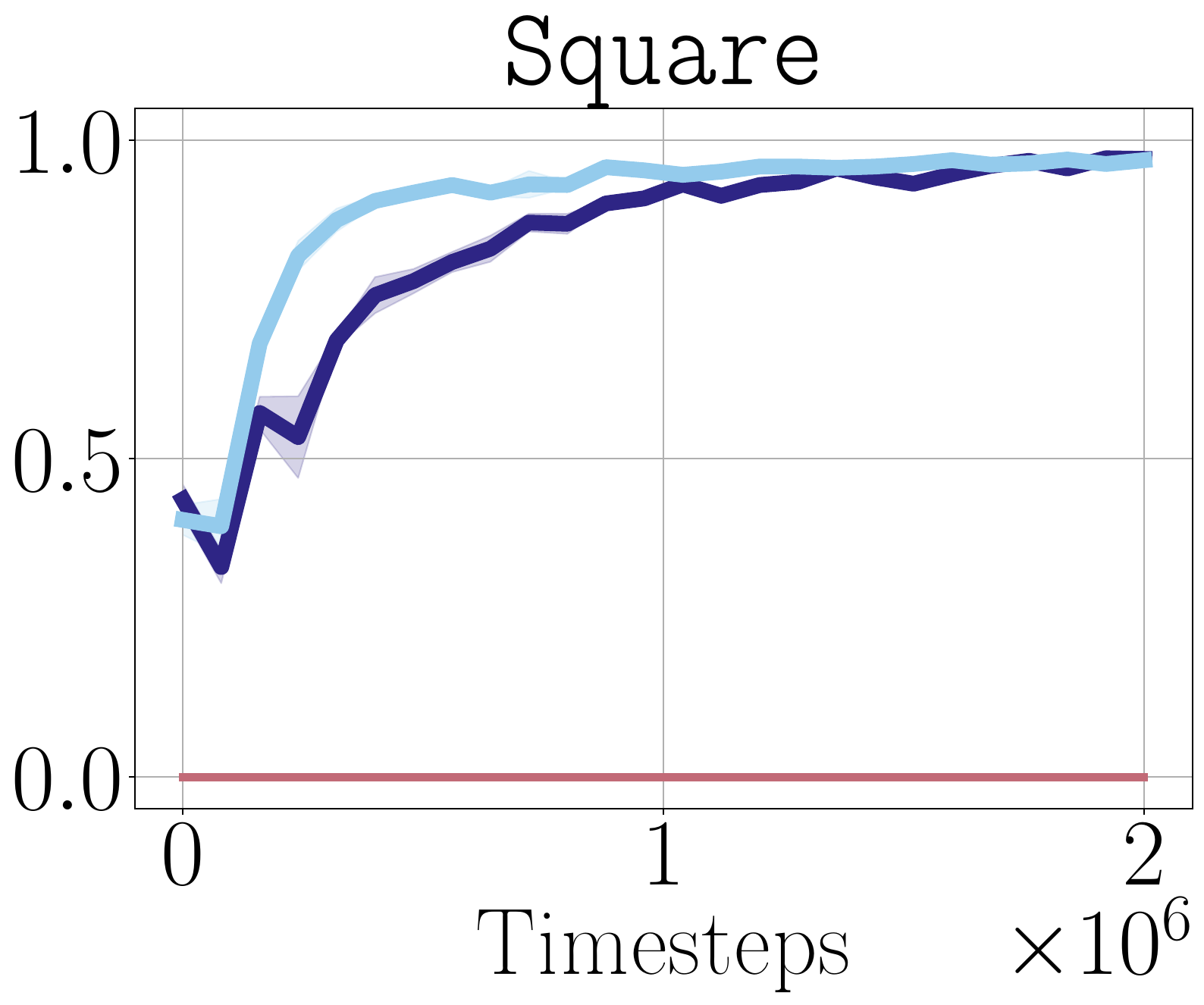}
    \end{subfigure}
    \vspace{-0.5em}
    \caption{\algname can make effective use of offline data to speed up online learning.}
    \label{fig:offline_to_online_results}
    \vspace{-0.5em}
\end{wrapfigure}
and \texttt{Square} tasks of \Cref{sec:exp_online},
 and as baselines compare against two standard approaches for offline-to-online RL, \textsc{Rlpd} \cite{ball2023efficient} and \textsc{Cal-QL} \cite{nakamoto2023cal}. We run \algname exactly as described in \Cref{alg:main}, with $\frakD_{\mathrm{off}}$ set to the \texttt{Robomimic} Multi-Human demonstration datasets. Our results are given in \Cref{fig:offline_to_online_results}. As can be seen, including the offline data improves the sample complexity of \algname by a factor of approximately 2$\times$, demonstrating that \algname can indeed effectively leverage offline data to speed up learning. In contrast, standard offline-to-online RL approaches are not able to learn at all on these tasks.

\iftoggle{arxiv2}{}{\vspace{-0.5em}}
\subsection{Adapting Real-World Single- and Multi-Task Robot Policies with \algname}\label{sec:exp_widowx}
\iftoggle{arxiv2}{}{\vspace{-0.5em}}

In this section, we demonstrate that \algname enables real-world improvement of robot control policies. We evaluate the performance of \algname in two settings: steering a single-task diffusion policy trained on a narrow set of expert demonstrations, and steering a multi-task diffusion policy trained on a much more diverse set of demonstrations.
For all experiments, we first roll out the pretrained diffusion policy $\pidp$ $n$ times ($n= 10$ for single-task experiments, and $20$ for multi-task experiments), then begin the online RL training, utilizing \algnamesac and initializing the replay buffer with the data from these rollouts of $\pidp$.  As a baseline, we run \rlpd \cite{ball2023efficient}, which has been shown to achieve excellent performance on real-world robot learning problems \cite{luo2024serl,luo2024precise}. In all case we use a 0-1 sparse reward.\loose

\begin{figure}[t]
\vspace{-0.5em}
\centering
\begin{subfigure}[b]{0.13\textwidth}
        \centering
         \includegraphics[width=\textwidth]{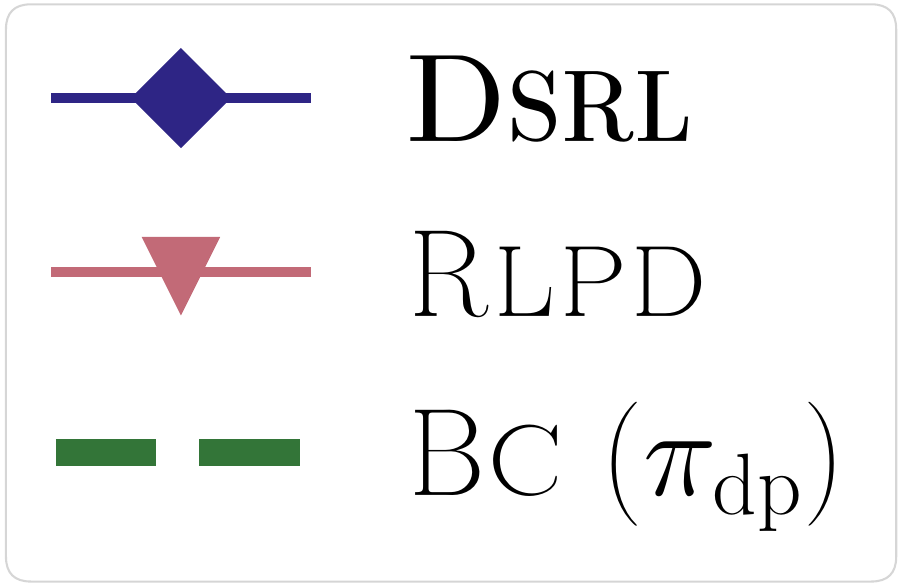}
        \vspace{1.1em}
    \end{subfigure}
    \hspace{0.5em}
    \begin{subfigure}[b]{0.2475\textwidth}
        \centering
        \includegraphics[width=\textwidth]{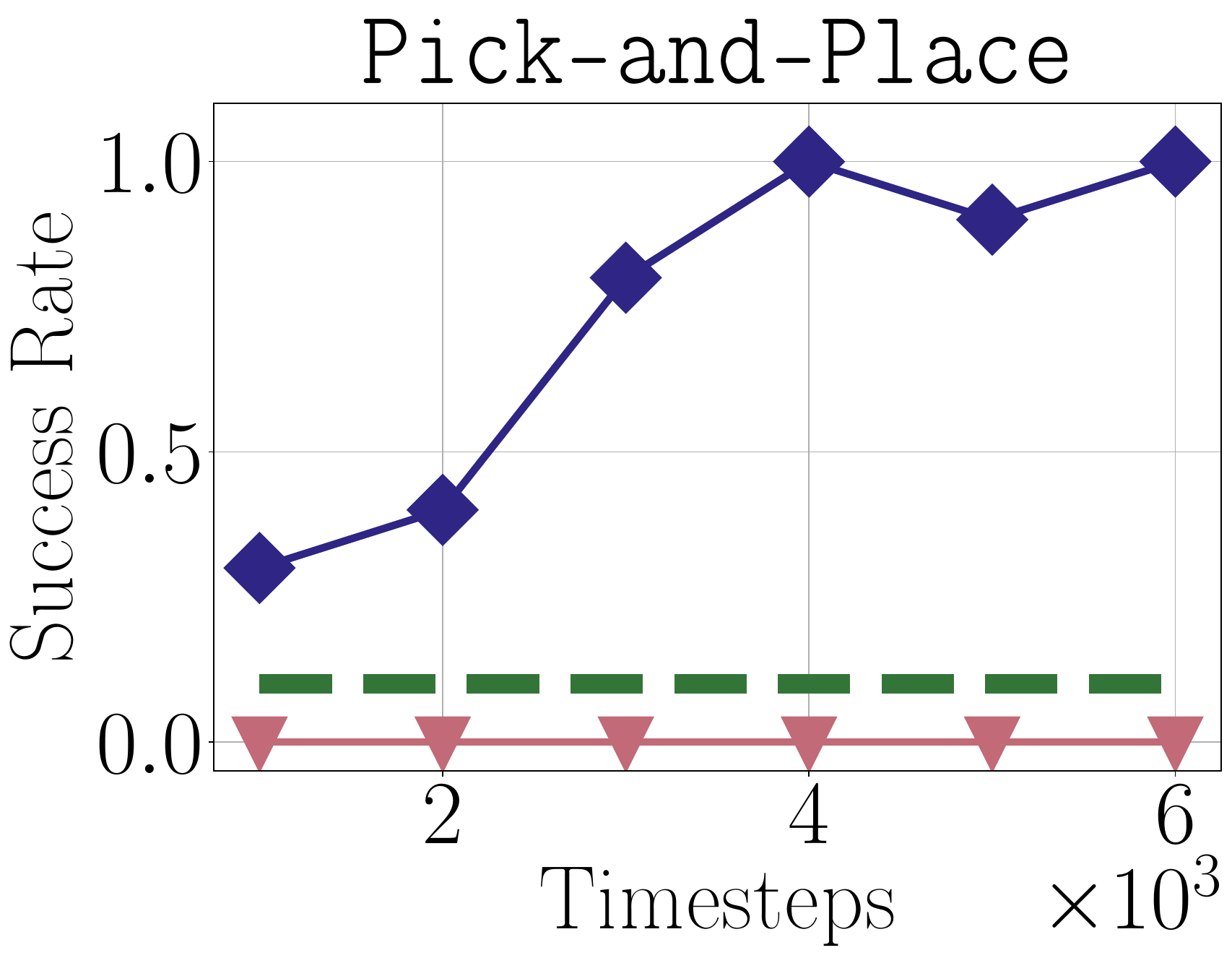}
    \end{subfigure}
    \hspace{0.5em}
    \begin{subfigure}[b]{0.23\textwidth}
        \centering
         \includegraphics[width=\textwidth]{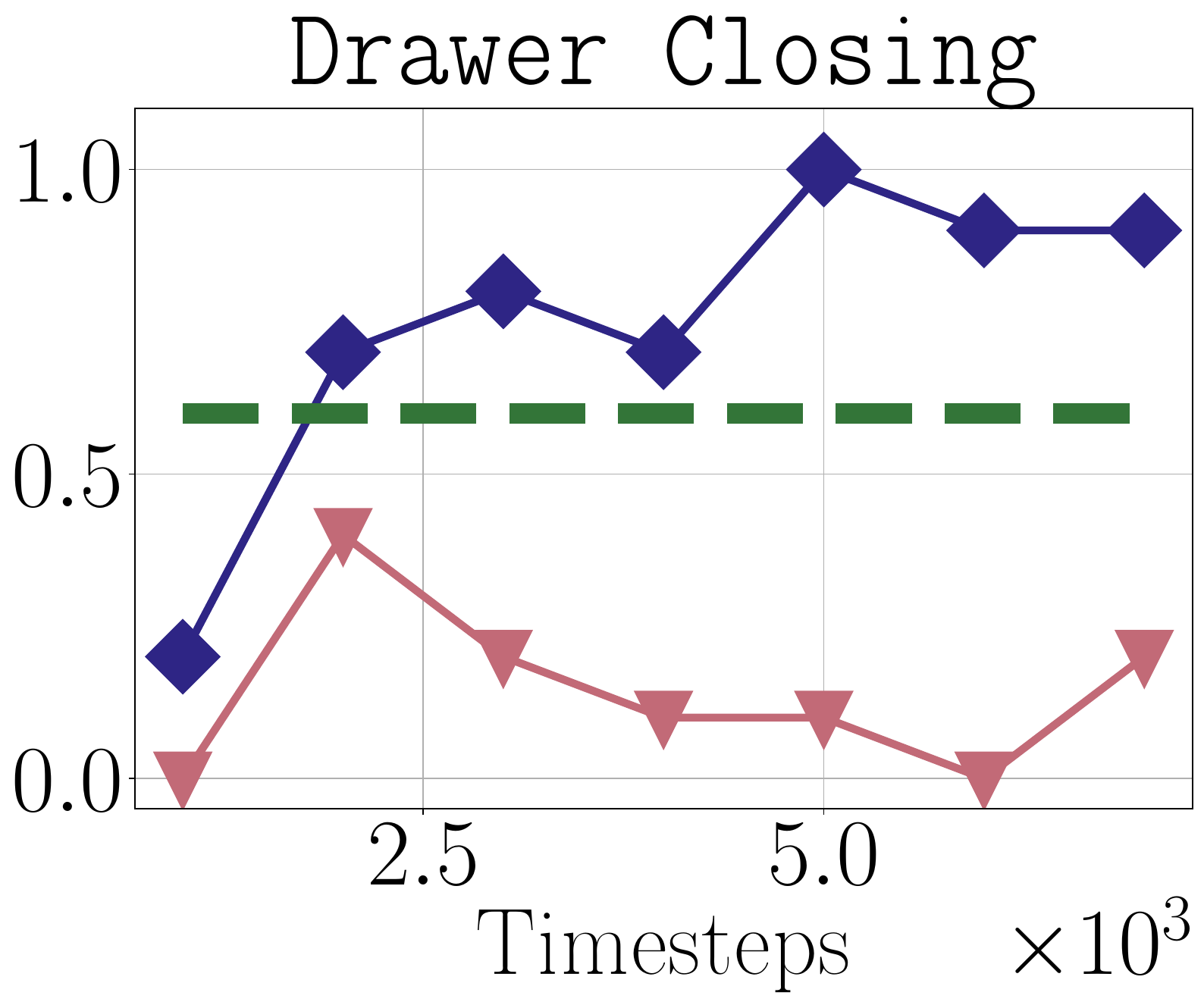}
    \end{subfigure}
    \hspace{0.5em}
     \begin{subfigure}[b]{0.23\textwidth}
        \centering
         \includegraphics[width=\textwidth]{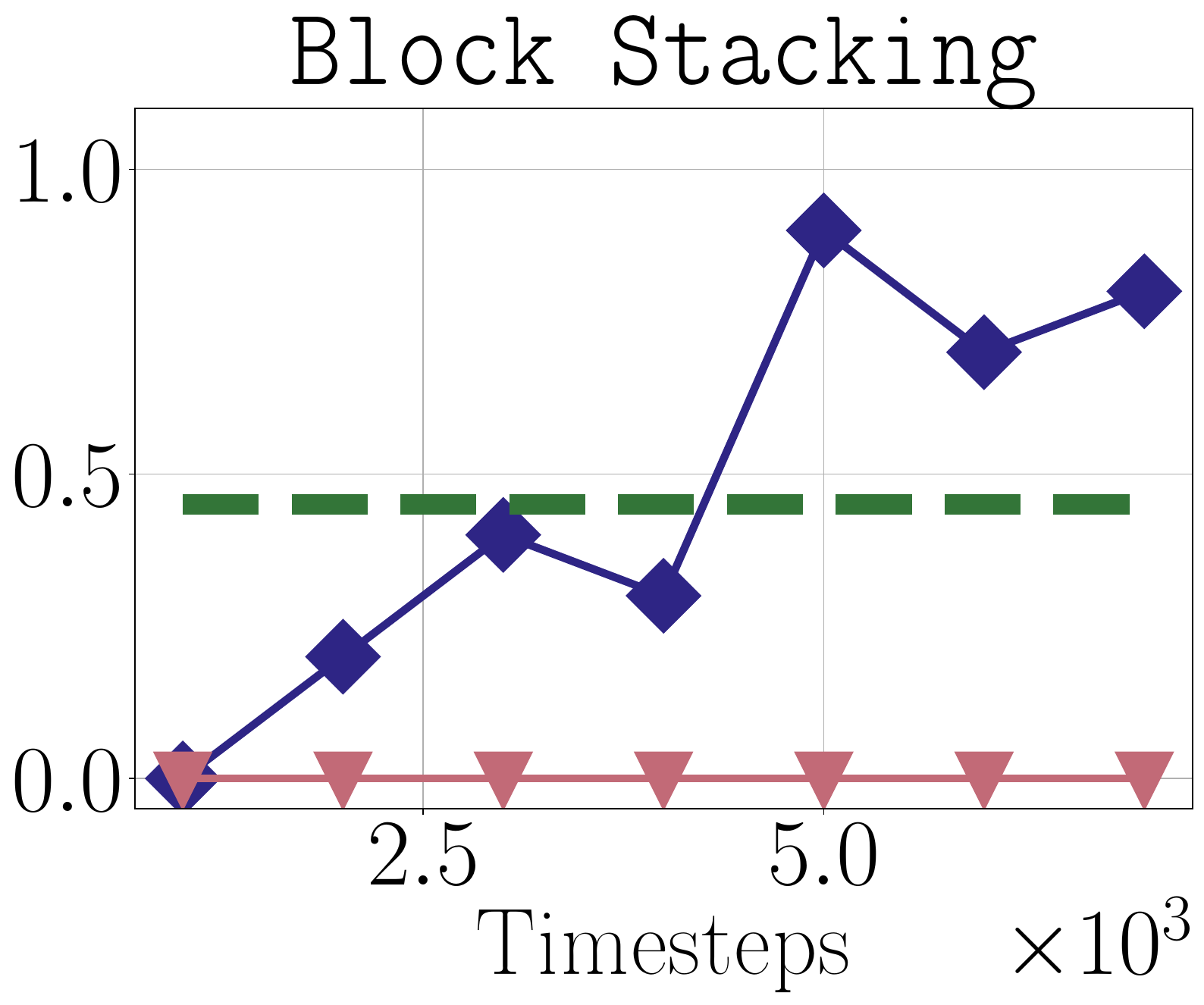}
    \end{subfigure}
    \caption{\algname efficiently improves performance of multi-task BC policy trained on Bridge V2 dataset on real-world WidowX robot.}
    \label{fig:robot_tasks}
\vspace{-1em}
\end{figure}

\begin{wraptable}{r}{0.45\textwidth}
\centering
\vspace{-1em}
\scalebox{1}
{
\begin{tabular}{lcccccccccccc}
\toprule
$\pidp$ & \rlpd & \parbox{1.75cm}{\rlpd +\\ interventions} & \algname \\
\midrule
$2/10$ & $0/10$ & $0/10$ & $\mathbf{9/10}$ \\
\bottomrule
\end{tabular}
}
\caption{
Performance of single-task diffusion policy $\pidp$, \rlpd, and \algname steering of $\pidp$.
}
\label{table:franka_real}
\end{wraptable}
\paragraph{\algname steering of single-task diffusion policy.}
For our single-task experiment, we consider a pick-and-place task, where the objective is to pick up a cube and place it in a bowl (see \Cref{fig:robot_tasks}(a)). We run on a Franka Emika Panda robot arm \cite{haddadin2022franka}, and train the base policy on 10 human demonstrations. For the \rlpd baseline, we utilize the human demonstrations as the offline dataset to initialize the algorithm. We also consider an additional variant of \rlpd, where a human periodically intervenes to correct \rlpd's actions, similar to the procedure described in \cite{luo2024precise}. We run all methods for 3500 online steps (approximately 40 episodes).\loose

\begin{figure}[t]
\centering
\includegraphics[width=0.9\textwidth]{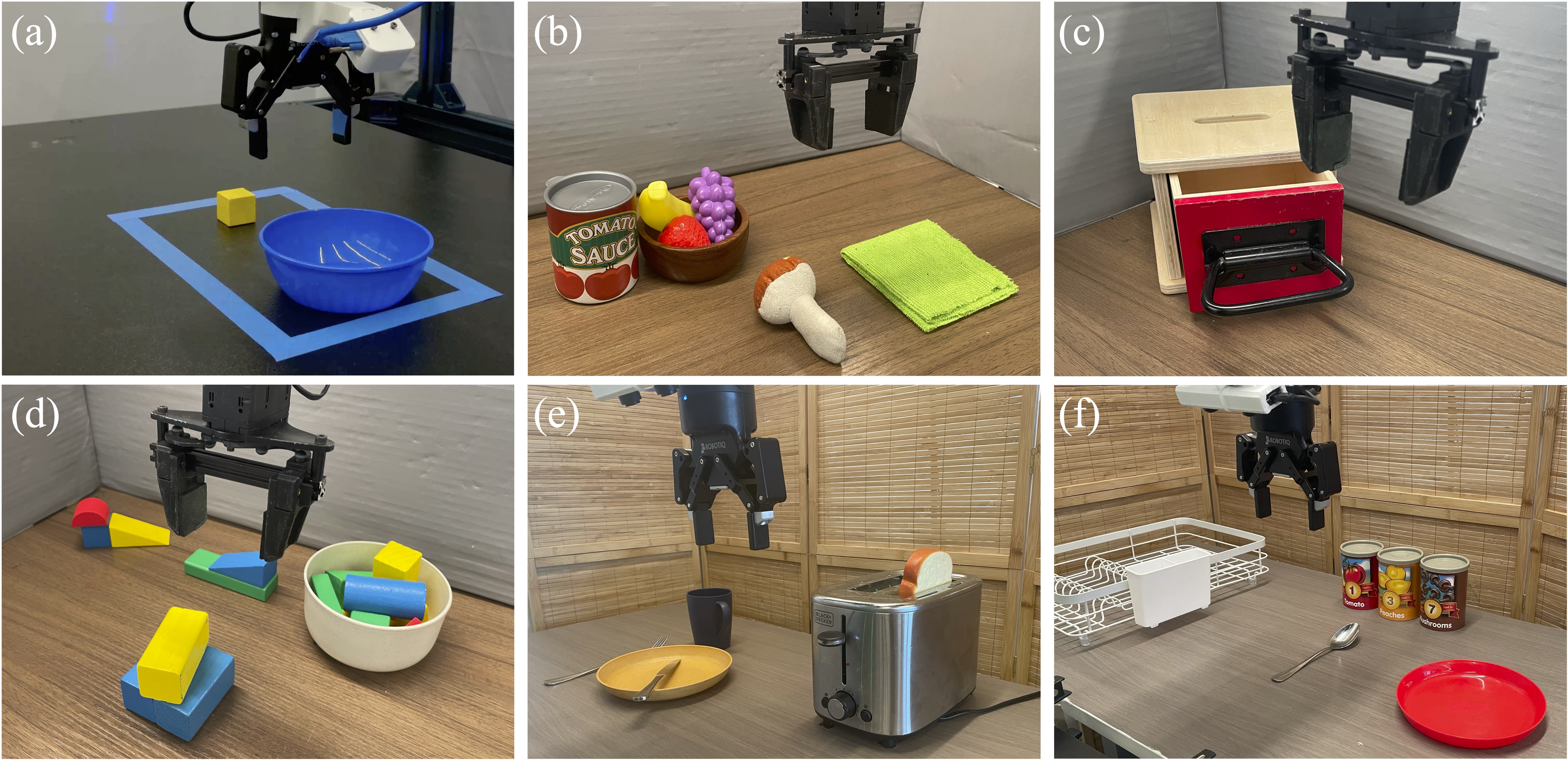}
\caption{Real-world robot tasks for \algname adaptation.}
\label{fig:widowx_results}
\end{figure}

Our results are given in \Cref{table:franka_real}, where we see that \algname is able to quickly improve the performance of $\pidp$ from 20\% to nearly 100\%. In contrast, \rlpd is unable to learn at all. We note that while augmenting \rlpd with human interventions does lead to more effective behavior than standard \rlpd, it is still unable to successfully solve the task.
This not only shows that \algname is sample-efficient enough to make adapting real-world diffusion policies practical,
but also shows that, given some demonstration data, rather than using this data to initialize \rlpd, a much more effective approach is to first train a diffusion policy on this data then run \algname on this diffusion policy.

\paragraph{\algname on multi-task diffusion policy.}
Next, we study if \algname can adapt multi-task real-world diffusion policies. We first train a diffusion policy on the Bridge V2 dataset \cite{walke2023bridgedata}. The Bridge V2 dataset contains over 60,000 robotic teleoperation trajectories collected on a diverse set of manipulation tasks on the WidowX 250 6-DoF robot arm. Pretraining on Bridge endows $\pidp$ with a variety of useful manipulation behaviors, yet such pretrained policies often still struggle when deployed in novel scenes. Here we parameterize $\pidp$ with the DiT policy architecture proposed by \citet{dasari2024ingredients}---a state-of-the-art transformer-based diffusion policy.

We evaluate on three tasks---\texttt{Pick-and-Place}, \texttt{Drawer Closing}, and \texttt{Block Stacking}---as illustrated in \Cref{fig:robot_tasks}(b)-(d). 
Our results are given in \Cref{fig:widowx_results}, where we plot the performance of the learned policies averaged over 20 evaluations for \texttt{Pick-and-Place}, and 10 for the other tasks. As can be seen, using a very small number of online adaptation steps (100-150 episodes), we are able to improve the performance of $\pidp$ significantly for each task; in contrast, \rlpd is unable to learn at all. These results demonstrate the ability of \algname to steer a real-world multi-task robotic control policy with sample efficiency high enough to make real-world adaptation practical.

\iftoggle{arxiv2}{}{\vspace{-0.5em}}
\subsection{Steering Pretrained Generalist Robot Policies with \algname}\label{sec:exp_generalist}
\iftoggle{arxiv2}{}{\vspace{-0.5em}}

\begin{figure}[t]
\vspace{-0.5em}
\centering
\begin{minipage}{0.62\textwidth}
\centering
    \begin{subfigure}[b]{0.2\textwidth}
        \centering
        \includegraphics[width=\textwidth]{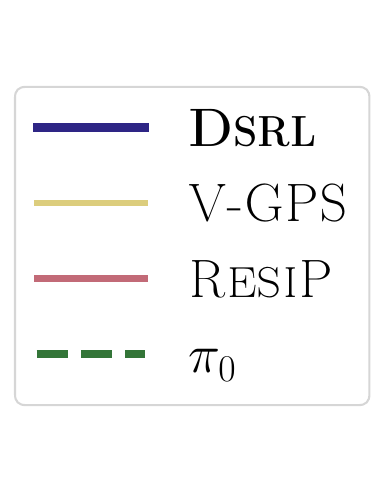}
        \vspace{0.9em}
    \end{subfigure}
    \hfill
    \begin{subfigure}[b]{0.39\linewidth}
    \centering
    \includegraphics[width=\linewidth]{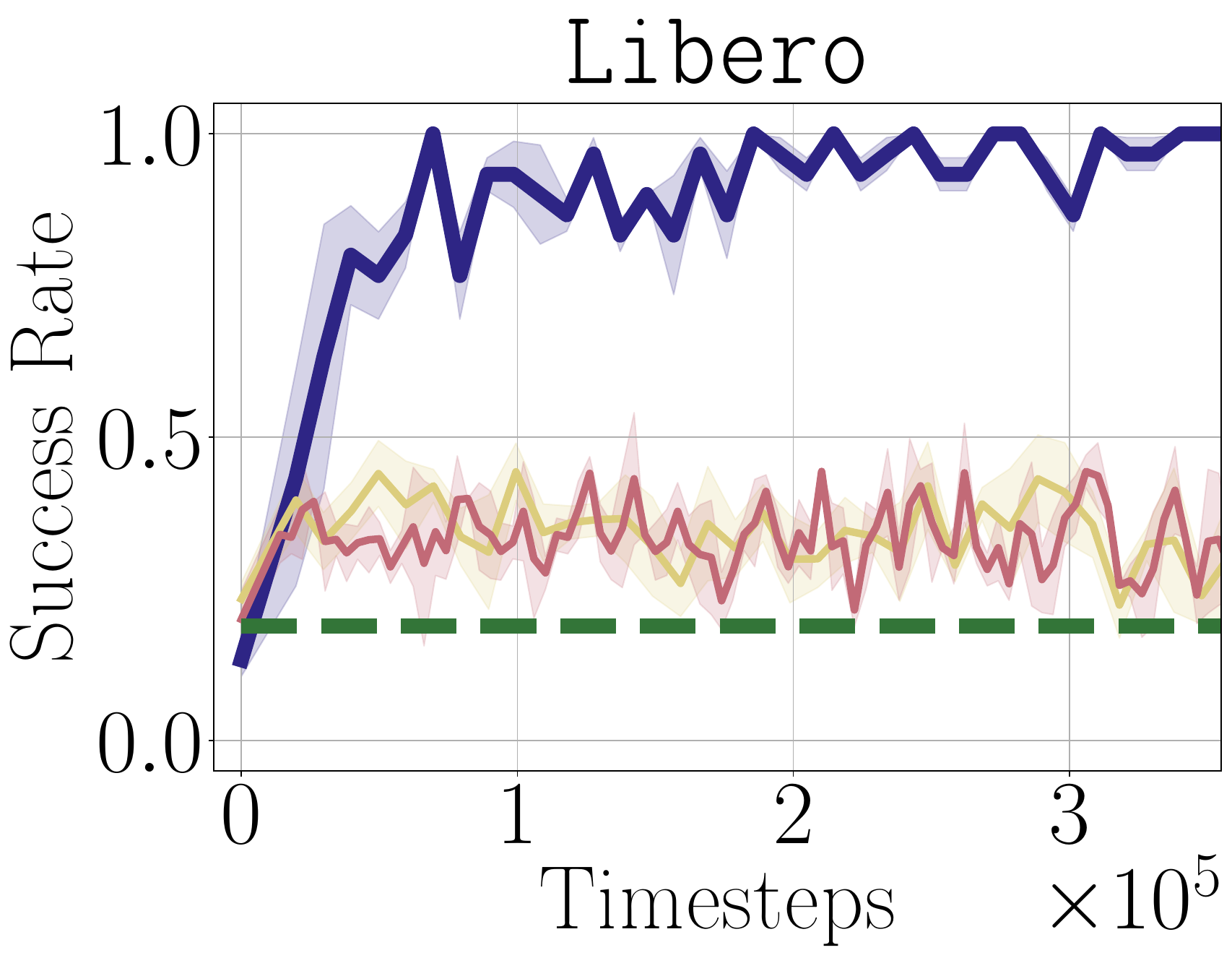}
    \label{fig:pizero_libero}
  \end{subfigure}%
  \hfill
  \begin{subfigure}[b]{0.37\linewidth}
    \centering
    \includegraphics[width=\linewidth]{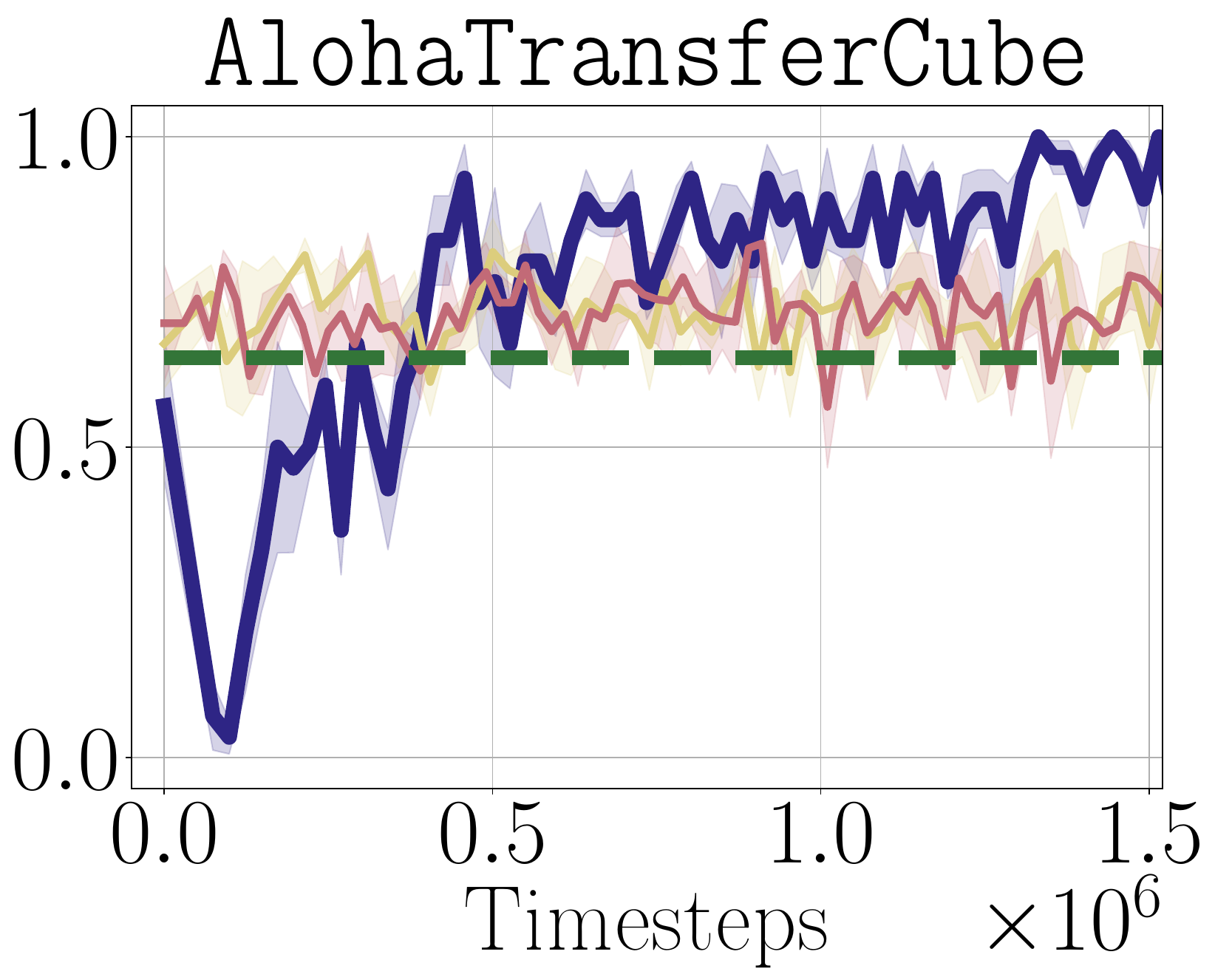}
    \label{fig:pizero_aloha}
  \end{subfigure}
  \vspace{-1.25em}
  \caption{\algname enables adaptation of $\pi_0$ \cite{black2024pi_0} on \texttt{Libero} and \texttt{Aloha} simulation tasks.}
  \label{fig:pizero_results}
\end{minipage}
\hfill
\begin{minipage}{0.35\textwidth}
        \centering

\scalebox{1}
{
\begin{tabular}{lcccccccccccc}
\toprule
\texttt{Task} & $\pi_0$ & \algname \\
\midrule
 \parbox{1.75cm}{Turn on \\ toaster} & $5/20$ & $\mathbf{18/20}$ \\ [1em]
 \parbox{1.75cm}{Put spoon\\ on plate} & $15/20$ & $\mathbf{19/20}$  \\
\bottomrule
\end{tabular}
}
\caption{
Real-world performance of $\pi_0$ and $\pi_0$ steered with \algname.
}
\vspace{-1em}
\label{table:pi0_real}
\end{minipage}
\vspace{-1em}
\end{figure}

We next evaluate if \algname can steer larger pretrained ``generalist'' policies. Here, we consider $\pi_0$ \cite{black2024pi_0}, a state-of-the-art generalist robot policy. $\pi_0$ is a 3.3B parameter model, trained to respond to language commands, and utilizes a VLM-based backbone and flow-based action head. 
Standard finetuning of policies of this size is challenging due to their scale and, furthermore, applying RL to $\pi_0$ is non-trivial due to its high-dimensional action space---$\pi_0$ uses an action chunk size of 50 and each action is 32-dimensional, resulting in a total action space dimension of 1,600.

We first evaluate \algnamesac steering of $\pi_0$ on two simulation tasks: (i) the \texttt{Libero}~\cite{liu2023libero} manipulation benchmark, where we study a pick-and-place task from the \texttt{Libero-90} suite as the downstream task, and (ii) \texttt{AlohaTransferCube}~\cite{zhao2023learning}, a challenging bimanual manipulation task that requires high precision. 
We compare against two baselines: \textsc{ResiP}~\cite{ankile2024imitation}, a residual RL-based method, and \textsc{V-GPS}~\cite{nakamoto2024steering}. Both \textsc{ResiP} and \textsc{V-GPS} require only black-box access to the pretrained policy, and aim to improve it by post-processing the actions. In all cases we use off-the-shelf public checkpoints for $\pi_0$ and average over 3 seeds. As shown in Figure~\ref{fig:pizero_results}, \algname dramatically improves $\pi_0$'s performance---for example, on \texttt{Libero}, improving from $\approx$$20\%$ to $\approx$$100\%$ after only $\approx$10,000 online samples---demonstrating the effectiveness of diffusion steering for adapting pretrained generalist policies. In contrast, neither \textsc{ResiP} nor \textsc{V-GPS} are able to significantly improve the performance of $\pi_0$. We would like to note that both tasks we study are extremely challenging for RL agents due to their complex nature---including high control frequency, long horizons, large observation and action spaces, and sparse rewards. The effectiveness of \algname in such settings, as compared to approaches such as \textsc{ResiP} or \textsc{V-GPS} which utilize more traditional RL techniques, underscores the advantages of running RL over the latent-noise action space in settings where traditional RL is challenging, and \algname's ability to more effectively exploit the pretrained policy's behavior than approaches which rely on post-processing the pretrained policy's actions.\loose

We next evaluate \algnamesac steering of $\pi_0$ in the real world. We utilize the publicly available checkpoint of $\pi_0$ trained on the DROID dataset \cite{khazatsky2024droid}, and deploy on a Franka robot arm. For our evaluation we consider two tasks: turning on a toaster, and picking up a spoon and placing it on a plate (see \Cref{fig:robot_tasks}(e) and (f)). We run \algname for 80 online episodes ($\sim$11000 total steps) on the toaster task, and 65 online episodes ($\sim$10000 total steps) on the pick-and-place task. We evaluate the performance of the base $\pi_0$ policy, and the final steered policy, and give the results in \Cref{table:pi0_real}. As these results illustrate, \algname is able to significantly improve the performance of $\pi_0$ in a relatively small number of samples on real-world tasks. To the best of our knowledge, this is the first successful demonstration of real-world RL-based finetuning of $\pi_0$.

\iftoggle{arxiv2}{\vspace{-0.5em}}{\vspace{-0.75em}}
\subsection{Understanding Diffusion Steering}\label{sec:understanding_ds}
\iftoggle{arxiv2}{}{\vspace{-0.75em}}

\begin{figure}[t]
\begin{minipage}{0.255\textwidth}
        \centering
        \includegraphics[width=\textwidth]{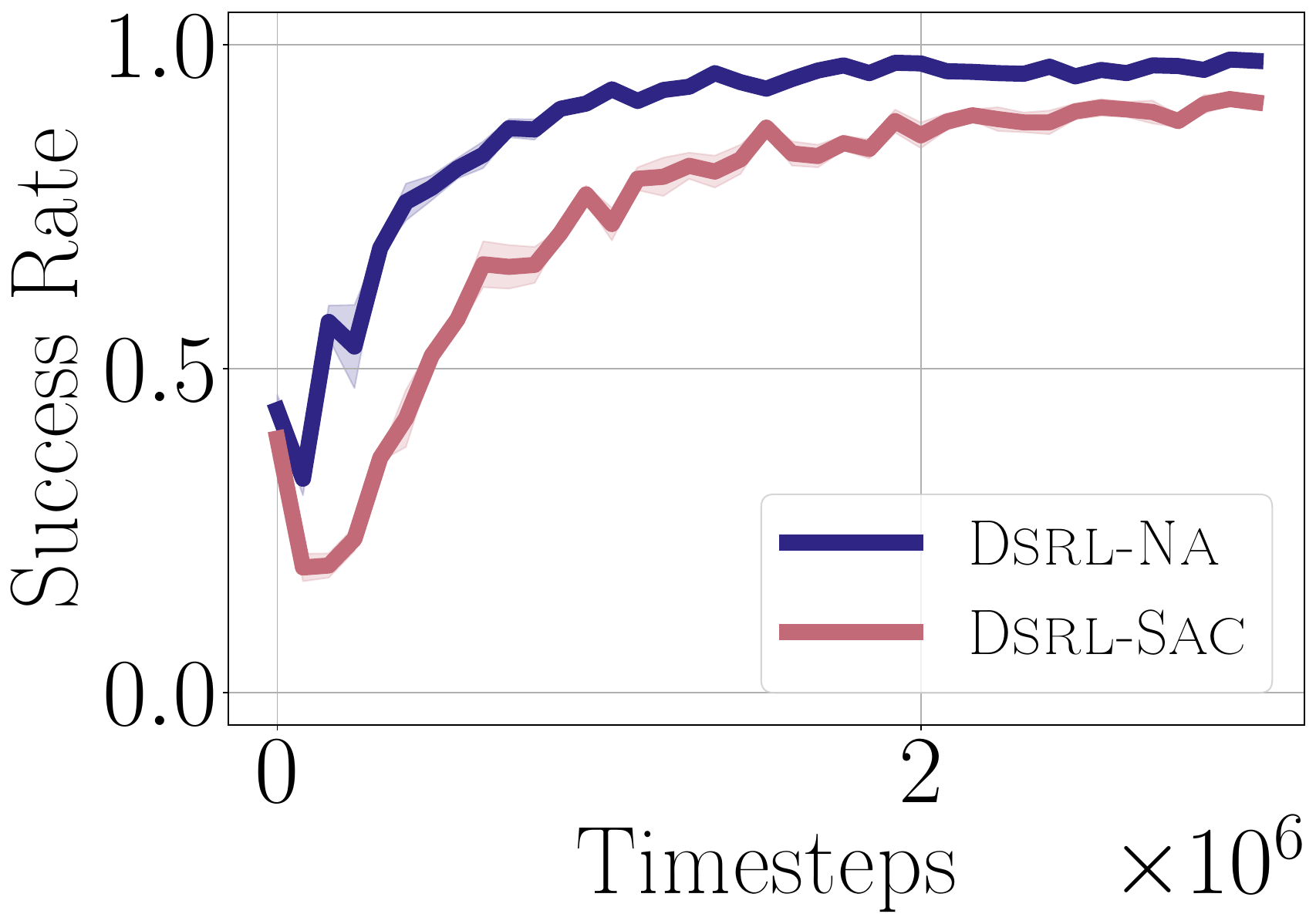}
        \caption{\algnamena improves on \algnamesac in online setting.}
        \label{fig:sac_ablation}
\end{minipage}
\hfill
\begin{minipage}{0.24\textwidth}
        \centering
        \includegraphics[width=\textwidth]{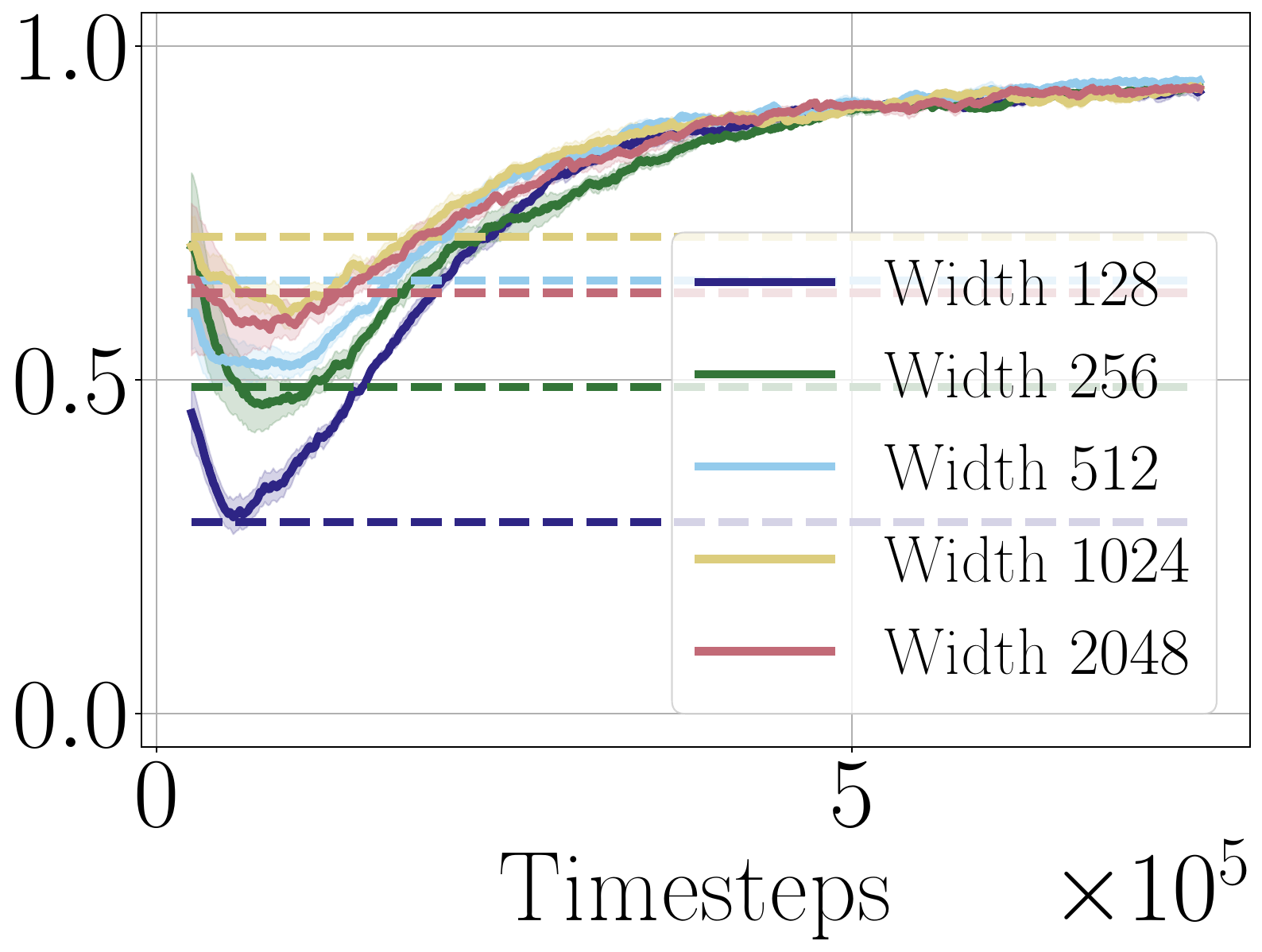}
        \caption{Sensitivity of \algname to $\pidp$'s hidden layer width.}
        \label{fig:pidp_size_ablation}
\end{minipage}
    \hfill
\begin{minipage}{0.24\textwidth}
    \centering
    \begin{subfigure}[b]{\textwidth}
        \centering
         \includegraphics[width=\textwidth]{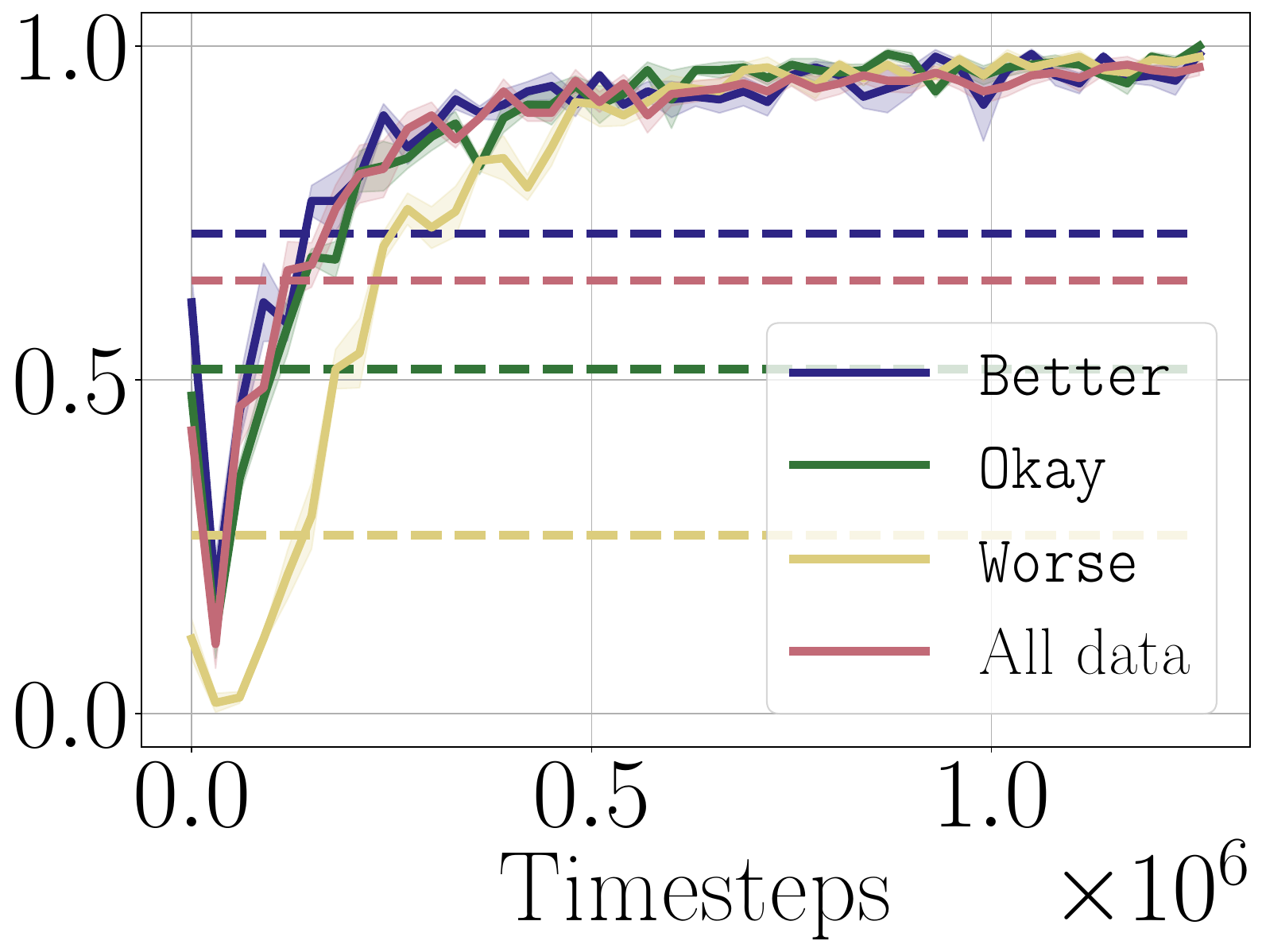}
    \end{subfigure}
    \caption{Sensitivity of \algname to quality of data $\pidp$ trained on.}
    \label{fig:ablation_data_quality}
\end{minipage}
\hfill
\begin{minipage}{0.24\textwidth}
    \centering
    \begin{subfigure}[b]{\textwidth}
        \centering
         \includegraphics[width=\textwidth]{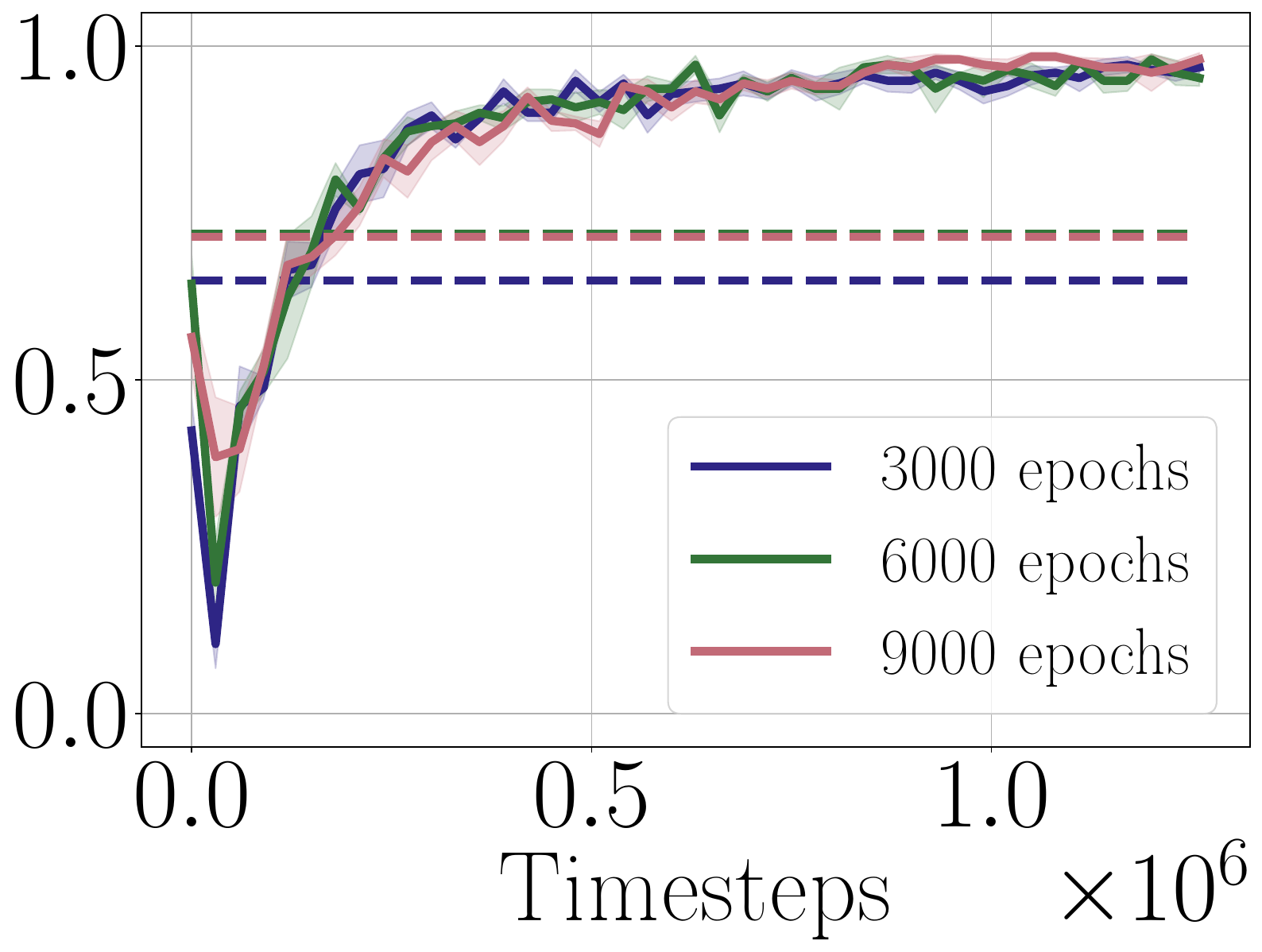}
    \end{subfigure}
    \caption{Sensitivity of \algname to number of epochs $\pidp$ trained on.}
    \label{fig:ablation_epochs}
\end{minipage}
\vspace{-1em}
\end{figure}

Finally, we seek to understand which design decisions are critical to \algname. We first study the importance of noise aliasing, \algnamena, as compared to \algnamesac. While \algnamena has advantages over \algnamesac in that it can incorporate offline data, we are interested in whether it also improves the performance of purely online learning. In \Cref{fig:sac_ablation} we compare \algnamena and \algnamesac on \texttt{Robomimic Square}. As can be seen, while \algnamesac is still able to learn, it requires $\approx$2$\times$ more samples than \algnamena, demonstrating the importance of noise aliasing for sample efficiency. 

We next consider how the base policy, $\pidp$, affects the performance of \algname. We vary three features of $\pidp$: its size, the quality of data it was trained on, and the number of epochs it was trained for. For all ablations, we run on the \texttt{Robomimic Can} environment. In all plots, the dashed line indicates the performance of $\pidp$ without steering. Please see \Cref{sec:app_ablations} for additional ablations.

To investigate the effect of the size of $\pidp$ on \algname, we consider parameterizing $\pidp$ as an MLP and varying its hidden layer width while keeping the training data fixed. We illustrate our results in \Cref{fig:pidp_size_ablation}, where we see that all model sizes can be successfully steered, despite the initial performance of $\pidp$ varying significantly. We also highlight that at sizes of 512 and 1024 the performance of $\pidp$ is essentially identical, yet \algname is able to steer the 1024 model more quickly than the 512 model. We hypothesize that this may be due to the higher capacity of the 1024 model enabling it to more effectively capture diverse, potentially successful behaviors.

To study the importance of data quality, we utilize the splits given in the Multi-Human dataset for \texttt{Robomimic Can}, which divides the data into sets collected by experienced operators (``\texttt{Better}''), adequate operators (``\texttt{Okay}''), and inexperienced operators (``\texttt{Worse}''). We train a new $\pidp$ on each split, and use this as the base policy for \algname. Our results are given in \Cref{fig:ablation_data_quality}. We see that the performance of $\pidp$ is affected significantly by the quality of data it is trained on, yet the performance of \algname only degrades for the \texttt{Worse} split, and quickly recovers to match its performance on the other splits. This demonstrates the robustness of \algname to the quality of $\pidp$---while its performance may be somewhat affected initially if $\pidp$ performs poorly, as long as $\pidp$ still exhibits some task-solving behavior, \algname is able to successfully steer it to solve the task.

Finally, we consider whether the number of epochs $\pidp$ was trained on affects the performance of $\pidp$. One might hypothesize that the larger the number of epochs $\pidp$ is trained for, the more difficult it will be to steer, as $\pidp$ becomes increasingly overfit to the training data. To test this, we train $\pidp$ for 3000, 6000, and 9000 epochs, and run \algname on each. As can be seen in \Cref{fig:ablation_epochs}, the number of epochs $\pidp$ is trained on (while affecting the performance of $\pidp$ somewhat) has almost no effect on the performance of \algname, suggesting that \algname is able to effectively steer policies that may be overfit to the demonstration data.

%% file: body/appendix.tex
\newcommand{\actmag}{b_{\cW}}

\section*{Contributions}
\textbf{Andrew Wagenmaker}: Project lead. Core algorithm development; primary paper writer; led the simulated online, offline-to-online, and ablation experiments; led the real-world multi-task diffusion policy experiments; supported $\pi_0$ experiments. \\
\textbf{Mitsuhiko Nakamoto}: Led the $\pi_0$ experiments. \\
\textbf{Yunchu Zhang}: Led the real-world single-task diffusion policy experiments and supported the simulated experiments. \\
\textbf{Seohong Park}: Led the offline experiments and contributed to algorithm development. \\
\textbf{Waleed Yagoub}: Supported the real-world single-task diffusion policy experiments. \\
\textbf{Anusha Nagabandi}: Investigated the application of \algname to dexterous manipulation. \\
\textbf{Abhishek Gupta}: Advised on the project and contributed to paper writing. \\
\textbf{Sergey Levine}: Advised on the project and contributed to paper writing.

\iftoggle{arxiv2}{
\section*{Limitations}

Our experiments show that \algname provides an effective and highly efficient method for improving diffusion and flow policies using either online or offline experience. However, our approach does have a number of limitations. First, the exploration capabilities of \algname are inherently determined by the underlying diffusion policy, and while this seems to work well in practice, we do not have a guarantee that \emph{all} diffusion policies are steerable, nor that all such policies provide adequate exploration. A highly concentrated action distribution (e.g., a policy trained on very narrow data) might not provide enough ``options'' for our method to select from, limiting its ability to improve, and our approach does not currently offer a clear way to quantify in advance how much improvement is possible with a given diffusion policy. Studying this question both empirically and theoretically could help us characterize the steerability of diffusion policies. Our method also has similar limitations to any RL approach: it requires reward signals, online rollouts (for the online RL variant), and resets (if learning in the real world). While the efficiency of our approach alleviates some of the challenges associated with RL in the real world, it still introduces additional requirements beyond the standard imitation learning methods that can be used to train the base diffusion policy. Reducing these limitations, for example by automating rewards or further improving sample efficiency, is an important direction for future work. We hope that studying these open questions will lead to even more practical methods for robotic RL, and that \algname will provide a practical and useful tool for autonomous improvement in robotic learning.}{}

\section{Additional Ablations of Diffusion Steering}\label{sec:app_ablations}

\begin{figure}[H]
\begin{minipage}{0.24\textwidth}
    \centering
    \begin{subfigure}[b]{\textwidth}
        \centering
         \includegraphics[width=\textwidth]{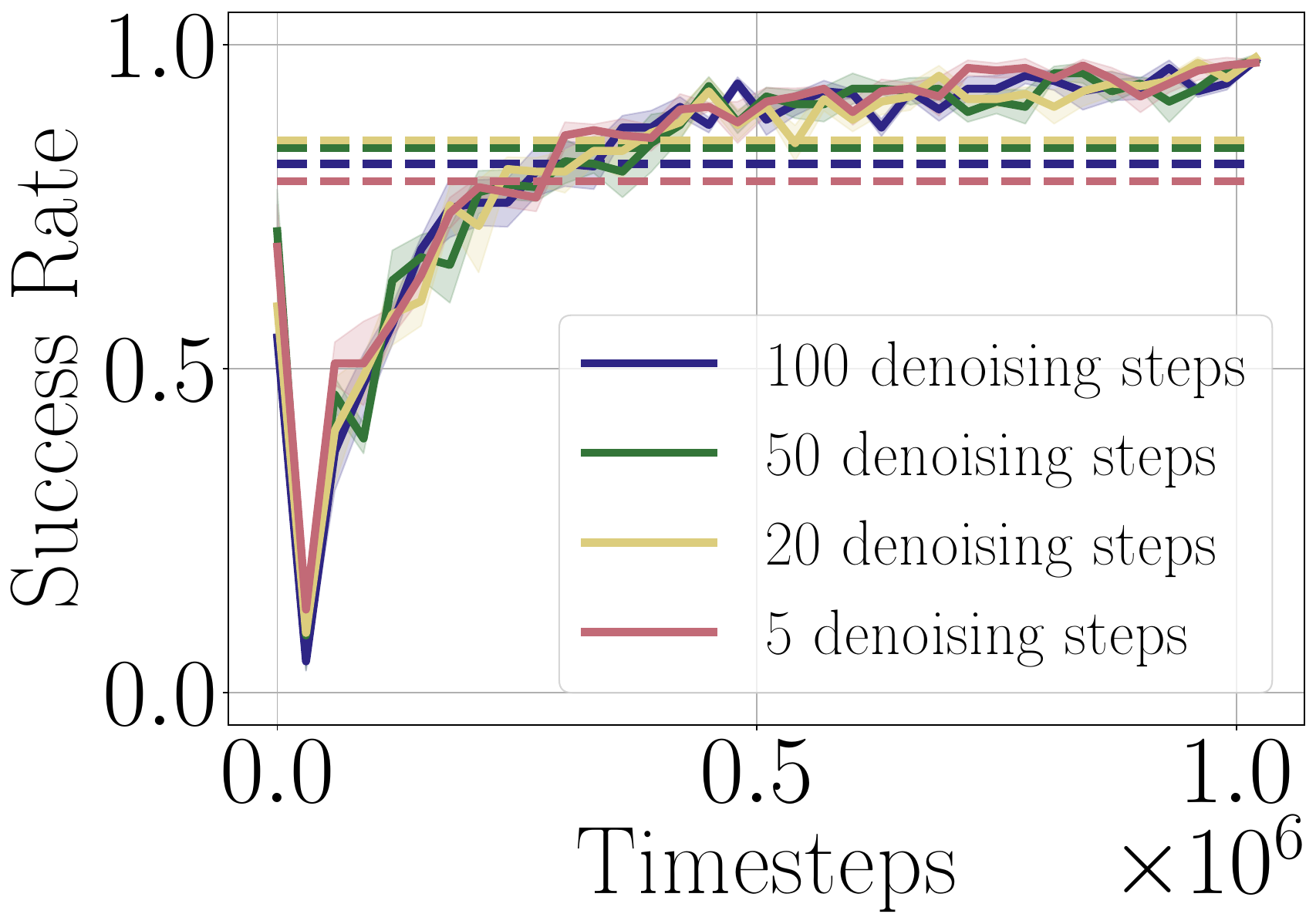}
    \end{subfigure}
    \caption{Sensitivity of \algname to number of $\pidp$ train denoising steps.}
    \label{fig:ablation_denois_steps}
\end{minipage}
\hfill
\begin{minipage}{0.24\textwidth}
    \centering
    \begin{subfigure}[b]{\textwidth}
        \centering
         \includegraphics[width=\textwidth]{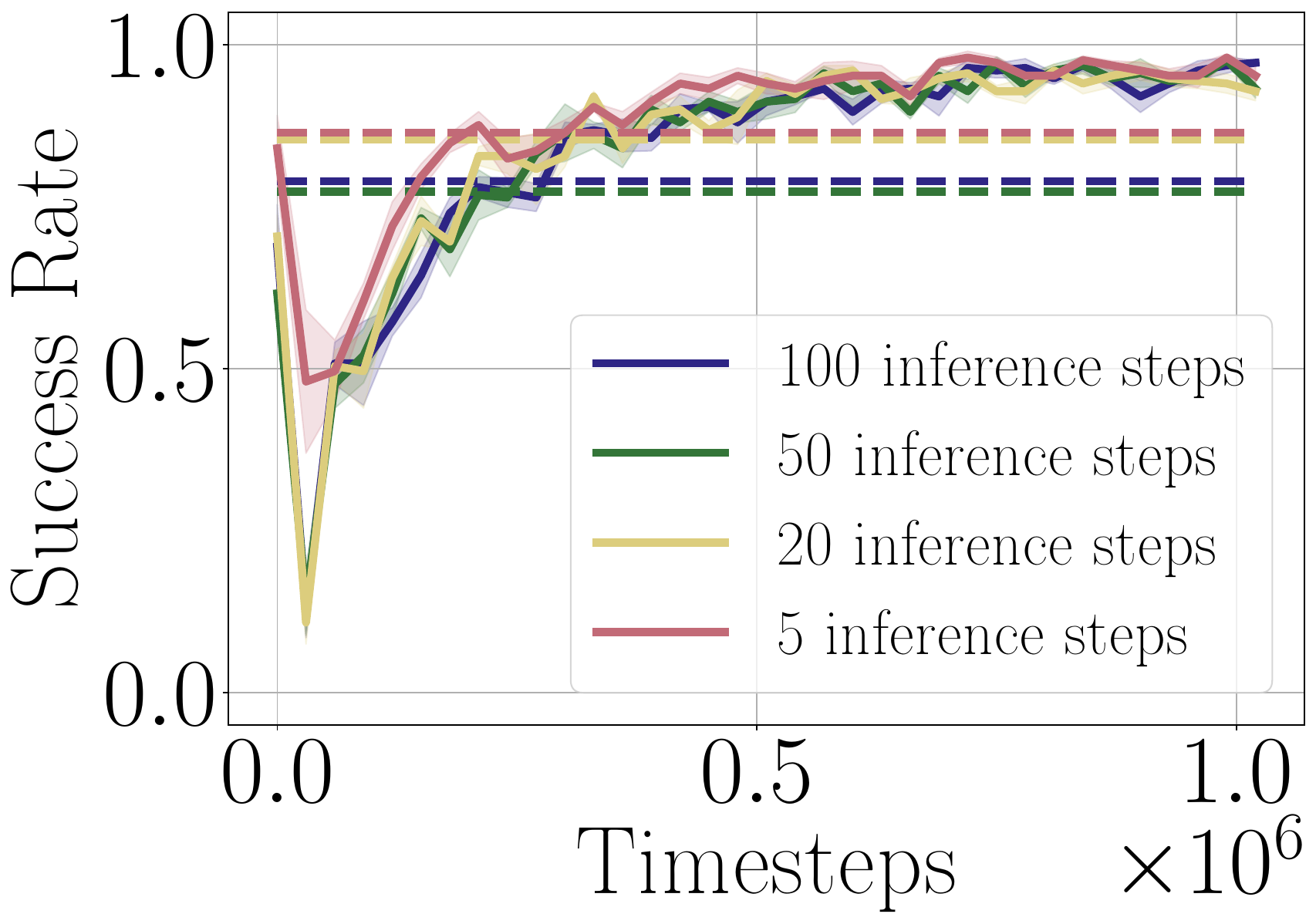}
    \end{subfigure}
    \caption{Sensitivity of \algname to number of $\pidp$ inference denoising steps.}
    \label{fig:ablation_ddim_steps}
\end{minipage}
\hfill
\begin{minipage}{0.24\textwidth}
\centering
\begin{subfigure}[b]{\textwidth}
        \centering
         \includegraphics[width=\textwidth]{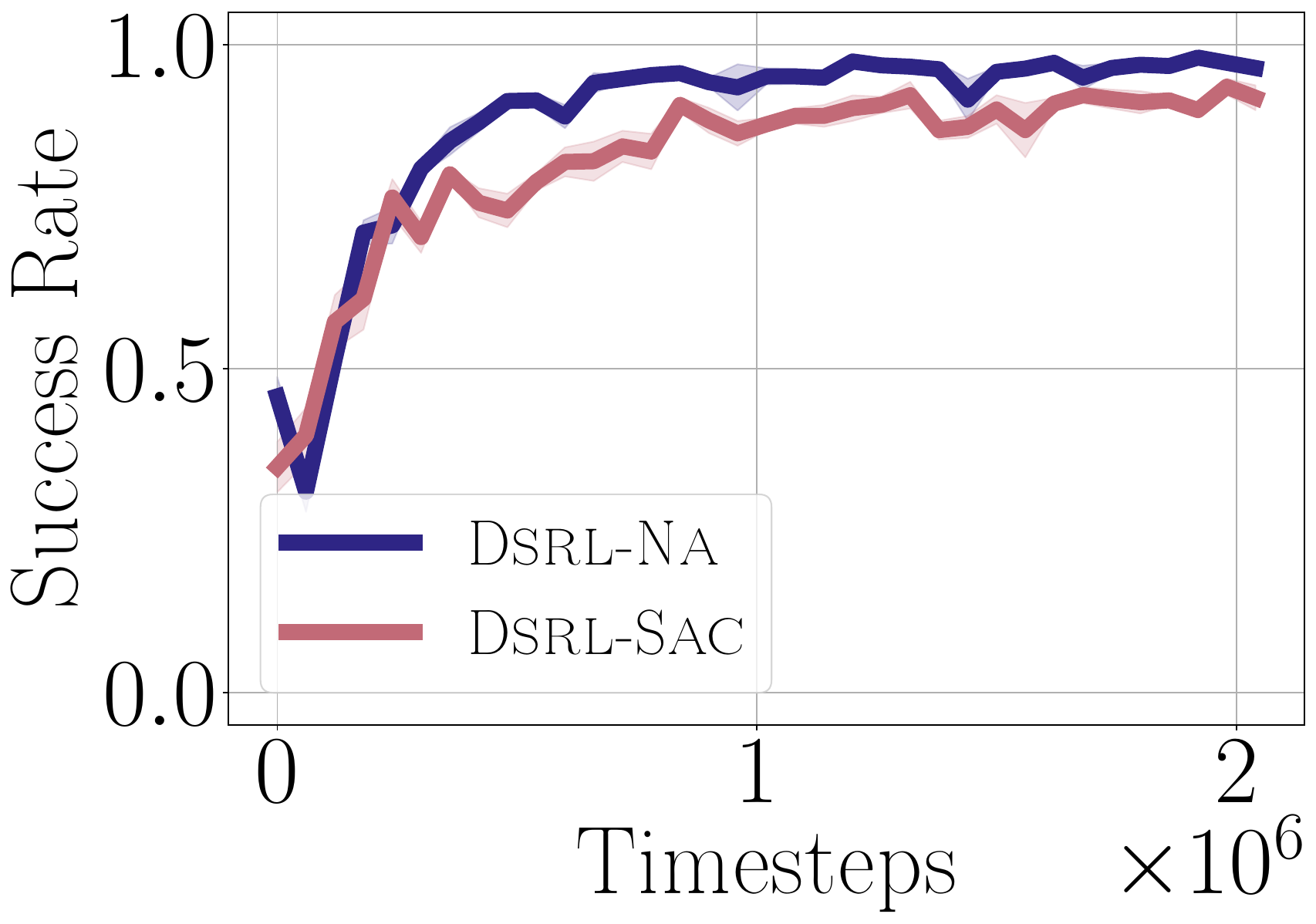}
    \end{subfigure}
    \caption{\algnamena improves on \algnamesac in online setting on \texttt{Robomimic Can}.}
    \label{fig:dsrl_sac_na_can}
\end{minipage}
\hfill
\begin{minipage}{0.24\textwidth}
    \centering
    \begin{subfigure}[b]{\textwidth}
        \centering
         \includegraphics[width=\textwidth]{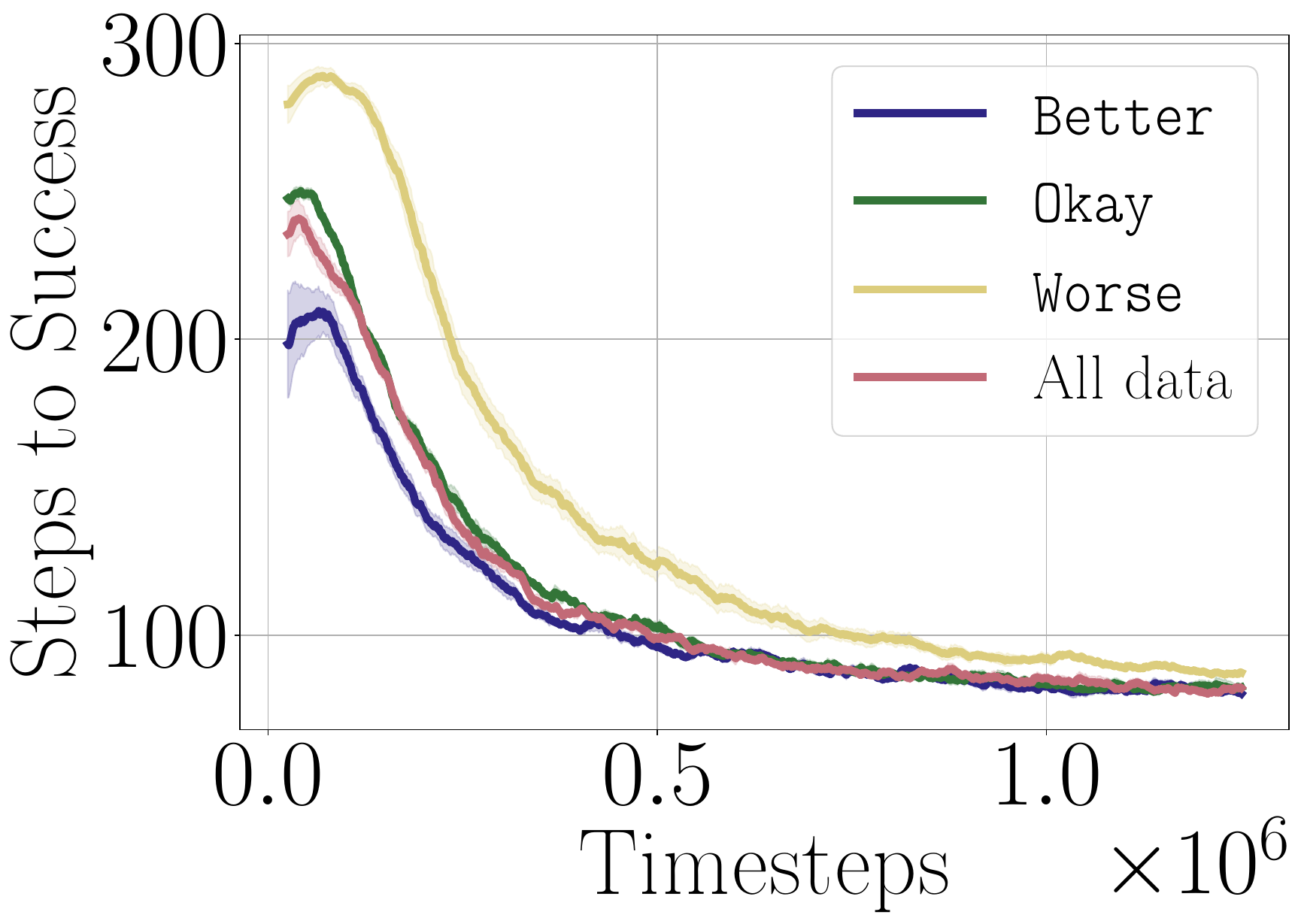}
    \end{subfigure}
    \caption{Sensitivity of \algname's solution quality to quality of data $\pidp$ trained on.}
    \label{fig:ablation_data_quality_time}
\end{minipage}
\end{figure}

In this section we expand on the results from \Cref{sec:understanding_ds}, providing several additional experiments that seek to understand when \algname is an effective approach.
All experiments in this section are run on the \texttt{Robomimic Can} environment. Unless otherwise noted, each line corresponds to the average over 5 random seeds, and we use the hyperparameters outlined in \Cref{sec:online_exp_details} for \texttt{Can}. In all plots, dashed lines denote the performance of the base diffusion policy, $\pidp$ (averaged over 200 trials).

\paragraph{Does the number of denoising steps $\pidp$ uses affect the performance of \algname?}
Both diffusion and flow policies rely on an iterative denoising process, \eqref{eq:reverse_process}, to generate actions. An important hyperparameter in the denoising process is the number of steps that it uses, with typically more steps leading to higher quality generation at the expense of longer inference times. We can split the number of steps used in denoising into the value used in training (at what intervals to add the noise), and the number of steps taken at inference time. We test how the number of steps affects the behavior of \algname, scaling both the number of train and inference steps. In \Cref{fig:ablation_denois_steps}, we vary the number of steps we train on, while keeping the number of steps used at inference time fixed to 5, while in \Cref{fig:ablation_ddim_steps} we train a single policy with 100 denoising steps, and vary the number of inference steps used. We see that neither the number of train steps or inference steps has virtually any impact on the performance of \algname.

\paragraph{Additional details on \Cref{fig:sac_ablation}}
For the experiment of \Cref{fig:sac_ablation}, for \algnamena we use the hyperparameters given in \Cref{sec:online_exp_details} for \texttt{Robomimic Square}. For \algnamesac we swept over hyperparameters and included the results for best-performing ones, which were identical to those used for \algnamena except for using a hidden size of $128$. 
In \Cref{fig:dsrl_sac_na_can} we also include results on \texttt{Robomimic Can}, comparing \algnamena to \algnamesac, where we see a similar trend as observed in \Cref{fig:sac_ablation}, albeit somewhat less pronounced.

\paragraph{Additional details on \Cref{fig:pidp_size_ablation}}
For the experiment of \Cref{fig:pidp_size_ablation}, we use the hyperparameters given in \Cref{sec:online_exp_details} for \texttt{Robomimic Can} (with the exception of $\pidp$'s layer size, which we vary).

\paragraph{Additional details on \Cref{fig:ablation_data_quality}.}

To complement \Cref{fig:ablation_data_quality}, in \Cref{fig:ablation_data_quality_time}, we also plot the number of steps to success the steered policy requires.
We see that the final steered policy \algname converges to for $\pidp$ trained on the \texttt{Worse} split is slightly less optimal in terms of time-to-success than the other cases. We conclude that the quality of $\pidp$ does play a role in how quickly \algname converges and the quality of the solution it converges to, yet find that \algname is surprisingly robust to variation in the performance of $\pidp$.

We note as well that, compared to policies trained on only a single split of the data, the policy trained on all splits sees a more diverse set of behaviors, covering a larger number of possible solution paths. One may hypothesize that a policy could be more easily steered given the larger number of possible behaviors it has learned, but we do not find that to be the case here---training on all datasets does not lead to a performance improvement for \algname as compared to training on a narrower dataset.

\section{Best Practices for \algname}
We found the following design choices led to the best results for \algname. Several of these are consistent with current best practices for sample-efficient RL (e.g. \cite{ball2023efficient}), while some are particular to \algname:
\begin{itemize}
    \item \textbf{High UTD}: In general we used a UTD of at least 20. We found that this led to faster convergence and higher sample efficiency, without sacrificing performance. Higher UTD seems to be particular valuable when running the \algnamena variant of \algname.
    \item \textbf{Layer-norm}: We found applying layer norm to all but the final layer for both the critic and actor networks also led to improved performance.
    \item \textbf{Large policy and critic networks}: For all our experiments we instantiate our latent-noise actor and critics, $\piw$ and $\Qw$, as MLPs. We found that using larger MLPs led to better performance---in many experiments we used networks with layer size up to 2048.
    \item \textbf{Action magnitude}: When transforming the MDP from $\cM$ to $\cMdp$, one design choice is how large a magnitude the actor can play in $\cW$. If $\pidp$ is trained by adding noise sampled from a standard Gaussian to the demonstration actions, as is typical, then only $\bw$ reasonably likely under a standard Gaussian would be ``in-distribution'' for $\pidp(\bs,\bw)$---$\pidp$ would have been unlikely to have denoised an example corresponding to $\bw$ outside this range during training, and therefore its behavior on such examples may be unreliable. To incorporate this, we simply constrain the actions to a rectangle of some width $\actmag$, i.e. in practice we set $\cW \leftarrow [-\actmag, \actmag]^d$. We found that choosing $\actmag$ in the range $[1,3]$ generally worked well for online RL and $[0.5,1.5]$ for offline RL. See below for the value of $\actmag$ used in each experiment.
\end{itemize}

\section{Additional Experimental Details}

\subsection{Details of Online Experiments (\Cref{sec:exp_online})}\label{sec:online_exp_details}

For all baselines, we use the implementations and hyperparameters given by \citet{ren2024diffusion} (please see Section E.3 of \cite{ren2024diffusion} for a detailed overview of baseline methods). As noted in the main text, for \texttt{Robomimic Lift} and \texttt{Can} and all \texttt{Gym} tasks, we utilize the diffusion policy checkpoints provided by \citet{ren2024diffusion} for $\pidp$ (in particular, we use the identical checkpoints as they utilize for their online finetuning experiments). For \texttt{Robomimic Square} and \texttt{Transport} we train new diffusion policies with a larger number of denoising steps than those provided by \citet{ren2024diffusion} (100 vs 20) which we found to increase the performance of $\pidp$ somewhat. To train these policies, we use the diffusion policy training code provided by \citet{ren2024diffusion}, and keep all hyperparameters fixed other than the number of denoising steps.
For all methods we utilize DDIM sampling, and note the number of denoising steps used at inference time in the tables below. As suggested in \citet{ren2024diffusion}, during training, we experiment with injecting additional noise into the DDIM denoising process to help induce additional exploration. For each baseline method, we run variants with and without this additional noise injection, and include results for the best-performing one. 
We utilize the default rewards for each task (in particular, for all \texttt{Robomimic} tasks the reward is a 0-1 success reward).

We utilize \algnamena for all environments but \texttt{Robomimic Transport} where we use \algnamesac. For \algname, in all experiments we run 4 environments in parallel. We update the actor and critic after every (parallel) environment step, taking the number of gradient steps at each update as stated below (we take the same number of steps on both the actor and critic). For updating the latent-noise critic $\Qw$ in \algnamena, we denote the number of gradient steps per update as ``$\Qw$ update steps'' in the following. For \algnamesac we use the implementation of \sac given by \cite{raffin2021stable}, and for \algnamena we use an entropy penalty identical to that used for \sac, otherwise running \algnamena as stated in \Cref{alg:main}. For each environment, we first roll out $\pidp$ with $\bw \sim \cN(0,I)$ for some number of steps (denoted as ``Initial rollouts'' in the following) and place the data collected in the replay buffer, before starting the actor and critic online training updates. The policies given by \citet{ren2024diffusion} all utilize action chunking (``Action chunk size'' in the following). To handle RL with action chunking, we treat the entire chunk as a single action and ignore the observations from within the chunk, treating the entire chunk as a single step in the environment (however, the number of timesteps stated in all results gives the total number of timesteps in the original environment).

In all plots, error bars denote 1 standard error.
All experiments in this section were run on an NVIDIA A5000 GPU.

\begin{table}[H]
\caption{
\footnotesize
\textbf{Common \algname hyperparameters for online experiments.}
}
\vspace{5pt}
\label{tab:dsrl_online_hyperparams}
\begin{center}
\scalebox{0.9}
{
\begin{tabular}{ll}
    \toprule
    \textbf{Hyperparameter} & \textbf{Value} \\
    \midrule
    Learning rate & $0.0003$ \\
    Batch size & $256$ \\
    Activation & Tanh \\
    Target entropy & $0$ \\
    Target update rate ($\tau$) & $0.005$ \\
    Number of actor and critic layers & $3$ \\
    Number of critics & $2$ \\
    Number of environments & $4$ \\
    \bottomrule
\end{tabular}
}
\end{center}
\end{table}

\begin{table}
\caption{
\footnotesize
\textbf{Hyperparameters for \algname \texttt{Robomimic} experiments.}
}
\vspace{5pt}
\begin{center}
\scalebox{0.9}
{
\begin{tabular}{lllll}
    \toprule
    \textbf{Hyperparameter} & \texttt{Lift} & \texttt{Can} & \texttt{Square} & \texttt{Transport}  \\
    \midrule
    Action chunk size & $4$ & $4$ & $4$ & $8$  \\
Hidden size & $2048$  & $2048$ & $2048$ & $128$  \\
Gradient steps per update & $30$ & $20$ & $20$ & $20$  \\
$\Qw$ update steps & $10$ & $10$ & $10$ & - \\
Discount factor & $0.99$ & $0.99$ & $0.999$ & $0.99$  \\
Action magnitude ($\actmag$) & $1.5$ & $1.5$ & $1.5$ & $1$  \\
Initial steps & $24000$ & $24000$ & $32000$ & $320000$ \\
$\pidp$ train denoising steps & $20$ & $20$ & $100$ & $100$ \\
$\pidp$ inference denoising steps & $8$ & $8$ & $8$ & $100$ \\
    \bottomrule
\end{tabular}
}
\end{center}
\end{table}

\begin{table}
\caption{
\footnotesize
\textbf{Hyperparameters for \algname \texttt{OpenAI Gym} experiments.}
}
\vspace{5pt}
\begin{center}
\scalebox{0.9}
{
\begin{tabular}{llll}
    \toprule
    \textbf{Hyperparameter} &  \texttt{Hopper-v2} & \texttt{Walker2D-v2} & \texttt{HalfCheetah-v2} \\
    \midrule
Action chunk size  & $4$ & $4$ & $4$ \\
Hidden size & $2048$ & $2048$ & $1024$ \\
Gradient steps per update  & $20$ & $20$ & $20$ \\
$\Qw$ update steps & $10$ & $10$ &  $10$ \\
Discount factor  & $0.99$ & $0.99$ & $0.99$ \\
Action magnitude ($\actmag$)  & $1.5$ & $2.5$ & $1.5$ \\
Initial steps & $32000$ & $32000$ & $32000$ \\
$\pidp$ train denoising steps  & $20$ & $20$ & $20$ \\
$\pidp$ inference denoising steps & $5$ & $5$ &  $5$ \\
    \bottomrule
\end{tabular}
}
\end{center}
\end{table}

\subsection{Details of Offline Experiments (\Cref{sec:exp_offline})}

In the offline setting, we utilize the flow policy implementation given by \citet{park2025flow}, and train each flow policy for 1M steps. For each environment we run on the default task given in \texttt{OGBench} \cite{park2024ogbench}, and specify the dataset used in the task name given in \Cref{table:offline}. As noted in the main text, all baseline values are taken directly from \cite{park2025flow}. While \algname was run with the same number of gradient steps and same network sizes as all baseline methods, we note that we utilize 10 critics for \algname (with mean aggregation), while all baselines were run with only 2 critics each, and thus the total number of parameters \algname utilizes is larger than that used by the baselines.

For hyperparameters, we mostly follow the original choices by \citet{park2025flow},
and present the full list of hyperparameters used in our offline experiments in \Cref{table:offline_hyp}.
Also, following \citet{park2025flow}, we individually tune the degree of conservatism (i.e., the action magnitude $\actmag$) for each task, and report the per-task hyperparameters in \Cref{table:offline_pertask_hyp}.
In \Cref{table:offline}, ``$\pm$'' denotes 1 standard deviation. All experiments in this section were run on an NVIDIA A5000 GPU.

\begin{table}
\caption{
\footnotesize
\textbf{Common \algname hyperparameters for offline experiments.}
}
\vspace{5pt}
\label{table:offline_hyp}
\begin{center}
\scalebox{0.9}
{
\begin{tabular}{ll}
    \toprule
    \textbf{Hyperparameter} & \textbf{Value} \\
    \midrule
    Learning rate & $0.0003$ \\
    Batch size & $256$ \\
    Activation & GELU \\
    Target update rate ($\tau$) & $0.005$ \\
    Hidden size & $512$ \\
    Number of actor and critic layers & $4$ \\
    Number of critics & $10$ \\
    Flow steps & $10$ \\
    Clipped double Q-learning & False \\
    \bottomrule
\end{tabular}
}
\end{center}
\end{table} 

\begin{table}[H]
\caption{
\footnotesize
\textbf{Per-task \algname hyperparameters for offline experiments.}
}
\vspace{5pt}
\label{table:offline_pertask_hyp}
\begin{center}
\scalebox{0.9}
{
\begin{tabular}{lll}
    \toprule
    \textbf{Task} & \textbf{Action Magnitude ($\actmag$)} & \textbf{Discount Factor} \\
    \midrule
    \texttt{antmaze-large-navigate-singletask} & $1.25$ & $0.99$ \\
    \texttt{antmaze-giant-navigate-singletask} & $1.25$ & $0.995$ \\
    \texttt{humanoidmaze-medium-navigate-singletask} & $0.5$ & $0.995$ \\
    \texttt{humanoidmaze-large-navigate-singletask} & $0.75$ & $0.995$ \\
    \texttt{antsoccer-arena-navigate-singletask} & $0.75$ & $0.995$ \\
    \texttt{cube-single-play-singletask} & $0.5$ & $0.99$ \\
    \texttt{cube-double-play-singletask} & $1.5$ & $0.99$ \\
    \texttt{scene-play-singletask} & $0.75$ & $0.99$ \\
    \texttt{puzzle-3x3-play-singletask} & $0.5$ & $0.99$ \\
    \texttt{puzzle-4x4-play-singletask} & $0.5$ & $0.99$ \\
    \bottomrule
\end{tabular}
}
\end{center}
\end{table}

\subsection{Details of Offline-to-Online Experiments (\Cref{sec:exp_offline_to_online})}
For \algname, we utilize identical hyperparameters as are used for \texttt{Robomimic Can} and \texttt{Square} in \Cref{sec:exp_online}. For \rlpd and \textsc{Cal-Ql}, we plot the numbers given in \cite{ren2024diffusion}.

\subsection{Details of Real-World Experiments (\Cref{sec:exp_widowx})}

\subsubsection{Single-Task Diffusion Policy}
In the Franka single-task experiments, for $\pidp$ we trained a diffusion policy using the CNN architecture proposed by \cite{chi2023diffusion} on a self-collected dataset. Please see \Cref{tab:serl_diffusion_params} for the hyperparameters used for $\pidp$. 
For \algname, we begin by running $\pidp$ for 10 episodes sampling $\bw \sim \cN(0,I)$, and initialize the replay buffer with the data collected from these rollouts. For \rlpd and \rlpd with intervention, every batch is sampled with half taken from the 10 human demonstrations used to train $\pidp$, and half taken from additional data collected online, as is outlined in \cite{ball2023efficient}. 
For \rlpd and \rlpd with intervention, we use the hyperparameters used by \citet{luo2024precise}, which were found to work well on a similar real-world robot policy learning task. For \algname, we use the hyperparameters given in \Cref{tab:dsrl_online_hyperparams} (with the exception of ``Number of environments'', which is 1 in the real-world setting) and \Cref{tab:serl_diffusion_params}. For both \algname and \rlpd we update the actor and critic at every environment step, for \algname taking the number of gradient steps given in \Cref{tab:serl_diffusion_params} per update.

In the cube pick-and-place task, we define an episode as successful if the cube is dropped into the bowl. We utilize a 0-1 success reward for each task. See \Cref{fig:widowx_setup2} for additional visualizations of our setups.

\begin{figure}[H]
    \centering
    \includegraphics[width=0.3\textwidth]{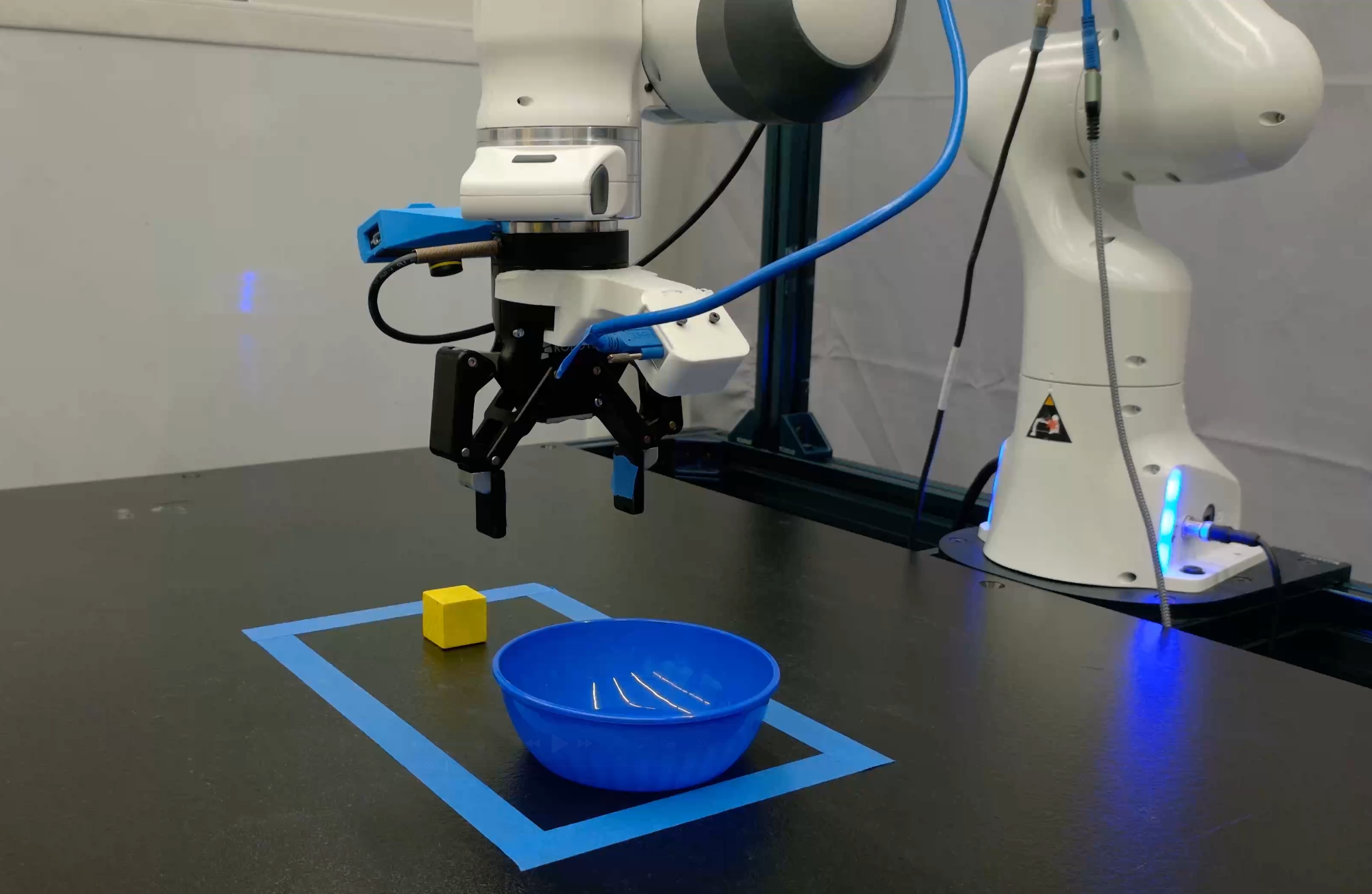}
    \caption{Single-task Franka setup.}
    \label{fig:franka_setup}
\end{figure}

\begin{table}
\caption{
\footnotesize
\textbf{Hyperparameters for $\pidp$ in Franka experiments.}
}
\label{tab:serl_diffusion_params}
\vspace{5pt}
\begin{center}
\scalebox{0.9}
{
\begin{tabular}{ll}
    \toprule
    \textbf{Hyperparameter} & Value \\
    \midrule
Batch size &  $256$ \\
Learning rate & $0.0003$ \\
Training steps & $3500$ \\
Warmup steps & $0$ \\
\hline
Action chunk size & $1$ \\
Train denoising steps & $100$ \\
Inference denoising steps & $8$ \\
Image encoder & ResNet-18 \\
Hidden size & $1024$ \\
Number of Heads & $1$ \\
Number of Layers & $3$ \\
Feedforward dimension & $512$ \\
    \bottomrule
\end{tabular}
}
\end{center}
\end{table}

\subsubsection{Multi-Task Diffusion Policy}
In the WidowX experiments, for $\pidp$ we trained a diffusion policy using the DiT architecture proposed by \cite{dasari2024ingredients} on the Bridge v2 dataset. Please see \Cref{tab:dit_params} for the hyperparameters used for $\pidp$. 
We trained $\pidp$ with image goal conditioning, and in deployment would condition it on a goal image of the desired final position. 

For both \algname and \rlpd, we begin by running $\pidp$ for 20 episodes sampling $\bw \sim \cN(0,I)$, and initialize the replay buffer with the data collected from these rollouts. For \rlpd, every batch is sampled with half taken from the 20 initial rollouts of $\pidp$, and half taken from additional data collected online, as is outlined in \cite{ball2023efficient}. For \rlpd, we use the hyperparameters used by \citet{luo2024precise}, which were found to work well on a similar real-world robot policy learning task. For \algname, we use the hyperparameters given in \Cref{tab:dsrl_online_hyperparams} (with the exception of ``Number of environments'', which is 1 in the real-world setting) and \Cref{tab:widowx_dsrl_params}. As input features for both \algname and \rlpd, we provide the proprioceptive state, and also the image features learned by $\pidp$'s pretrained ResNet encoder.
For both \algname and \rlpd we update the actor and critic every other environment step, for \algname taking the number of gradient steps given in \Cref{tab:widowx_dsrl_params} per update.

\begin{figure}
    \centering
    \begin{subfigure}[b]{0.3\textwidth}
        \centering
        \includegraphics[width=\textwidth]{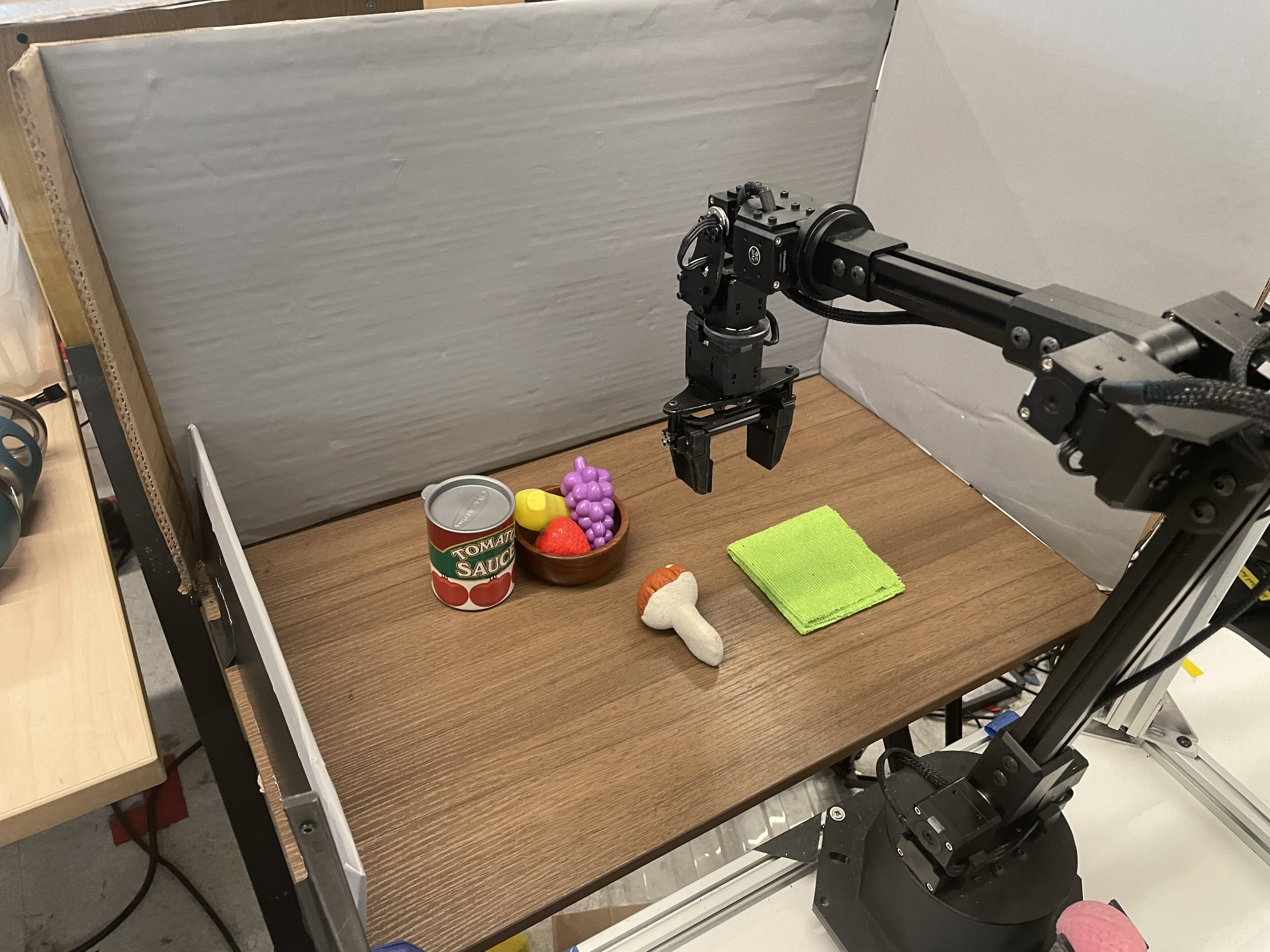}
        \caption{Pick-and-place task.}
        \label{fig:mushroom}
    \end{subfigure}
    \hfill
    \begin{subfigure}[b]{0.3\textwidth}
        \centering
        \includegraphics[width=\textwidth]{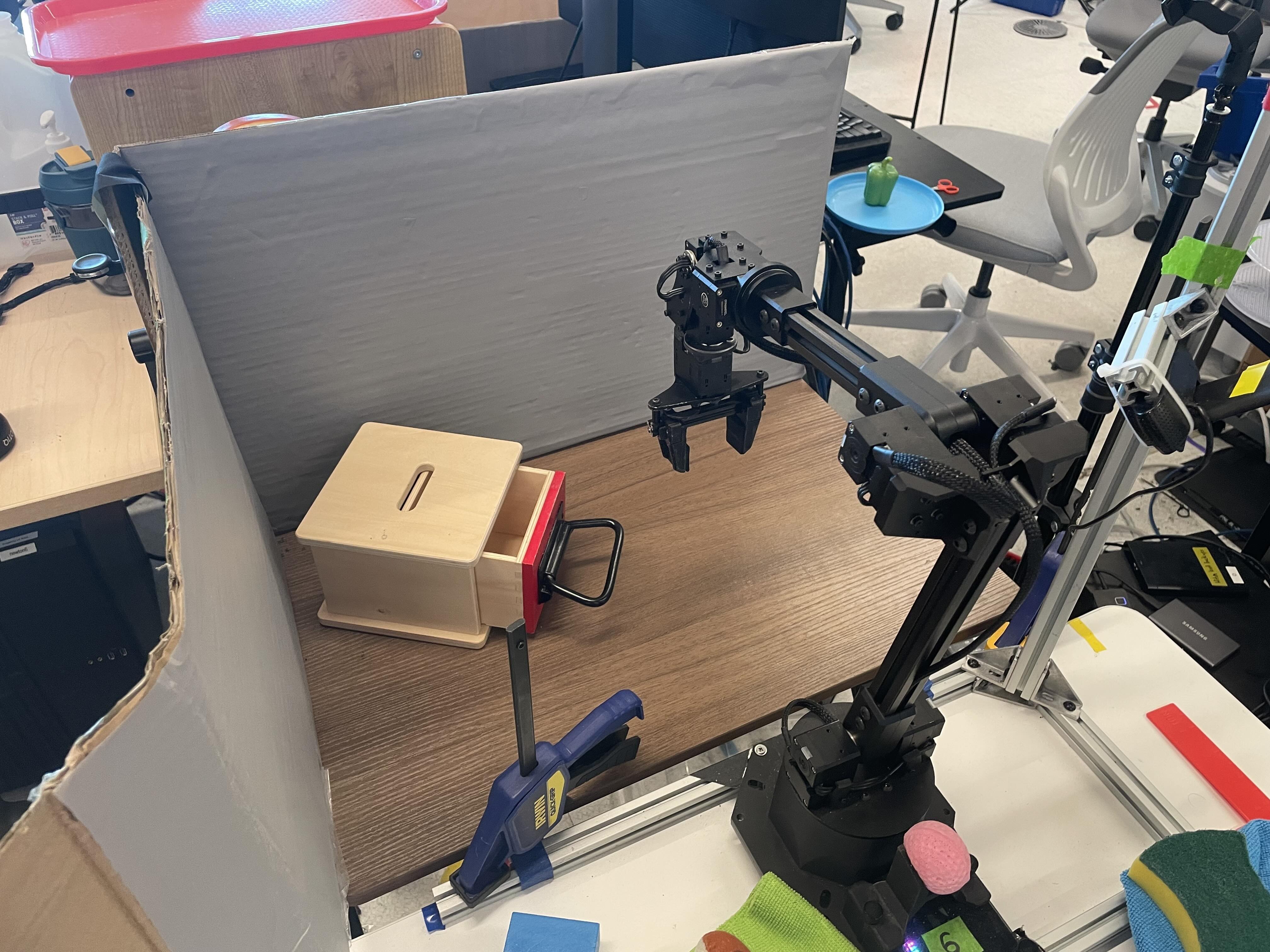}
        \caption{Drawer closing task.}
        \label{fig:drawer}
    \end{subfigure}
    \hfill
    \begin{subfigure}[b]{0.3\textwidth}
        \centering
        \includegraphics[width=\textwidth]{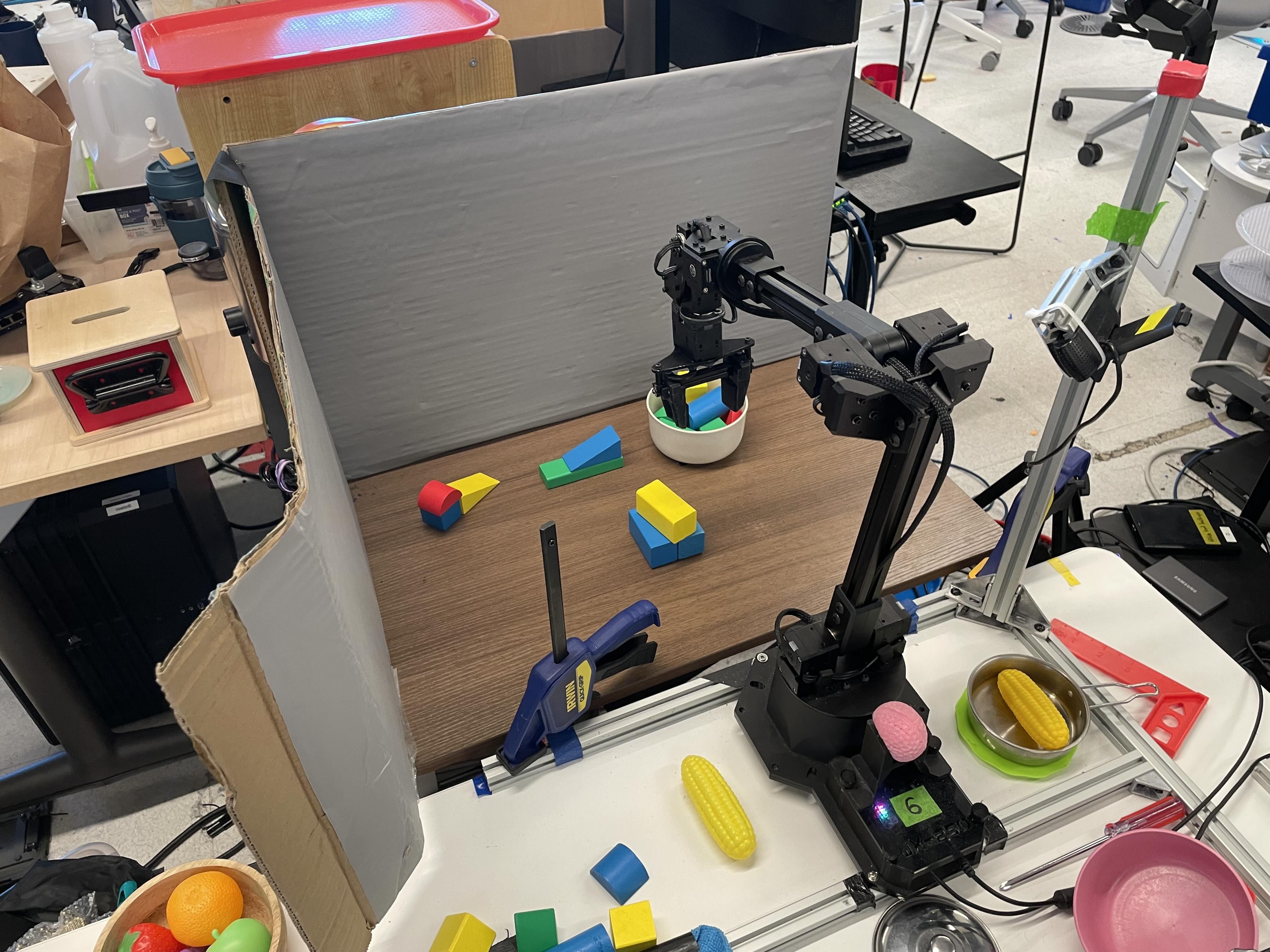}
        \caption{Block stacking task.}
        \label{fig:third}
    \end{subfigure}
    \caption{Setups for multi-task WidowX diffusion policy experiments.}
    \label{fig:widowx_setup2}
\end{figure}

For the pick-and-place task, we define an episode as successful if the mushroom is at least half on the cloth, and the WidowX has released the mushroom. For the drawer closing task we define an episode as successful if the drawer is successfully shut. For the block stacking task, we define an episode as successful if the yellow block is on the blue blocks and the WidowX has released the yellow block. We utilize a 0-1 success reward for each task. See \Cref{fig:widowx_setup2} for additional visualizations of our setups.

For \algname training, we utilize an NVIDIA GeForce RTX 4090 GPU.

\begin{table}
\caption{
\footnotesize
\textbf{Hyperparameters for DiT $\pidp$ in WidowX experiments.}
}
\label{tab:dit_params}
\vspace{5pt}
\begin{center}
\scalebox{0.9}
{
\begin{tabular}{ll}
    \toprule
    \textbf{Hyperparameter} & Value \\
    \midrule
Batch size &  $2048$ \\
Learning rate & $0.0003$ \\
Training steps & $100000$ \\
LR scheduler & cosine \\
Warmup steps & $2000$ \\
\hline
Action chunk size & $1$ \\
Train denoising steps & $100$ \\
Inference denoising steps & $8$ \\
Image encoder & ResNet-34 \\
Hidden size & $256$ \\
Number of Heads & $1$ \\
Number of Layers & $3$ \\
Feedforward dimension & $512$ \\
    \bottomrule
\end{tabular}
}
\end{center}
\end{table}

\begin{table}
\caption{
\footnotesize
\textbf{Hyperparameters for \algname WidowX experiments.}
}
\label{tab:widowx_dsrl_params}
\vspace{5pt}
\begin{center}
\scalebox{0.9}
{
\begin{tabular}{llll}
    \toprule
    \textbf{Hyperparameter} & \texttt{Pick-and-Place} & \texttt{Drawer Closing} & \texttt{Block Stacking}  \\
    \midrule
Hidden size & $1024$  & $2048$ & $1024$  \\
Gradient steps per update & $30$ & $20$ & $20$   \\
Discount factor & $0.97$ & $0.97$ & $0.97$  \\
Action magnitude ($\actmag$) & $2$ & $2$ & $1.5$  \\
    \bottomrule
\end{tabular}
}
\end{center}
\end{table}

\subsection{Details of $\pi_0$ Experiments (\Cref{sec:exp_generalist})}

\begin{figure}[b]
\centering
\hspace{-2em}
\begin{minipage}{0.55\textwidth}
    \begin{subfigure}[b]{0.45\textwidth}
        \centering
        \includegraphics[height=\textwidth]{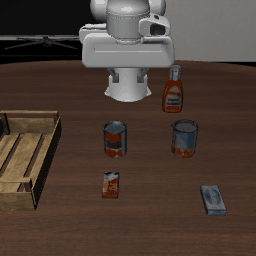}
        \caption{\texttt{Libero}}
        \label{fig:libero}
    \end{subfigure}
    \begin{subfigure}[b]{0.45\textwidth}
        \centering
        \includegraphics[height=\textwidth]{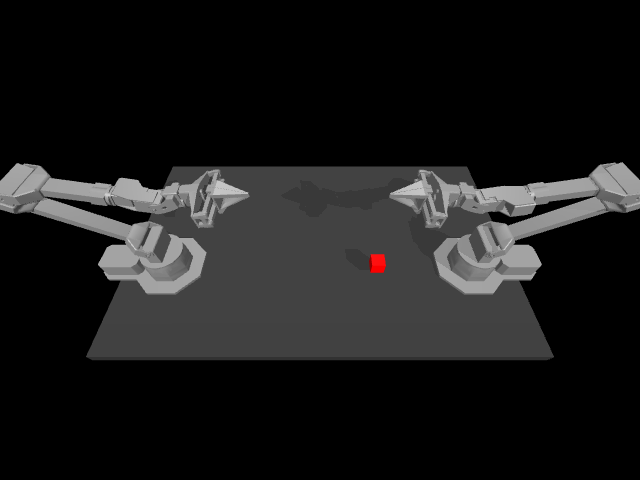}
        \caption{\texttt{AlohaTransferCube}}
        \label{fig:aloha}
    \end{subfigure}
    \caption{Simulation tasks for $\pi_0$.}
    \label{fig:pi0_tasks}
\end{minipage}
\end{figure}

\textbf{Tasks.} In the $\pi_0$ experiments, we study two simulation tasks: \texttt{Libero}~\cite{liu2023libero} and \texttt{AlohaTransferCube}~\cite{zhao2023learning}, as shown in Figure~\ref{fig:pi0_tasks}. The \texttt{Libero} benchmark consists of 130 tasks which are divided into separate suites---\texttt{Libero-\{Spatial, Object, Goal, 10, 90\}}. We study the task ``pick up the cream cheese and put it in the tray'' from the \texttt{Libero-90} suite as our target task for online fine-tuning. The \texttt{AlohaTransferCube} task requires the bimanual robots to pick up the cube with one hand and transfer it to the other. Both tasks use a sparse reward scheme: the agent receives –1 at every time step until it succeeds, at which point it receives 0.

\textbf{Base policy.} We use the public weights from $\pi_0$ for both tasks. Specifically, we use the checkpoint at \texttt{s3://openpi-assets/checkpoints/pi0\_libero} for the \texttt{Libero} task, and \texttt{s3://openpi-assets/checkpoints/pi0\_aloha\_sim} for the \texttt{AlohaTransferCube} task. Both checkpoints were obtained by fine-tuning the $\pi_0$ flow-matching base policy, which was pre-trained on over 10k hours of robot data. The \texttt{Libero} checkpoint was fine-tuned for 30k steps on the \texttt{Libero-\{Spatial, Object, Goal, 10\}} suites, containing a total of 1,693 demonstrations on 40 tasks. Note that our target task is from the \texttt{Libero-90} suite, which is not included in the dataset. This allows us to study how to rapidly adapt the base policy to master a new, unseen task. The \texttt{Aloha} checkpoint was fine-tuned for 20k steps using 50 demonstrations on the same task.

\textbf{Training details.} We use the \algnamesac variant of our method for these experiments. Since both tasks receive visual observations, we train the actor and critic with vision encoders composed of four convolutional layers. 
The $\pi_0$ policy is trained in a noise and action space with a chunk size of 50, where each noise and action is 32-dimensional, resulting in a total dimensionality of 1,600. Naively training the actor $\piw(\bs)$ and critic $\Qw(\bs,\bw)$ is challenging due to this high dimensionality. 

Here, we propose a simple yet effective approach to train \algnamesac on large action chunks. To avoid training the actor and critic on the whole noise space $\cW = (\mathbb{R}^d)^C = \bigl\{\, \bw = (w_1,\dots,w_C)\mid w_i \in \mathbb{R}^d\bigr\},$ where $C$ is the action chunk size and $d$ is the per-step noise dimension, we instead train a single-step actor $\piw_\text{single}(\bs)$ and critic $\Qw_\text{single}(\bs,\bw_\text{single})$ where $\cW_\text{single} = \mathbb{R}^d$. At inference time, we simply sample $w_\text{single} \sim \piw_\text{single}$ and repeat it across the action chunk axis $\bw = \{(w_1,\dots,w_C)\mid w_i = w_\text{single} \}$, and use it to query the $\pi_0$ policy. We find that this makes SAC training more efficient and stable, while the repeated-noise formulation remains expressive enough to steer $\pi_0$ to produce effective actions. This shows that the noise space of pre-trained $\pi_0$ policy is a well-structured latent space, highlighting another advantage of running RL in this latent-noise action space. We provide the hyperparameters in Table~\ref{table:pi0_hyp}.

\paragraph{Real-world experiments.}
For the real-world $\pi_0$ experiments, we utilize the pubic $\pi_0\text{-DROID}$ checkpoint at \texttt{s3://openpi-assets/checkpoints/pi0\_droid}. We adopt the same method to handle large action chunks as is described above, though note that this $\pi_0\text{-DROID}$ variant we used only has an action chunk of size 10. Please see \Cref{fig:pi0_tasks} for visualizations of our setup. We input the proprioceptive state, the final token’s last hidden feature from $\pi_0$’s VLM backbone (a 2,048-dimensional vector), and visual features into the noise policy. The visual features are extracted by training a shallow convolutional encoder on the concatenated images from the left, right, and wrist cameras, following the same procedure described above.
For text commands to $\pi_0$, we use ``turn on the toaster'' and ``put the spoon on the plate''.

We run $\pi_0$ inference on a remote policy server, but run \algname training locally on a NVIDIA GeForce RTX 3070 GPU with 8GB of VRAM. We note that this is a significantly smaller amount of compute than is typically required for $\pi_0$ finetuning---as stated in the $\pi_0$ repository, LoRA of $\pi_0$ requires at least 22.5GB of VRAM, and full finetuning requires 70GB.

\begin{figure}[H]
    \centering
    \begin{subfigure}[b]{0.3\textwidth}
        \centering
        \includegraphics[width=\textwidth]{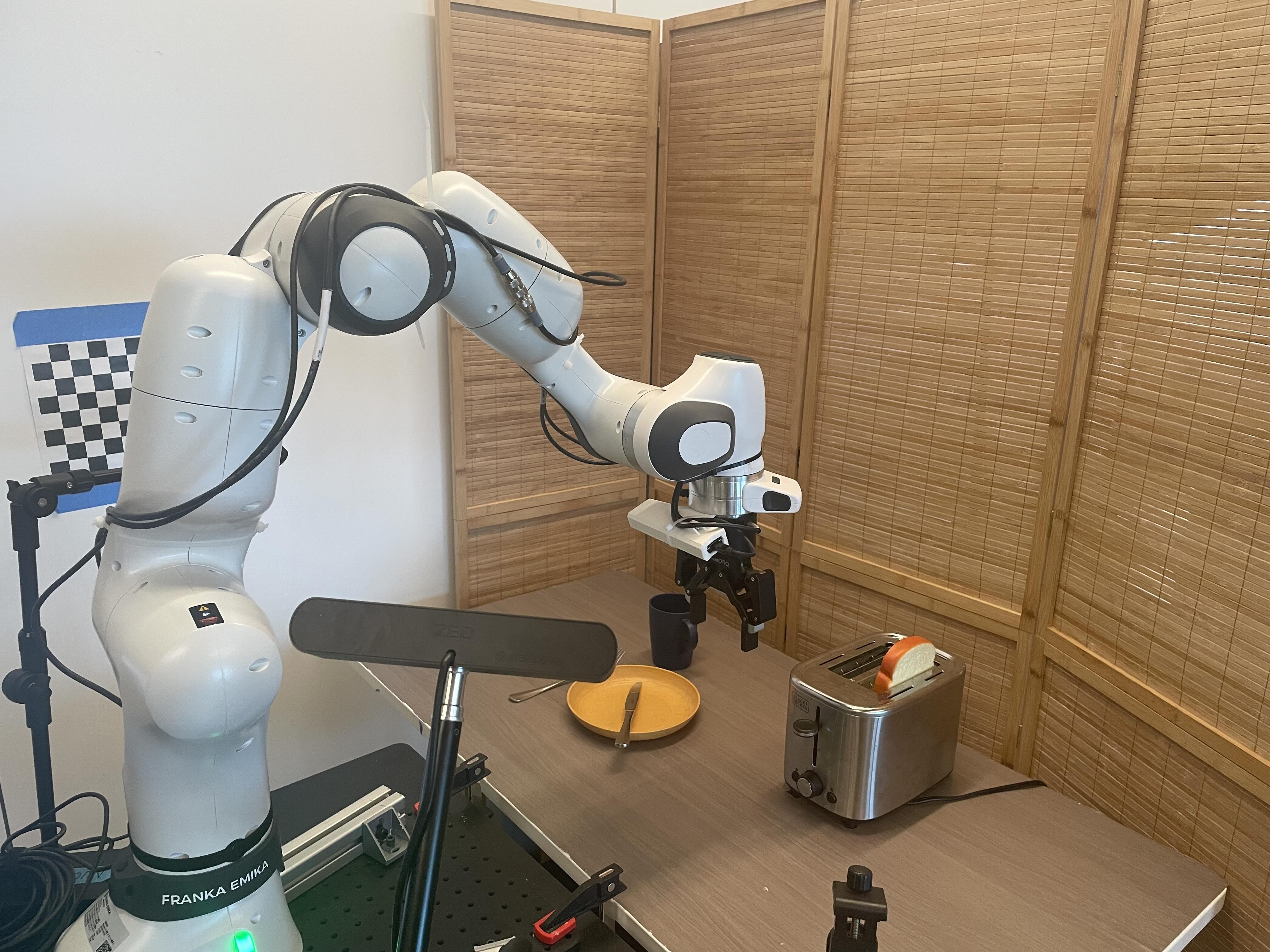}
        \caption{Toaster task.}
        \label{fig:toaster_setup}
    \end{subfigure}
    \hspace{2em}
    \begin{subfigure}[b]{0.3\textwidth}
        \centering
        \includegraphics[width=\textwidth]{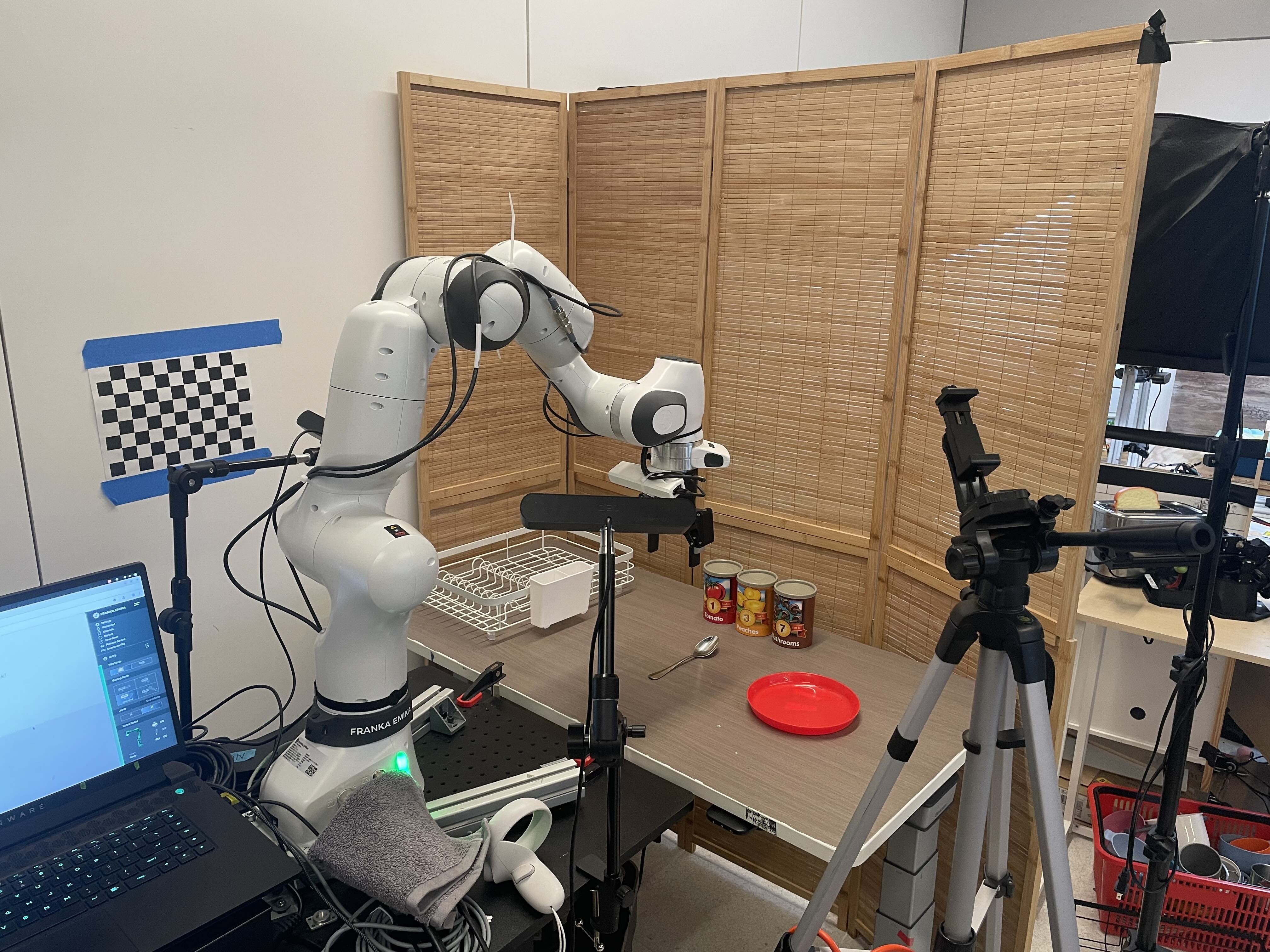}
        \caption{Spoon task.}
        \label{fig:third}
    \end{subfigure}
    \caption{$\pi_0$ real-world task setups.}
    \label{fig:pi0_setup}
\end{figure}

\begin{table}[t]
\caption{
\footnotesize
\textbf{Hyperparameters for \algname \texttt{$\pi_0$} experiments.}
}
\vspace{5pt}
\label{table:pi0_hyp}

\begin{center}
\scalebox{0.9}
{
\begin{tabular}{llll}
    \toprule
    \textbf{Hyperparameter} & \texttt{Libero} & \texttt{AlohaTransferCube} & Real-world \\
    \midrule
Batch size & $256$ & $256$ & $256$ \\
Hidden size & $128$ & $128$ & $1024$ \\
Number of layers & $3$ & $3$ & $3$ \\
Actor learning rate  & $0.0001$ & $0.0001$ & $0.0001$ \\
Critic learning rate & $0.0003$ & $0.0003$ & $0.0003$ \\
Discount factor & $0.999$ & $0.999$ & $0.99$ \\
Target entropy & $-d/2$ & $-d/2$ & 0 \\
Target update rate ($\tau$) & $0.005$ & $0.005$ & $0.005$ \\
Image observation size & $64 \times 64$ & $64 \times 64$ & $128 \times 128$  \\
CNN features & $(32, 32, 32, 32)$ & $(32, 32, 32, 32)$ & $(32, 32, 32, 32)$  \\
Gradient steps per update & $20$ & $20$ & $30$ \\
Action magnitude ($\actmag$) & $1.0$ & $2.0$ & $2.5$ \\
Action chunk size & $50$ & $50$ & $10$ \\
Number of actions to execute & $20$ & $50$ & $10$  \\
Number of critics & $10$ & $10$ & $2$ \\
Flow steps & $10$ & $10$ & $10$ \\
Clipped double Q-learning & False & False & False \\
    \bottomrule
\end{tabular}
}
\end{center}
\end{table}